\documentclass[journal]{IEEEtran} 
\usepackage{graphicx}
\usepackage{cite}
\usepackage{url}
\usepackage{colortbl}
\usepackage{multirow}
\usepackage{amssymb}
\usepackage{mathtools}
\usepackage{bbm}
\usepackage{xcolor}
\usepackage{booktabs}

\usepackage[linesnumbered,ruled,vlined]{algorithm2e}

\usepackage{hyperref}
\hypersetup{
    colorlinks=true,
    linkcolor=red,
    filecolor=magenta,      
    urlcolor=blue,  % colorlinks=true,
}

\definecolor{limegreen}{rgb}{0.2, 0.8, 0.2}
\definecolor{forestgreen}{rgb}{0.13, 0.55, 0.13}
\definecolor{greenhtml}{rgb}{0.0, 0.5, 0.0}
\definecolor{black}{rgb}{0.0, 0.0, 0.0}

\usepackage[font=small,labelfont=bf,tableposition=top]{caption}

\usepackage{blindtext}
\title{here title}

\let\oldtwocolumn\twocolumn
\renewcommand\twocolumn[1][]{%
    \oldtwocolumn[{#1}{
    \begin{center}
           \includegraphics[width=18cm]{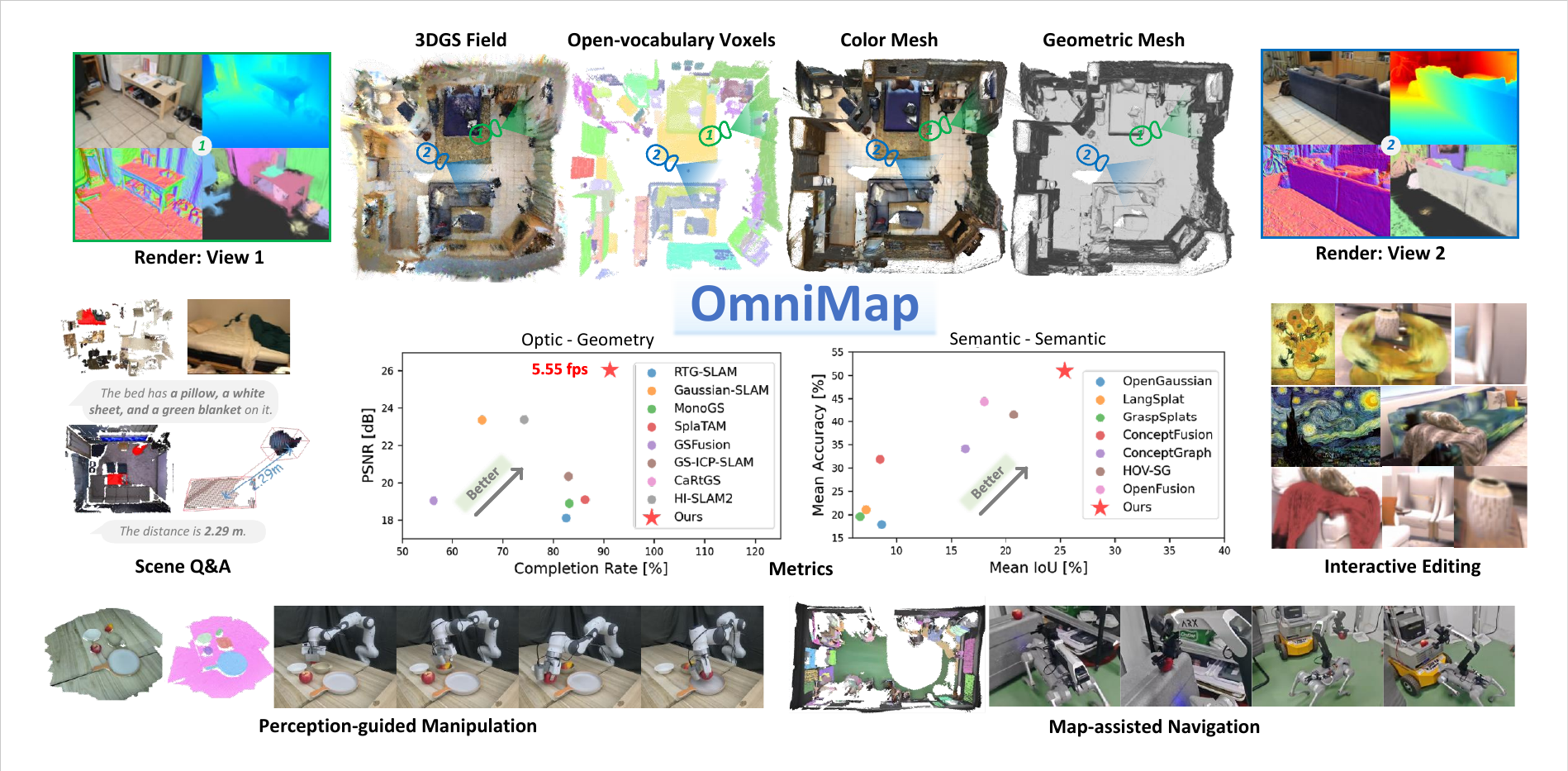}
           \captionof{figure}{We introduce \textbf{OmniMap}, a general online mapping framework integrating optics, geometry, and semantics. OmniMap incrementally maintains an open-vocabulary instance-level voxel representation and a 3DGS (3D Gaussian Splatting) representation, from which color and geometric meshes are derived. OmniMap supports multi-modal rendering (RGB / depth / normal / instance), and achieves state-of-the-art performance in rendering fidelity, mesh quality, and semantic understanding. This holistic framework enables versatile support for a wide range of downstream applications.}
           \label{first}
        \end{center}
    }]
}

\begin{document}

% \pagestyle{empty}
% \markboth{IEEE TRANSACTIONS ON ROBOTICS}
% {Yinan Deng \MakeLowercase{\textit{et al.}}: OmniMap}

\title
{	
OmniMap: A General Mapping Framework Integrating Optics, Geometry, and Semantics

\thanks{This work is supported by the National Natural Science Foundation of China under Grant 92370203, 62473050, 62233002, Beijing Natural Science Foundation Undergraduate Research Program QY24180. \textit{(Corresponding Author: Yufeng Yue)}}
\thanks{Yinan Deng, Yufeng Yue, Jianyu Dou, Jingyu Zhao, Jiahui Wang, Yujie Tang, and Yi Yang are with School of Automation, Beijing Institute of Technology, Beijing 100081, China (e-mail: dengyinan@bit.edu.cn; yueyufeng@bit.edu.cn; BruceDou030806@163.com; unique$\_$zhao0210@163.com; wjh@bit.edu.cn; 3120235697@bit.edu.cn; yang$\_$yi@bit.edu.cn).}
\thanks{Mengyin Fu is with the School of Automation, Beijing Institute of Technology, Beijing 100081, China, and the School of Automation, Nanjing University of Science and Technology, Nanjing 210018, China (e-mail: fumy@bit.edu.cn).}
\thanks{The project page of OmniMap is available at \url{https://omni-map.github.io/}.}
\thanks{Accepted  by IEEE Transactions on Robotics (TRO)}
}

\author{Yinan Deng,  Yufeng Yue$^{*}$, Jianyu Dou, Jingyu Zhao, Jiahui Wang, Yujie Tang, Yi Yang, and Mengyin Fu}

\maketitle

\begin{abstract}

Robotic systems demand accurate and comprehensive 3D environment perception, requiring simultaneous capture of photo-realistic appearance (optical), precise layout shape (geometric), and open-vocabulary scene understanding (semantic). Existing methods typically achieve only partial fulfillment of these requirements while exhibiting optical blurring, geometric irregularities, and semantic ambiguities.
To address these challenges, we propose OmniMap. Overall, OmniMap represents the first online mapping framework that simultaneously captures optical, geometric, and semantic scene attributes while maintaining real-time performance and model compactness. 
At the architectural level, OmniMap employs a tightly coupled 3DGS–Voxel hybrid representation that combines fine-grained modeling with structural stability.
At the implementation level, OmniMap  identifies key challenges across different modalities and introduces several innovations: 
adaptive camera modeling for motion blur and exposure compensation, hybrid incremental representation with normal constraints, and probabilistic fusion for robust instance-level understanding.
Extensive experiments show OmniMap's superior performance in rendering fidelity, geometric accuracy, and zero-shot semantic segmentation compared to state-of-the-art methods across diverse scenes. 
The framework's versatility is further evidenced through a variety of downstream applications, including multi-domain scene Q\&A, interactive editing, perception-guided manipulation, and map-assisted navigation.

\end{abstract}


\begin{table*}[]
\centering
\scriptsize
\caption{Comparison of representations of some classic and recent mapping domains.}
\label{mapping_domain}
\renewcommand\arraystretch{1.1}
\setlength{\tabcolsep}{2mm}{
\begin{tabular}{c|c|c|c|c|c}
\toprule
\begin{tabular}[c]{@{}c@{}}Robotic Mapping\\Research Area \end{tabular} & \begin{tabular}[c]{@{}c@{}}\textbf{Optics}\\      High-Fidelity Rending ($\checkmark$)\\      Discrete Appearance ($\sqrt{}\mkern-9mu{\smallsetminus}$)\end{tabular} & \begin{tabular}[c]{@{}c@{}}\textbf{Geometry}\\      Detailed Mesh ($\checkmark$)\\      Sparse Structure ($\sqrt{}\mkern-9mu{\smallsetminus}$)\end{tabular} & \begin{tabular}[c]{@{}c@{}}\textbf{Semantic}\\      Open Understanding ($\checkmark$)\\      Closed-Set Labels ($\sqrt{}\mkern-9mu{\smallsetminus}$)\end{tabular} & \textbf{Online} & Representative Works \\ \midrule
(Semantic)   Volume Mapping & $\sqrt{}\mkern-9mu{\smallsetminus}$ & $\sqrt{}\mkern-9mu{\smallsetminus}$ & $\sqrt{}\mkern-9mu{\smallsetminus}$ & $\checkmark$ & OctMap \cite{OctoMap}, Voxblox \cite{Voxblox} \\ 
Surface Reconstruction & $\sqrt{}\mkern-9mu{\smallsetminus}$ & $\checkmark$ & $\times$ & $\checkmark$ & KinectFusion \cite{Kinectfusion}, BundleFusion \cite{Bundlefusion} \\ 
Maps For View Synthesis & $\checkmark$ & $\checkmark$ & $\times$ & $\times$ & NeRF \cite{nerf}, 3DGS \cite{3DGS} \\
OV Incremental Mapping & $\times$ & $\sqrt{}\mkern-9mu{\smallsetminus}$ & $\checkmark$ & $\checkmark$ & ConceptGraphs \cite{conceptgraphs}, HOV-SG \cite{hov-sg} \\
OV NeRF/3DGS & $\checkmark$ & $\sqrt{}\mkern-9mu{\smallsetminus}$ & $\checkmark$ & $\times$ & LERF \cite{lerf}, LangSplat \cite{langsplat} \\ 
\midrule
OmniMap & $\checkmark$ & $\checkmark$ & $\checkmark$ & $\checkmark$ & -
\\ \bottomrule
\end{tabular}
}
\\
\vspace{0.5em} 
{\footnotesize
\raggedright
\textbf{Note:} This table outlines a partial yet representative selection of research areas in robotic mapping.
}
\end{table*}

\section{Introduction}

% 3D perception and representation of the surrounding environment form a critical foundation for robots to perform downstream tasks. The accuracy and dimensionality of these representations directly influence task performance and operational capabilities.
% An ideal representation should comprehensively capture all essential scene attributes, including optics (high-fidelity appearance), geometry (detailed structure), and semantics (high-level understanding).
% However, most widely adopted mapping algorithms remain constrained by fixed-resolution geometric representations \cite{cai2023occupancy, chen2023continuous, yuan2024uni} and closed-set semantic \cite{see-csom, semantic-octree, hd-ccsom}. 
% While these approaches adequately support conventional tasks like obstacle avoidance and point-to-point navigation, the absence of photorealistic appearance modeling and open-set semantic recognition limits their effectiveness in meeting the growing demands for environmental representation in the era of embodied intelligence.

The quality of a robot's 3D environmental representation, measured by its accuracy and dimensionality, fundamentally impacts the robot's task operational performance and execution capabilities. 
The emergence of Embodied Artificial Intelligence (Embodied AI) has positioned robots as crucial physical agents that bridge AI systems with the real world, driving increasingly sophisticated representation requirements.
An ideal environmental representation should enable high-precision, multi-dimensional understanding of the scene, while simultaneously addressing a variety of critical aspects: (1) \textbf{optics:} high-fidelity appearance rendering from arbitrary views, (2) \textbf{geometry:} fine-grained structural reconstruction, and (3) \textbf{semantics:} open-vocabulary object recognition and scene understanding. 
This representation should support diverse task modes, including interactive virtual operations (scene Q\&A and editing) and multi-granular physical interactions (desktop-level manipulation and room-level navigation).

However, current mapping frameworks remain narrowly focused, capturing only limited aspects of the scene. Tab. \ref{mapping_domain} summarizes and compares several classic and recent representations in the domain of robotic mapping.
Most classic (semantic) volume algorithms rely on fixed-resolution geometry \cite{cai2023occupancy, chen2023continuous, yuan2024uni} and closed-set semantic \cite{see-csom, semantic-octree, hd-ccsom}, restricting their functionality to basic tasks like obstacle avoidance and point-to-point navigation. 
Surface reconstruction methods \cite{Kinectfusion, Bundlefusion} are capable of recovering fine geometric details, but they typically lack semantic modeling and still fall short in achieving high-quality rendering.
While recent advances in Neural Radiance Fields (NeRF) \cite{nerf} and 3D Gaussian Splatting (3DGS) \cite{3DGS} have demonstrated promising capabilities for photorealistic view synthesis, and have even been adapted into SLAM systems \cite{Gaussian-slam, Splatam, monogs} for online reconstruction, they still primarily address only optical and geometric recovery.
Open-vocabulary (OV) mapping leverages the cognitive power of large language models (LLMs) and vision-language models (VLMs) to achieve rich semantic scene interpretation. Yet, existing incremental approaches \cite{conceptfusion, conceptgraphs, opengraph, hov-sg, openin} in this domain often rely on sparse point cloud representations, which inherently lack the density required for high-quality rendering and detailed scene reconstruction. 
The integration of OV techniques with NeRF/3DGS appears promising for addressing these limitations, however, their batch-style processing makes them challenging to deploy in real-time robotic systems.

This naturally leads us to a fundamental question: \textbf{\textit{Is it feasible to develop an online mapping algorithm capable of holistically integrating all these scene attributes while simultaneously improving the precision of each individual attribute?}}
To answer this question, we present OmniMap, as shown in Fig. \ref{first}. To better highlight its contributions, we break down this challenge into the following key sub-questions:

\textbf{1) Representation: how to integrate diverse scene attributes within a unified framework?}
A key challenge in 3D scene mapping lies in effectively representing and integrating diverse scene attributes within a unified framework.
Existing methods often fall short in online robotic systems, being constrained to a single representation \cite{OctoMap, Kinectfusion, nerf, 3DGS, conceptgraphs} or a naïve combination thereof \cite{Gsfusion, nice-slam, activegs}.
To address this, OmniMap adopts a highly complementary and tightly coupled hybrid 3DGS-Voxel representation.
Voxels possess a stable and structured spatial format and often serve as the cornerstone for local probabilistic fusion, though their expressive capacity is inherently constrained by resolution.
Differentiable 3DGS provides theoretically arbitrarily fine-grained representations, but due to its positional immobility, it is difficult to directly correlate semantics during optimization, and it faces ambiguity in discriminating newly added Gaussians during incremental mapping.
Therefore, OmniMap organically combines the two: voxels act as the fundamental unit for incremental mapping, encapsulating probabilistic semantics within their structure, while newly assigned voxels guide the initialization of Gaussian primitives that underpin precise optical and geometric reconstruction.
This integration preserves the fine detail modeling of 3DGS while leveraging the voxel grid’s strengths in probabilistic fusion and structural stability.

\textbf{2) Optics: how to efficiently handle camera motion blur and exposure inconsistencies?}
Camera motion, whether from a mobile chassis or a moving end-effector, frequently induces motion blur in captured images. 
While some algorithms address this issue through techniques like ray distortion \cite{deblur-nerf}, primitive shifting \cite{deblurring-3dgs}, or local linearization \cite{bad-gs}, their high time complexity makes them unsuitable for online dense mapping systems.
Additionally, these methods fail to handle exposure variations, a common challenge in unevenly illuminated environments.
In contrast, OmniMap employs a lightweight differentiable camera model,  which enhances appearance reconstruction quality while maintaining low computational overhead.
% utilizing only two offset parameters to simulate motion blur and two weighting parameters for adaptive exposure compensation. 
% Experimental results demonstrate that this approach significantly enhances appearance reconstruction quality while maintaining low computational overhead.

\textbf{3) Geometry: how to improve geometry stability during 3DGS incremental mapping?}
The incremental update of Gaussian primitives poses significant challenges for online 3DGS systems. Current approaches typically follow a brute-force approach: they first back-project RGB-D data to generate colored point clouds, then apply downsampling for Gaussian initialization. 
This process fails to differentiate observed from unobserved regions, causing redundant primitives in over-scanned areas and insufficient coverage elsewhere.
OmniMap addresses these limitations by integrating TSDF-Fusion's incremental framework (replaceable with other voxel-based mapping methods), strategically initializing Gaussians only within newly assigned voxels to maintain model compactness and geometric stability.
Additionally, OmniMap incorporates normal-based supervision directly derived from the rendered depth, significantly enhancing the accuracy of detail modeling and the smoothness of the mesh surface.

\textbf{4) Semantics: how to robustly recognise and fuse instances in open environments?}
While VLMs like CLIP enable open-set object recognition within instance masks, their textual inference capabilities remain limited. OmniMap enhances this through an LLM-augmented front-end pipeline. 
Furthermore, segmentation models often exhibit poor intra-frame accuracy (under/over-segmentation) and inter-frame consistency. Exciting incremental methods using IoU-based association with 3D boxes \cite{conceptgraphs} or 2D masks \cite{open-fusion} prove sensitive to segmentation noise. OmniMap's novel solution employs probabilistic instance tuples over TSDF-Fusion voxels, decomposing fusion into two robust subtasks: instance association and live map evolution, modeled as MLE and MAP problems, respectively. This probabilistic formulation naturally preserves uncertainty for improved noise resilience.

\begin{figure}[!t]\centering
	\includegraphics[width=7.5cm]{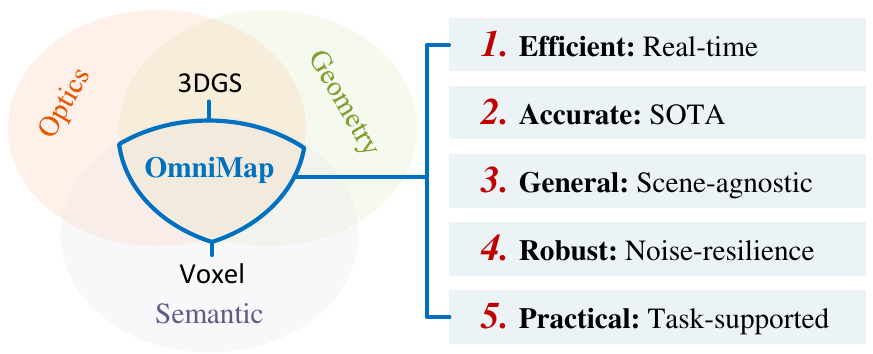}
	\caption{OmniMap facilitates a holistic representation that integrates optics, geometry, and semantics, and provides five key advantages.}
	\label{map_domain} 
\end{figure}

In summary, OmniMap successfully integrates three critical scene attributes within a general online mapping framework while achieving significant performance improvements in all dimensions: high-fidelity optical rendering, geometrically stable reconstruction, and open-vocabulary semantic understanding. 
As demonstrated in Fig. \ref{first}, our framework not only surpasses all SOTA methods in quantitative metrics, but also enables versatile downstream applications through its holistic representation.
OmniMap offers five key advantages: \textbf{a) Efficient}: Real-time operation at sufficient frame rate with compact model size; \textbf{b) Accurate}: High precision in rendering, mesh, and cognition; \textbf{c) General}: No pre-training or scene-specific optimization required; \textbf{d) Robust}: Resilient to sensor and segmentation noise; \textbf{e) Practical}: Supports diverse downstream tasks.
Our contributions are summarized as follows:
\begin{itemize}
    \item \textbf{Comprehensive Representation:} OmniMap establishes the first general framework bridging optics, geometry, and semantics in online mapping, based on a hybrid 3DGS-Voxel representation.
    \item \textbf{High-Fidelity Optics:} We model the camera motion blurring and exposure process based on four differentiable parameters that efficiently and effectively improve rendering quality.
    \item \textbf{Detailed Geometry:} We propose an incremental 3DGS strategy where newly assigned voxels initialize Gaussian primitives, and introduce higher-order normal supervision for recovering fine-scale structures.
    \item \textbf{Open Semantics:} We design an efficient instance understanding pipeline and mathematically model the incremental fusion process as two subtasks, enabling adaptation and recovery from noise.
    \item \textbf{SOTA Performance:} OmniMap achieves superior metrics in image rendering, mesh reconstruction, and zero-shot semantic segmentation, while maintaining a high frame rate and compact model size.
\end{itemize}

% The rest of this article is organized as follows. Section \ref{RW} provides an overview of related works. 
% We then present the complete mapping framework and the details of each module in Sections \ref{OV}, \ref{2L}, \ref{PR}, and \ref{MR}, respectively. 
% Section \ref{BR} presents the benchmark experiments and ablation studies conducted on public datasets. 
% In Section \ref{Ap}, we demonstrate some applications.
% Finally, Section \ref{C} concludes this article.

\section{Related Works} \label{RW}

In this section, we review previous related research on robotic mapping, including discrete reconstruction, 3DGS-based dense mapping, and open-vocabulary representation.

\subsection{Discrete Reconstruction}

Classical representation structures for robotic mapping, such as point clouds and voxels, have been extensively developed over the past decades. The widely adopted OctoMap \cite{OctoMap} offers an efficient 3D probabilistic volumetric representation via octrees. Meyer-D et al. \cite{Occupancy} model dynamic environments by introducing state transition probabilities in Hidden Markov Models for occupancy grids. 
% Beyond occupancy, TSDF and ESDF representations are also widely integrated into online reconstruction frameworks, including KinectFusion \cite{Kinectfusion}, BundleFusion \cite{Bundlefusion}, Voxblox \cite{Voxblox}, and FIESTA \cite{Fiesta}.
The advancement of deep learning has significantly accelerated progress in robotic semantic understanding \cite{Ssgm, tang2023multi, tang2023robust}, enabling numerous recent breakthroughs. 
% DA-RNN \cite{DA-RNN} introduces Data Associated Recurrent Neural Networks for semantic annotation of RGB-D video using a novel recurrent architecture. 
SemanticFusion \cite{semanticfusion} integrates CNN-based semantic predictions with ElasticFusion and refines them via CRF-based map updates. Semantic-OcTree \cite{semantic-octree} proposes a Bayesian multi-class octree mapping method, where semantic categories are probabilistically updated. However, these methods are purely discrete and overlook the inherently continuous nature of the real world.

To construct smoother occupancy maps, many methods relax the assumption of voxel independence. GPmap \cite{GPmap} models point dependencies via nonparametric Bayesian inference. Hilbert maps \cite{hilbert} leverage fast kernel approximations for efficient training, enabling real-time incremental 3D mapping. Recently, Bayesian kernel inference with lower computational complexity has gained attention. BGKOctoMap \cite{BGK} applies sparse kernels and nonparametric inference to enhance efficiency.
% , while Lu et al. \cite{gan2022multitask} extends binary occupancy mapping to multiclass and multitask domains, enriching map representations. 
To address over-inflation issues, S-MKI \cite{S-MKI} and SEE-CSOM \cite{see-csom} introduce adaptive multi-entropy kernel inference, improving both efficiency and accuracy in continuous semantic occupancy mapping.

Although these frameworks have served as milestones in robotic mapping, their reliance on discrete sparse representations limits scalability. While continuous occupancy mapping alleviates some issues, it remains constrained by map resolution, lacks rendering capabilities, and is restricted to closed-set semantics, falling short of the representation demands in embodied intelligence applications.

\subsection{3DGS-Based Dense Mapping}

In recent years, learning-based implicit mapping methods have emerged, inspired by NeRF's strengths in continuous representation and appearance modeling. These approaches have been applied to incremental SLAM \cite{Lgsdf, MACIM, imap}, enabling the generation of photorealistic renderings with compact models.
Although some efforts focus on accelerating vanilla NeRF, the computational burden of volumetric rendering and implicit representation continues to pose a significant challenge for the practical deployment of NeRF-like methods.
More recently, 3DGS \cite{3DGS} achieved real-time novel view rendering without sacrificing quality by leveraging a tile-based differentiable splatting technique. However, its original design is tailored for offline processing and typically requires tens of minutes to batch process small-scale datasets, making it unsuitable for exploratory robotics where full-scene sensor data is unavailable upfront.

To address this, recent studies have integrated 3DGS with visual SLAM systems and proposed targeted strategies for online optimization. 
GS-SLAM \cite{Gs-slam} employs an adaptive scaling strategy to dynamically add new or remove noisy Gaussians, while selecting Gaussians for camera pose optimization through a coarse-to-fine approach. By decomposing the scene into independently optimized sub-maps, Gaussian-SLAM \cite{Gaussian-slam} enables online optimization of newly explored regions regardless of scene scale. SplaTAM \cite{Splatam} leverages rendered depth and silhouette images to effectively capture scene density distribution to guide Gaussian insertion. RTG-SLAM \cite{Rtg-slam} densifies Gaussians based on geometry and optics, enforcing complete opacity or transparency to better represent dominant and residual surface colors, respectively. MonoGS \cite{monogs} places Gaussians through depth back-projection and applies isotropic regularization to reduce artifacts in novel views. Splat-SLAM \cite{Splat-slam} dynamically adapts to keyframe pose and depth updates through active map deformation.

Recent approaches have focused on balancing the quality and efficiency of dense 3DGS mapping. GS-ICP-SLAM \cite{gs-icp-slam} integrates Generalized Iterative Closest Point with 3DGS and adds mapping keyframes alongside tracking keyframes. CaRtGS \cite{CaRtGS} an adaptive computational alignment strategy that effectively tackles insufficient optimization, long-tail convergence, and weakly constrained densification. 
HI-SLAM2 \cite{hi-slam2}, built upon DROID-SLAM, incorporates monocular geometry priors (depth and normals) to enhance 3DGS geometric estimation. 
However, these methods lack explicit 3D Gaussian specification, making them susceptible to over- and under-fitting. 
Most closely related to our work is GSFusion \cite{Gsfusion}, which combines TSDF-Fusion and 3DGS in a hybrid representation. Yet, it still relies on depth backprojection for Gaussian initialization, using voxels only as a filtering criterion. 
% In contrast, OmniMap initializes new Gaussians directly from new assigned voxels, improving structural stability while reducing the number of Gaussians.

More importantly, existing methods often neglect two key challenges introduced by camera motion: motion blur during movement and exposure inconsistency across viewpoints in non-uniformly lit scenes. While offline techniques \cite{deblurring-3dgs} exist for blur mitigation, their high computational cost hinders real-time use. OmniMap addresses this by explicitly modeling per-frame camera parameters with four differentiable variables to capture blur and exposure configurations, and experimental results validate the effectiveness of this design.

\begin{figure*}[!t]\centering
	\includegraphics[width=17.5cm]{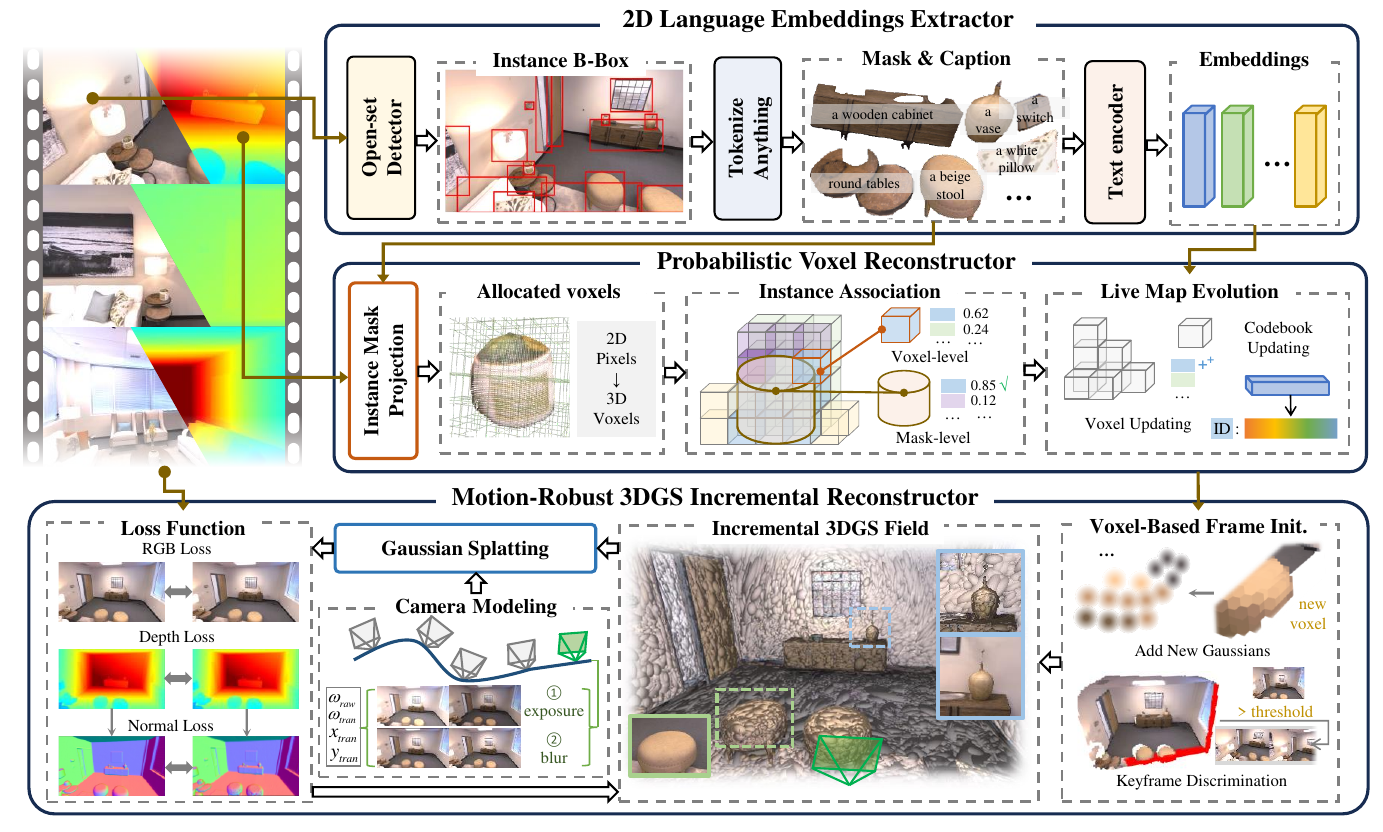}
	\caption{The system framework of OmniMap consists of three main modules. 
    The \textbf{2D Language Embeddings Extractor} sequentially combines multiple open-set models for detection, segmentation, captioning, and embedding extraction of objects. 
    The \textbf{Probabilistic Voxel Reconstructor} incrementally integrates per-frame instance masks and embeddings into 3D space while maintaining voxel-aligned probabilistic instance tuples and a global embedding codebook.
    The \textbf{Motion-Robust 3DGS Incremental Reconstructor} initializes new Gaussians from new assigned voxels and renders 2D images under RGB, depth, and normal modalities with parameterized camera models for supervision.}
	\label{framework} 
\end{figure*}

\subsection{Open-Vocabulary Representation}

To extend semantics from closed-set pre-trained labels to open-set categories, researchers have developed open-vocabulary mapping systems leveraging VLMs and LLMs for zero-shot generalization. Early approaches typically extracted pixel-level 2D VLM features, which are then projected or distilled into 3D space. ConceptFusion \cite{conceptfusion} combines global and local features into pixel-aligned ones, which are subsequently fused into a 3D point cloud via multi-view fusion, showcasing cross-modal inference. Based on NeRF, DFF \cite{dff} distills the knowledge of self-supervised 2D image feature extractors such as into a 3D feature field optimized in parallel to the radiance field. Similarly, LERF \cite{lerf} learns a dense, multi-scale language field inside NeRF by volume rendering CLIP embeddings. OpenObj \cite{Openobj} and OpenMulti \cite{openmulti} builds multi-granularity language fields by distilling part features into objects' NeRFs.

Some other methods enhance the field by leveraging a collection of 3D Gaussians. LangSplat \cite{langsplat} leverages SAM's multi-granularity masks for feature extraction and CLIP encoding, incorporating an autoencoder to embed compressed semantics into Gaussians. Gaussian Grouping \cite{Gaussian_grouping} augments each Gaussian with a compact Identity Encoding, enabling grouping of Gaussians based on object instances in the 3D scene. OpenGaussian \cite{opengaussian} trains feature consistency within instances and employs a two-phase codebook to distinguish between coarse and fine features.
Although integration with 3DGS enables fast feature rendering in these methods, most employ a two-stage optimization process: geometry first, followed by semantics.
To date, no approach has directly incorporated open-vocabulary understanding into the incremental mapping of 3DGS, as the continuous displacement of Gaussian centers during optimization disrupts the stability of their associated semantic attributes.

Recently, several algorithms have advanced incremental instance-level or region-level open-vocabulary mapping. ConceptGraphs \cite{conceptgraphs} compute associations between frames according to the IoU of the bounding box of the point cloud obtained by back-projection of the instance segmentation mask, and fuse features from different instances to obtain the final instance-level point cloud. OpenGraph \cite{opengraph} extends ConceptGraphs to a wide range of outdoor environments, and builds hierarchical scene graphs for structured querying. OpenFusion \cite{open-fusion} extracts region-level features with SEEM and achieves fast cross-frame correlation through 2D mask and confidence rendering. 
However, these methods rely on empirically determined IoU threshold parameters and cannot fully mitigate the impact of robust segmentation at the front end. 
In contrast, OmniMap improves noise recovery by attaching probabilistic instance tuples to each voxel, utilizing probabilistic modeling techniques derived from a theoretical extension of OpenVox \cite{OpenVox}. 
Furthermore, OmniMap significantly extends the capabilities of OpenVox by incorporating more comprehensive scene attributes, including optical and geometric properties.
% In contrast, OmniMap appends probabilistic instance tuples to each voxel, using probabilistic modeling to improve noise recovery . Although front-end segmentation models may lack intra-frame accuracy and inter-frame consistency, OmniMap achieves precise open-vocabulary instance mapping.

\section{Overview} \label{OV}

OmniMap presents a general framework for robotic online mapping, enabling the simultaneous realization of optically, geometrically, and semantically accurate representations.
OmniMap processes RGB-D video streams in real-time, including RGB image frames $\mathcal{C} =\left\{ {{\textbf{C}}_{1}},{{\textbf{C}}_{2}},...,{{\textbf{C}}_{t}} \right\}$, depth image frames $\mathcal{D} =  \left\{ {{\textbf{D}}_{1}},{{\textbf{D}}_{2}},...,{{\textbf{D}}_{t}} \right\}$, and camera poses $\mathcal{P} = \left\{ {{\textbf{P}}_{1}},{{\textbf{P}}_{2}},...,{{\textbf{P}}_{t}} \right\}$ (where $t$ is the timestamp index). 

As shown in Fig. \ref{first}, it constructs a complete 3DGS field $\mathcal{G}_t$ consisting of Gaussian primitives $\mathcal{G}_t=\{g^i_t\}$ (where $i$ is the Gaussian index) and open-vocabulary voxel map $\mathcal{M}_t$ consisting of probabilistic instance-level voxels $\mathcal{V}_t=\{v^j_t\}$ (where $j$ is the voxel index) and an embedding codebook $\mathcal{B}_t=\{f^{\gamma }_t \mid \gamma \in \Gamma \}$ (where $\gamma$ is the instance index) that captures instance-level semantic understanding $f^{\gamma}_t$, where $\Gamma$ is the set of all current instances. 
Fig. \ref{framework} illustrates the system framework, which consists of three core modules:

\textbf{i) 2D Language Embeddings Extractor:} 
This module extracts image embeddings for instance-level scene understanding in the 2D domain. OmniMap employs an efficient, robust pipeline that enhances semantic reasoning through the textual encoding of object captions.

\textbf{ii) Probabilistic Voxel Reconstructor:} 
This module performs incremental open-set instance fusion in the 3D domain. To address front-end segmentation noise, OmniMap employs a voxel-based probabilistic modeling framework.

\textbf{iii) Motion-Robust 3DGS Incremental Reconstructor:} 
This module performs incremental 3DGS scene construction. By deriving new Gaussians and keyframes from voxels, OmniMap implements a lightweight and regularized Gaussian field. Meanwhile, camera parametric modeling and normal supervision improve the optical and geometric quality.

\section{2D Language Embeddings Extractor} \label{2L}

Objects in open environments exhibit diverse types and structural distributions, posing challenges for classical deep learning-based semantic segmentation methods to maintain robust generalization across varied scenes.
Recent advances in VLMs and LLMs offer new paradigms for open understanding. 
Unlike pixel-aligned feature extraction approaches \cite{conceptfusion, dff, lerf}, OmniMap adopts an instance-centric paradigm, treating instances as fundamental decomposable primitives for physical world representation, and focuses on capturing instance-level scene information.

\subsection{Open-set Instance Decection and Segmentation} \label{PIDS}

At time $t$, we first apply the real-time open-vocabulary detection model YOLO-World \cite{Yolo-world}, denoted as $\operatorname{Det}(\cdot)$, to detect instances in the image ${\textbf{C}}_{t}$. Detected targets with confidence scores exceeding a predefined threshold are localized via bounding boxes. Although YOLO-World provides coarse category predictions, it lacks fine-grained attributes (e.g., color, shape) and precise segmentation masks.

To address this, we seamlessly integrate the TAP model \cite{tap}, denoted as $\operatorname{SegCap}(\cdot)$, as a backend to YOLO-World. Prompted by YOLO-World's bounding boxes, TAP concurrently produces pixel-level segmentation masks and corresponding descriptive captions. The output masks are denoted as $\left\{m_{t}^{k}\right\}$ (where $k$ is the mask index) and the captions as $\left\{c_{t}^{k}\right\}$, encapsulating essential visual characteristics. To handle overlapping masks in the raw outputs, we apply a post-processing step that prioritizes smaller masks over larger ones and performs edge erosion.
This process can be expressed as:
\begin{equation}
\begin{aligned}
    \label{tap}
    \{m_{t}^{k}, c_{t}^{k}\} =  \operatorname{SegCap} \left( \operatorname{Det}(\textbf{C}_t) \right)
\end{aligned}
\end{equation}

\subsection{LLM-Enhanced Embedding Encoding}

The textual captions $\left\{c_{t}^{k}\right\}$ , while human-interpretable, present challenges for robotic systems. 
Trained on large-scale corpora, LLMs demonstrate human-like reasoning capabilities. Inspired by this, OmniMap leverages LLMs $\operatorname{Enc}(\cdot)$ to encode captions for enhanced semantic representations.  Specifically, we utilize the SBERT model \cite{sbert}, which satisfies the critical requirement of mapping variable-length texts to fixed-dimensional embeddings. The resulting caption embeddings are denoted as $\left\{ f_{t}^{k} \right\}$:
\begin{equation}
\begin{aligned}
    \label{sbert}
     \{f_{t}^{k}\} = \operatorname{Enc} \left( c_{t}^{k} \right)
\end{aligned}
\end{equation}

After processing with above complete pipeline, we derive for the 2D image-size masks $\left\{m_{t}^{k}\right\}$ and caption embeddings $\{f_{t}^{k}\}$ from current RGB image ${\textbf{C}}_{t}$, which will be grounded into 3D space in the next module.

% \begin{figure}[!t]\centering
% 	\includegraphics[width=8.8cm]{FIG/under_seg.pdf}
% 	\caption{A 2D illustration of incremental instance mapping for OmniMap and ConceptGraphs \cite{conceptgraphs} is shown here. Probabilistic modeling allows OmniMap to achieve more robust instance association and fusion, while ConceptGraphs is prone to failure in such cases. These failures will compound subsequent errors in a continuous incremental setting. Note that at time 11 we only show the correlation calculation for the upper half of the region.}
% 	\label{under_seg} 
% \end{figure}

\section{Probabilistic Voxel Reconstructor} \label{PR}

After segmenting and comprehending the current frame $\textbf{C}_t$, we perform incremental instance-level reconstruction to incorporate the results $\left\{m_{t}^{k}, f_{t}^{k}\right\}$ into the voxel map $\mathcal{M}_{t-1}$. Throughout this process, we adopt a probabilistic modeling framework to enhance the robustness of mapping, as shown in Fig. \ref{asso}. The map is represented by probabilistic voxels $\mathcal{V}$, with each voxel $v^j$ storing a probabilistic instance ID tuple $\theta^j$.
In practice, we apply the TSDF-Fusion algorithm \cite{tsdf-fusion} integrated with the Open3D framework. To optimize memory usage, the voxel map is composed of active hashing blocks, distributed only in the vicinity of obstacles, with each block containing $8^3$ voxels.
Open-vocabulary understanding is maintained via a separate embedding codebook $\mathcal{B}$ that links each instance ID $\gamma$ to its fused caption embedding $f^{\gamma}$.

\subsection{Problem Definition and Decomposition}

The incremental instance-level voxel mapping problem is defined as follows:

Given the current frame's sensor observations and 2D instance cognition results ${\mathcal{Q}_{t}}=\{\{m_{t}^{k}, f_{t}^{k}\},{{\textbf{D}}_{t}},{{\textbf{P}}_{t}}\}$, determine $\mathcal{I}_{t}$ the updated map ${\mathcal{M}_{t}}=\{\mathcal{V}_{t}, \mathcal{B}_{t}\}$ from the map at the last moment ${\mathcal{M}_{t-1}}$:
\begin{equation}
\begin{aligned}
    \label{problem}
    P(\mathcal{I}_{t},{\mathcal{M}_{t}}\mid{\mathcal{M}_{t-1}},{\mathcal{Q}_{t}})
\end{aligned}
\end{equation}
where $\mathcal{I}_{t}=\{\mathcal{I}_{t}^k\}$ represents the instance IDs assigned to all masks $\{m_{t}^{k}\}$, indicating their correspondence to the existing map instances $\Gamma_{t-1}$.

Applying the chain rule, we derive the problem as:
\begin{equation}
\begin{aligned}
    \label{problem_refine}
    &P({{\mathcal{I}}_{t}},{{\mathcal{M}}_{t}}\mid {{\mathcal{M}}_{t-1}},{\mathcal{Q}_{t}})\\
    =&\underbrace{P({{\mathcal{I}}_{t}}\mid {{\mathcal{M}}_{t-1}},{{\mathcal{Q}}_{t}})}_{\text{instance association}}\cdot \underbrace{P({{\mathcal{M}}_{t}}\mid {{\mathcal{M}}_{t-1}},{\mathcal{Q}_{t}},{{\mathcal{I}}_{t}})}_{\text{live map evolution}}
\end{aligned}
\end{equation}
This involves two subtasks: the instance association task $P({\mathcal{I}_{t}} \mid {\mathcal{M}_{t-1}},{\mathcal{Q}_{t}})$ and the live map evolution task $P({\mathcal{M}_{t}} \mid {\mathcal{M}_{t-1}},\mathcal{Q}_{t},{\mathcal{I}_{t}})$.

\subsection{Instance Association}

\begin{figure}[!t]\centering
	\includegraphics[width=8.8cm]{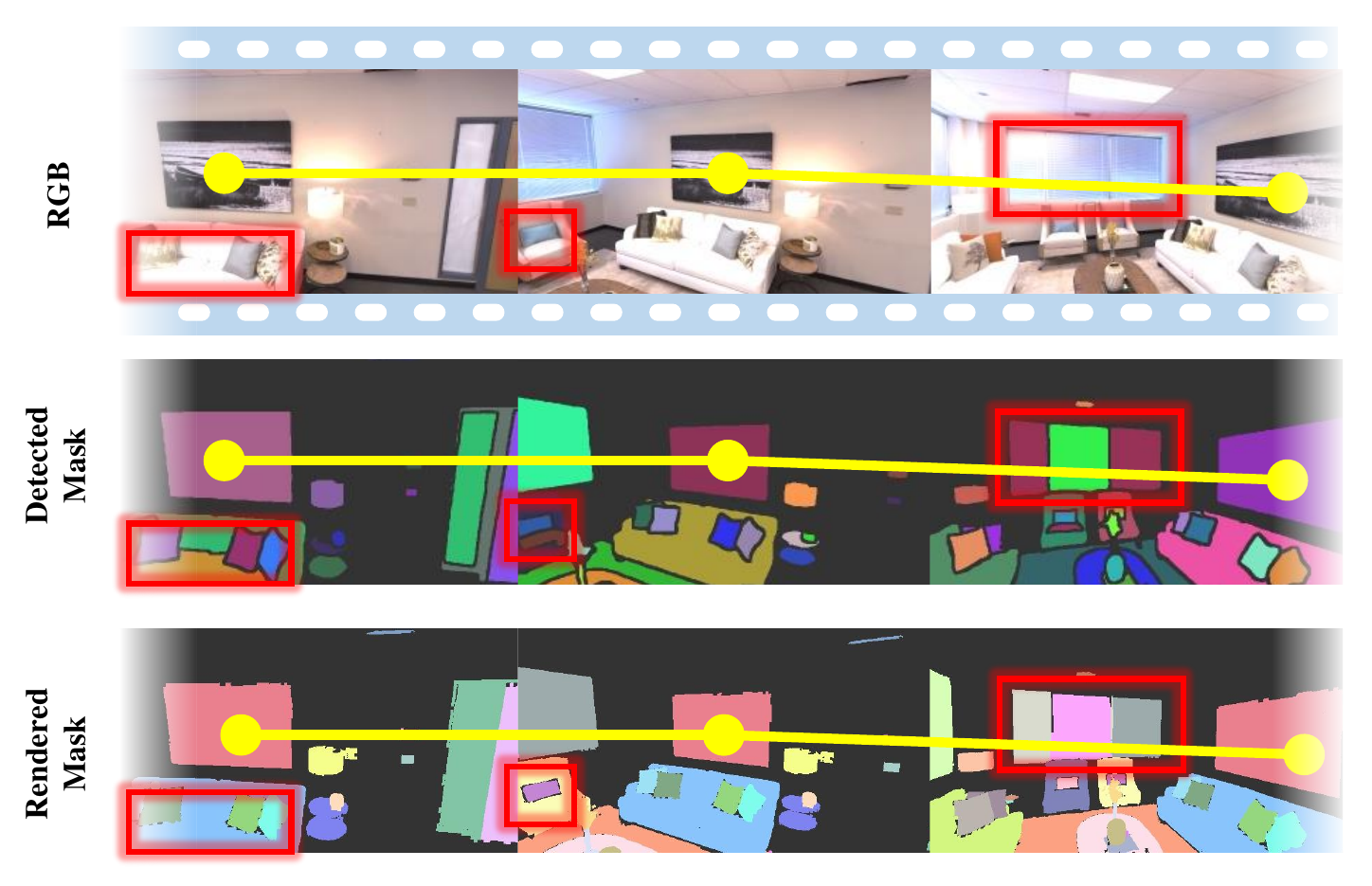}
	\caption{The comparison between rendered and detected masks highlights its effectiveness in associating instances across frames (yellow lines) and mitigating missing, under-, or over-segmentation (red boxes).}
	\label{asso} 
\end{figure}

Unlike traditional closed-set semantic frameworks, where semantics are inherently associated across frames, instance segmentation in open environments lacks mask inherent associations among frames, and the number of instances is unlimited. Furthermore, as shown in Fig. \ref{asso}, instance segmentation quality fluctuates due to challenging conditions including uneven lighting, oblique viewpoints, and object clutter. Thus, robustly associating new-frame masks $\{m_{t}^{k}\}$ with existing instances $\Gamma_{t-1}$ or initializing new ones becomes crucial.

Instance association involves mapping each segmented mask $m_{t}^{k}$ to the instance ID set $\Gamma_{t-1}$ from the current map $\mathcal{M}_{t-1}$:
\begin{equation}
\begin{aligned}
    \label{ins_ass}
    P({\mathcal{I}_{t}}\mid{\mathcal{M}_{t-1}},{\mathcal{Q}_{t}}) = P({\mathcal{I}_{t}}\mid\mathcal{V}_{t-1}, \mathcal{B}_{t-1}, \{m_{t}^{k}, f_{t}^{k}\},{{\textbf{D}}_{t}},{{\textbf{P}}_{t}})
\end{aligned}
\end{equation}

Given that mask overlaps are resolved in Subsection \ref{PIDS}, each mask's instance association can be treated independently. Under the conditional independence assumption, this reduces to solving separate association tasks for individual masks $m_{t}^{k}$:
\begin{equation}
\begin{aligned}
    \label{ins_ass_refine}
    \prod\limits_{i}{P({\mathcal{I}_{t}^{k}} \mid \mathcal{V}_{t-1}, \mathcal{B}_{t-1}, m_{t}^{k}, f_{t}^{k},{{\textbf{D}}_{t}},{{\textbf{P}}_{t}})}
\end{aligned}
\end{equation}
where $\mathcal{I}_{t}^{k}$ is the map instance ID associated with mask $m_{t}^{k}$.

We first project the mask $m_{t}^{k}$ into the voxel map $\mathcal{V}$ according to the depth image ${\textbf{D}}_{t}$ to get the corresponding associated voxel region $V_{m_{t}^{k}}$:
\begin{equation}
\begin{aligned}
    \label{pixel2voxel}
    V_{m_{t}^{k}}=\operatorname{Vox}\left( \left\{ {{\textbf{D}}_{t[u,v]}}{{\textbf{P}}_{t}}{{\textbf{K}}^{-1}}\cdot [u,v] \mid [u,v]\in m_{t}^{k} \right\} \right)
\end{aligned}
\end{equation}
where $[u,v]$ are the pixel coordinates within the mask $m_{t}^{k}$, $\textbf{K}$ is the camera internal parameter matrix, and $\operatorname{Vox}(\cdot)$ is the 3D point-to-voxel transformation. 
 
Consequently, our approach centers on computing the optimal instance assignment for the voxel set $V_{m_{t}^{k}}$. In cases where $V_{m_{t}^{k}}$ lacks prior instance information (indicating an initial observation), we perform two simultaneous operations: expanding the instance set $\Gamma_{t-1}$ by creating a new instance $\gamma_{new}$, and initializing its associated embedding in the codebook $\mathcal{B}_{t-1}$ using the extracted mask feature $f_{t}^{k}$. This approach efficiently handles previously unobserved instances while maintaining the integrity of the incremental mapping framework.

When $V_{m_{t}^{k}}$ contains existing instance information, the probability that the mask $m_{t}^{k}$ is associated with these instances must be determined. 
Using a Bayesian formulation, we convert this task into a Maximum Likelihood Estimation (MLE) problem:
\begin{equation}
\begin{aligned}
    \label{MLE}
    P(\mathcal{I}_{t}^{k} \mid V_{m_{t}^{k}})\propto P(\mathcal{I}_{t}^{k}) {P( V_{m_{t}^{k}}\mid \mathcal{I}_{t}^{k})}
\end{aligned}
\end{equation}
where $P(\mathcal{I}_{t}^{k})$ denotes the prior probability, assumed to be uniform across all instances $\Gamma_{t-1}$. 
Therefore, the goal is to solve for $\mathcal{I}_{t}^{k}$ such that the likelihood of all voxels $v^{j}_{m_{t}^{k}} \in V_{m_{t}^{k}}$ having the current instance  $\theta^{j}_{m_{t}^{k}}$ is maximized.

\begin{algorithm}[t]
\caption{Instance Association Algorithm}
\label{algo:instance_association}
\KwIn{
  Segmented masks $\{m_{t}^{k}, f_{t}^{k}\}$; \\
  \quad\quad\quad \  Depth image $\textbf{D}_t$, camera pose $\textbf{P}_t$; \\
  \quad\quad\quad \ Current voxel map $\mathcal{V}_{t-1}$ and codebook $\mathcal{B}_{t-1}$
}
\KwOut{
  Instance IDs $\{\mathcal{I}_t^k\}$ for each mask $\{m_t^k\}$
}

\ForEach{mask $m_t^k$}{
  Project pixels $[u,v] \in m_t^k$ to 3D using $\textbf{D}_t$ and $\textbf{P}_t$ \\
  Obtain corresponding voxel region $V_{m_t^k}$ via Eq.~\eqref{pixel2voxel} \\

  \uIf{$V_{m_t^k}$ has no prior instance}{
    Assign new instance ID $\gamma_{\text{new}}$, add to $\Gamma_{t-1}$ \\
    Initialize $\mathcal{B}_{t-1}[\gamma_{\text{new}}] \leftarrow f_t^k$
  }
  \Else{
    \ForEach{existing instance $\gamma \in \Gamma_{t-1}$}{
      Compute geometric similarity $S^{geo}_{\mathcal{I}_t^k=\gamma}$ via Eq.~\eqref{geo_sim} \\
      Compute embedding similarity $S^{emb}_{\mathcal{I}_t^k=\gamma}$ via Eq.~\eqref{fea_sim} \\
      Fuse to get final score $A_{\mathcal{I}_t^k=\gamma}$
    }
    \If{$\max_{\gamma} A_{\mathcal{I}_t^k=\gamma} > \xi$}{
      Assign $\mathcal{I}_t^k = \arg\max_\gamma A_{\mathcal{I}_t^k=\gamma}$
    }
    \Else{
       Assign new instance ID $\gamma_{\text{new}}$, add to $\Gamma_{t-1}$ \\
    Initialize $\mathcal{B}_{t-1}[\gamma_{\text{new}}] \leftarrow f_t^k$
    }
  }
}
Update instance set: $\Gamma_t \leftarrow \Gamma_{t-1}$; \\
\Return $\{\mathcal{I}_t^k\}$
\end{algorithm}

Since the voxels are relatively independent, a rigorous approach would involve multiplying the likelihood probabilities of each voxel. However, this introduces high computational complexity and is prone to arithmetic underflow. To mitigate this, we simplify the MLE process by accumulating evidence. By applying the law of large numbers, if the number of voxels $V_{m_{t}^{k}}$ is sufficiently large and the observed probability distribution for each voxel is accurate, averaging these likelihood probabilities provides a reliable estimate of the geometric similarity $S^{geo}_{\mathcal{I}_{t}^{k}=\gamma}$ for mask $m_t^k$ and instance $\gamma$:
\begin{equation}
\begin{aligned}
    \label{geo_sim}
    S^{geo}_{\mathcal{I}_{t}^{k}=\gamma} = \underset{j}{\mathbb{E}}(\{\theta^{j}_{m_{t}^{k}}[\gamma]\})
\end{aligned}
\end{equation}

The geometric similarity metric $S^{geo}_{\mathcal{I}_{t}^{k}=\gamma}$ operates on the principle that spatially co-located masks are more likely to correspond to the same object.
However, this assumption may break down for adjacent objects with tight spatial couplings, such as books on a table, especially in voxelized configurations. To address this limitation, we augment the geometric measure with embedding cosine similarity $S^{emb}_{\mathcal{I}_{t}^{k}=\gamma}$, enabling robust instance association through semantic-level feature matching:
\begin{equation}
\begin{aligned}
    \label{fea_sim}
    S^{emb}_{\mathcal{I}_{t}^{k}=\gamma} = \operatorname{Cos\_Sim}(f^{\gamma}_{t-1}, f_{t}^{k})
\end{aligned}
\end{equation}
where $f^{\gamma}_{t-1}$ is the embedding for instance $\gamma$ in codebook $\mathcal{B}_{t-1}$.

The final association probability $A_{\mathcal{I}_{t}^{k}=\gamma}$ is obtained by the weighted fusion of the two similarities $S^{geo}_{\mathcal{I}_{t}^{k}=\gamma}$ and $S^{emb}_{\mathcal{I}_{t}^{k}=\gamma}$. 
Instance association occurs if the maximum probability $\underset{\gamma}{\operatorname{Max}}(A_{\mathcal{I}_{t}^{k}=\gamma})$ exceeds the instance fusion threshold $\xi$; otherwise, a new instance is initialized to $\Gamma_{t-1}$. The complete instance association algorithm is detailed in Alg. \ref{algo:instance_association}.

\subsection{Live Map Evolution}

The live map evolution jointly updates both the voxel representation $\mathcal{V}_{t-1}$ and the embedding codebook $\mathcal{B}_{t-1}$. The voxel update process employs probabilistic local fusion to maintain robustness against segmentation noise, while the codebook progressively integrates multi-view instance perceptions through feature aggregation.

\textbf{Voxel Updating:}
For voxel updating, we propose the Instance Counting Sensor Model (ICSM), inspired by the semantic counting sensor model in \cite{s-bki}.  As previously mentioned, each voxel $v^j$ in $\mathcal{V}$ stores a probabilistic instance ID tuple $\theta^j=\{\theta^{j,\gamma}\mid \gamma \in \Gamma \}$, where $\theta^{j,\gamma}>0$ and $\sum\nolimits_{\gamma \in \Gamma }{{{\theta }^{j,\gamma }}}=1$.
During the instance association phase, we process observation data $\{\mathcal{V}_{m^k_t}, \mathcal{I}^k_t\}$, simplified as $\{(v_t^j,y_t^j)\}$, where $v_t^j$ denotes the observed voxels in the current frame, and $y_t^j$ is a one-hot-encoded measurement tuple representing the observed instance ID.

Voxel updating is essentially a Maximum A Posteriori estimation (MAP) task:
\begin{equation}
\begin{aligned}
    \label{map}
    p(\theta _{t}^{j} \mid y_{t}^{j})\propto p(y_{t}^{j} \mid \theta _{t}^{j})p(\theta _{t}^{j})
\end{aligned}
\end{equation}
where the prior probability $p(\theta _{t}^{j})$ is modeled as equivalent to the previous timestep's posterior probability $p(\theta_{t-1}^{j})$. 
The likelihood $p(y_{t}^{j} \mid \theta_{t}^{j})$ follows a Categorical distribution, representing the probability distribution over possible labels $y_t^{j,\gamma}$ for voxel $v^j$ given the current observation:
\begin{equation}
\begin{aligned}
    \label{likelihood}
    p(y_{t}^{j} \mid \theta _{t}^{j})={{\prod\limits_{\gamma \in \Gamma }{(\theta _{t}^{j,\gamma })}}^{y_{t}^{j,\gamma }}}
\end{aligned}
\end{equation}

When applying the Dirichlet distribution, the conjugate prior for the Categorical distribution, to the prior probability, the posterior probability remains of the same distribution type:
\begin{equation}
\begin{aligned}
    \label{prior}
    p(\theta _{t-1}^{j})\propto {{\prod\limits_{\gamma \in \Gamma }{(\theta _{t}^{j,\gamma })}}^{\alpha _{t-1}^{j,\gamma }-1}}
\end{aligned}
\end{equation}
\begin{equation}
\begin{aligned}
    \label{posterior}
    p(\theta _{t}^{j} \mid y_{t}^{j})\propto {{\prod\limits_{\gamma \in \Gamma }{(\theta _{t}^{j,\gamma })}}^{\alpha _{t}^{j,\gamma }-1}}
\end{aligned}
\end{equation}
where $\alpha _{t-1}^{j}$ and $\alpha _{t}^{j}$ are the concentration parameters of the prior and posterior, respectively.

Substituting Eq. \eqref{likelihood}, Eq. \eqref{prior} and Eq. \eqref{posterior} into \eqref{map}, we can deduce that:
\begin{equation}
\begin{aligned}
    \label{count}
    \alpha_{t}^{j,\gamma }=\alpha_{t-1}^{j,\gamma }+y_{t}^{j,\gamma }
\end{aligned}
\end{equation}
Since $\alpha_{t}^{j,\gamma }$ counts the number of times voxel $v^j$ is associated with instance label $\gamma$, we refer to this model as the instance counting sensor model. 
Given $\alpha_{t}^{j}$, the probabilistic instance tuple of the voxel $v^j$ is the closed-form expected value of the posterior Dirichlet \cite{see-csom}:
\begin{equation}
\begin{aligned}
    \label{instance_tuple}
    \theta _{t}^{j,\gamma }=\frac{\alpha _{t}^{j,\gamma }}{\sum\limits_{\gamma' \in \Gamma }{\alpha _{t}^{j,\gamma'}}}
\end{aligned}
\end{equation}

The Dirichlet concentration parameters $\alpha _{t}^{j}$ act as pseudocounts that accumulate evidence over time. This formulation provides three key properties: (i) the update rule Eq. \eqref{count} becomes a simple counting operation, (ii) the expected value Eq. \eqref{instance_tuple} yields interpretable probabilities, and (iii) the variance naturally decreases with accumulating evidence.

\begin{algorithm}[t]
\caption{Live Map Evolution}
\label{algo:map_evo}
\KwIn{Observation set $\{(v^j_t, y^j_t)\}$; \\
\quad\quad\quad Mask set $\{(m^k_t, \mathcal{I}^k_t, f^k_t)\}$;\\
\quad\quad\quad Previous voxel map $\mathcal{V}_{t-1}$, Codebook $\mathcal{B}_{t-1}$}
\KwOut{Updated voxel map $\mathcal{V}_t$, Codebook $\mathcal{B}_t$}
% \Statex \Comment{This is a full-line comment.}
\ForEach{voxel observation $(v^j_t, y^j_t)$}{
    $\alpha^{j}_{t} \gets \alpha^{j}_{t-1} + y^j_t$\\
    $\theta^{j}_{t,\gamma} \gets \alpha^{j}_{t,\gamma} / \sum_{\gamma'} \alpha^{j}_{t,\gamma'}$
}
\ForEach{mask $m^k_t$ with instance ID $\mathcal{I}^k_t$ and feature $f^k_t$}{
    Compute credibiliy $w^k_t$ via Eq.~\eqref{credibility} and Eq.~\eqref{vis_ratio} \\
    $W_t^{\mathcal{I}^k_t} \gets W_{t-1}^{\mathcal{I}^k_t} + w^k_t$\\
    $f_t^{\mathcal{I}^k_t} \gets \left(W_{t-1}^{\mathcal{I}^k_t} f_{t-1}^{\mathcal{I}^k_t} + w^k_t f^k_t\right) / W_t^{\mathcal{I}^k_t}$
}
Update voxel map: $\mathcal{V}_t \leftarrow \mathcal{V}_{t-1}$ \\
Update codebook: $\mathcal{B}_t \leftarrow \mathcal{B}_{t-1}$ \\
\Return{$\mathcal{V}_t$, $\mathcal{B}_t$}
\end{algorithm}

\textbf{Codebook Updating:}
For the codebook updating, we use a weighted fusion strategy. For each mask $m_{t}^{k}$, the associated instance ID $\mathcal{I}_{t}^{k}$ is obtained in the instance association step. The credibility $w^t_i$ of its current frame observation features $f^{k}_{t}$ is evaluated by combining the association probability $A_{\mathcal{I}_{t}^{k}}$ and the visibility ratio $R^k_{t}$:
\begin{equation}
\begin{aligned}
    \label{credibility}
    w^t_k = A_{\mathcal{I}_{t}^{k}} \cdot R^k_{t}
\end{aligned}
\end{equation}
\begin{equation}
\begin{aligned}
    \label{vis_ratio}
    {R^k_{t}}=\frac{\left| {{V}_{m_{t}^{k}}} \right|}{\left| \left\{ \underset{\gamma }{\mathop{\arg \max }}\,\left( \theta _{t-1}^{j}[\gamma ] \right)  =\mathcal{I}_{t}^{k} \right\}_j \right|}
\end{aligned}
\end{equation}
The visibility ratio $R_{k}$ represents the proportion of the instance’s size observed by the current mask $m_{t}^{k}$ relative to the total size of the instance. This helps prevent mask features with poor viewing (e.g., only a corner of a couch is visible) from contaminating the global codebook.
Based on this, the updating of the codebook can be derived as:
\begin{equation}
\begin{aligned}
    \label{codebook_updating}
    f_{t}^{\mathcal{I}_{t}^{k}}=\left({W_{t-1}^{\mathcal{I}_{t}^{k}}f_{t-1}^{\mathcal{I}_{t}^{k}}+w_{t}^{i}f_{t}^{k}}\right)/{W_{t}^{\mathcal{I}_{t}^{k}}}
\end{aligned}
\end{equation}
\begin{equation}
\begin{aligned}
    \label{codebook_weight}
    W_{t}^{\mathcal{I}_{t}^{k}}=W_{t-1}^{\mathcal{I}_{t}^{k}}+w_{t}^{k}
\end{aligned}
\end{equation}
where $f_{t}^{\mathcal{I}_{t}^{k}}$ and $W_{t}^{\mathcal{I}_{t}^{k}}$ are the updated features and weights of instance $\mathcal{I}_{t}^{k}$ in embedding codebook $\mathcal{B}_{t}$, respectively. The complete instance association algorithm is detailed in Alg. \ref{algo:map_evo}.

\section{Motion-Robust 3DGS Incremental Reconstructor} \label{MR}

The preceding sections have robustly achieved incremental scene understanding through voxel-based representations. While voxels enable locally fused updates, their inherent resolution limitations fundamentally restrict the ability to reconstruct high-fidelity geometric and photometric details of the scene. To address these limitations, we now introduce an incremental representation based on 3DGS \cite{3DGS}, which reconstructs scene observations using more flexible and differentiable Gaussian primitives. This approach overcomes the rigid structural constraints of voxel grids while maintaining the incremental update capabilities essential for scene reconstruction.

\subsection{Preliminary for 3DGS}
3DGS \cite{3DGS} is composed of 3D explicit Gaussian ellipsoids $\{g^i\}$ parameterized by
\begin{enumerate}
    \item \textit{3D center:} \quad ${{\mu}^i} = ({{x}^i},{{y}^i},{{z}^i})\in {{\mathbb{R}}^{3}}$
    \item \textit{3D rotation:} \quad ${{{{q}}}^i} = (q{{w}^i},q{{x}^i},q{{y}^i},q{{z}^i})\in {{\mathbb{R}}^{4}}$
    \item \textit{RGB color:} \quad ${c}^i = (r^i,g^i,b^i)\in {{\mathbb{R}}^{3}}$
    \item \textit{3D size (scaling factor):} \quad $s^i = (sx^i,sy^i,sz^i)\in {{\mathbb{R}}^{3}}$
    \item \textit{Opacity logit:} \quad $o^i\in [0,1]$
\end{enumerate}

Each Gaussian occupies a spatial region, and RGB values ${c}^i$ can alternatively be represented as view-dependent spherical harmonics coefficients. For an arbitrary query point $x \in {{\mathbb{R}}^{3}}$, the i-th Gaussian $g^i$'s contribution is weighted by its opacity $o^i$, computed via the standard Gaussian function:
\begin{equation}
\begin{aligned}
    \label{contribution}
    	{{\eta }^{i}}(x)={{o}^{i}}\exp \left( -\frac{1}{2}(x-{{\mu }^{i}}){{\Sigma }^{i}}^{T}(x-{{\mu }^{i}}) \right)
\end{aligned}
\end{equation}
where ${{\Sigma }^{i}} \in {{\mathbb{R}}^{3 \times 3}}$ is the covariance matrix calculated by combining the scaling component $S^i=\operatorname{darg}([\begin{matrix}
   s{{x}^{i}} & s{{y}^{i}} & s{{z}^{i}}  \\
\end{matrix}]$ and the rotation component $R^i=\operatorname{q2R}([\begin{matrix}
   qw^i & qx^i & qy^i & qz^i  \\
\end{matrix}]$.

Due to their formulation, Gaussians have theoretically infinite support, which leads to gradients having a global influence during optimization. To fit observed scenes, 3DGS differentiably renders Gaussians $\{g^i\}$ into images. During rendering, Gaussian primitives are depth-sorted and projected via a 2D version of Eq. \eqref{contribution}. The 2D density of $g^i$ is denoted as $\alpha^i$.
The color $\hat{\textbf{C}}(p)$ and depth $\hat{\textbf{D}}(p)$ of each pixel $p$ are then computed using alpha-blending:
\begin{equation}
\begin{aligned}
    \label{color_render}
    \hat{\textbf{C}}(p)=\sum\limits_{i}{{{c}^{i}}{{\alpha }^{i}}\prod\limits_{j=1}^{i-1}{(1-{{\alpha }^{j}})}}
\end{aligned}
\end{equation}
\begin{equation}
\begin{aligned}
    \label{depth_render}
    \hat{\textbf{D}}(p)=\sum\limits_{i}{{{h}^{i}}{{\alpha }^{i}}\prod\limits_{j=1}^{i-1}{(1-{{\alpha }^{j}})}}
\end{aligned}
\end{equation}
where ${h}^{i}$ represent the depth of the Gaussian primitive $g^i$ in given camera pose.

\subsection{Voxel-Based Frame Initialization}

\begin{figure}[!t]\centering
	\includegraphics[width=7.8cm]{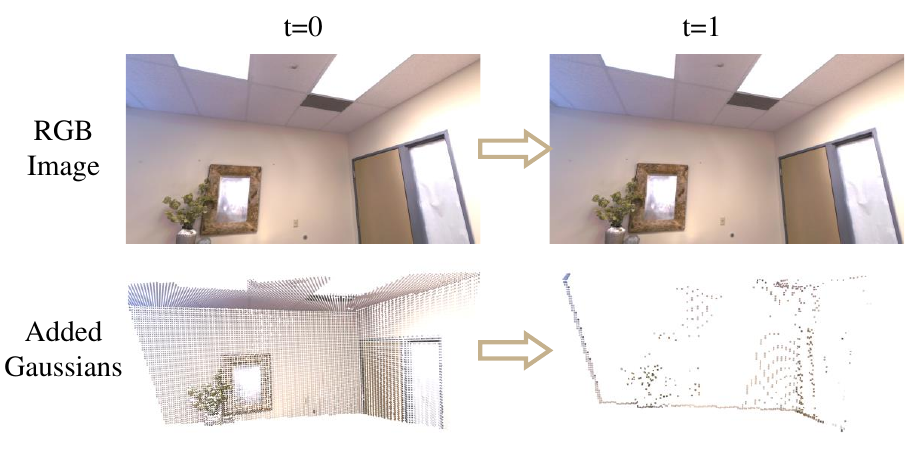}
	\caption{OmniMap initializes the Gaussian by distinguishing the newly added voxels, providing an efficient yet effective coverage of newly observed regions. After initialization, only minimal Gaussian additions are required for subsequent updates.}
	\label{add_gaussians} 
\end{figure}

\textbf{Add New Gaussians:}
Unlike the original 3DGS paradigm, online incremental 3DGS requires the continuous addition of new Gaussians to adapt to newly observed regions, which are unpredictable a priori. Some existing methods \cite{hi-slam2} adopt a brute-force incremental approach, back-projecting each keyframe to initialize Gaussian centers from point clouds. However, this leads to uneven scene distribution, where frequently observed regions accumulate excessive Gaussian density compared to rarely viewed areas. To mitigate this, alternative methods \cite{Gaussian-slam} use rendered alpha-blending values to identify newly observed pixels for back-projection. Yet, this approach remains vulnerable to artifacts from large-scale Gaussians in novel views and lacks 3D physical consistency.

Thanks to TSDF-Fusion's standardized voxel representation, OmniMap achieves regular and efficient Gaussian initialization. Specifically, OmniMap identifies candidate voxels for Gaussian addition through a simple filter operation on the counting values from Eq. \eqref{count}:
\begin{equation}
\begin{aligned}
    	\mathcal{V}_{t}^\text{new}=\left\{ {{v}^{j}}|\alpha _{t-1}^{j}=0\text{ and }\alpha _{t}^{j}>0 \right\}
\end{aligned}
\end{equation}
$\mathcal{V}_{t}^\text{new}$ is the new voxel centers, from where new Gaussians $\mathcal{G}_{t}^\text{new}$ are initialized.
The colors of $\mathcal{G}_{t}^\text{new}$ are set as the voxel colors from TSDF-Fusion. Their spatial extent $s$ is scaled to 0.2 times the voxel size, with an initial opacity $o$ of 0.5.
Fig. \ref{add_gaussians} demonstrates OmniMap's Gaussian addition process in a representative case.

\textbf{Keyframe Discrimination:}
Similar to other 3DGS-based SLAM systems, OmniMap maintains a keyframe buffer to prevent catastrophic forgetting during incremental optimization. To ensure effective scene coverage, keyframe selection prioritizes maximally divergent viewpoints, thereby enhancing the quality of multi-view reconstruction.
Our approach quantifies view dissimilarity by computing the ratio of unregistered to observed regions at the voxel level. 
To do this, we introduce a boolean variable $r(v^j_t)$ for each voxel $v^j$, initialized as unregistered and updated to registered upon keyframe inclusion. When the unregistered region ratio $\tau_t$ exceeds a predefined threshold, the current frame is added to the keyframe buffer:
\begin{equation}
\begin{aligned}
    {{\tau }_{t}}=\frac{1}{n}\sum\limits_{j=1}^{n}{\mathbb{I}(r(v_{t}^{j}))}
\end{aligned}
\end{equation}
where $n$ denotes the total number of observable voxels at time $t$ with pose $\textbf{P}_t$, and $\mathbb{I}$ is the indicator function.
In OmniMap, keyframes are selected not only based on the unregistered ratio but also when more than $N_{key}$ consecutive frames are observed without inserting a new keyframe. Such temporal density control avoids under-sampling of the scene, which is critical for dense reconstruction accuracy.

\begin{figure}[!t]\centering
	\includegraphics[width=8.5cm]{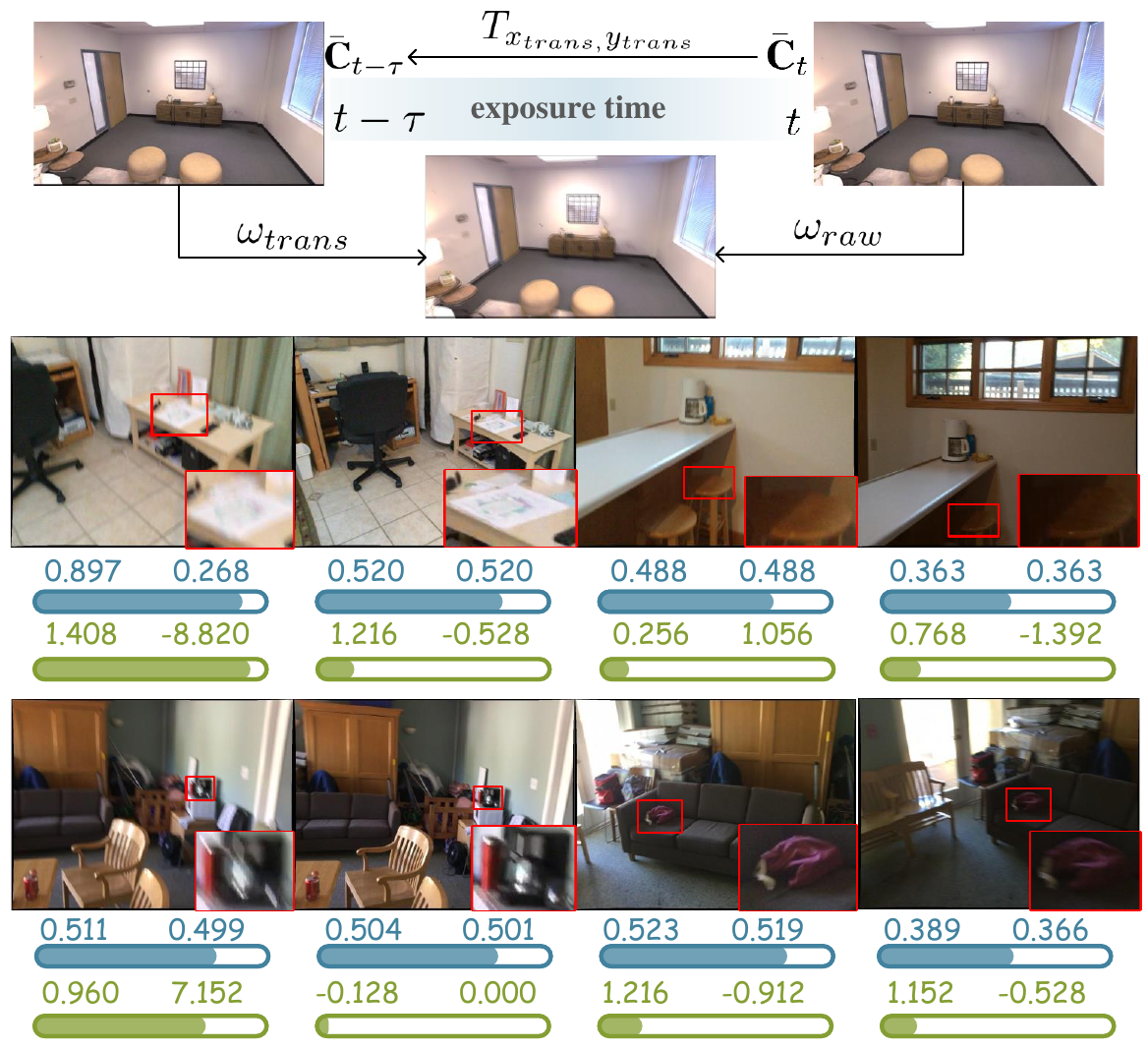}
	\caption{Schematic of camera modeling. The numbers under images correspond to the optimized parameter set $[{\omega }_{raw}, {\omega }_{trans},  {{x}_{trans}},{{{y}_{trans}}}]$. 
    Quantitative analysis reveals: (1) high-brightness images show elevated ${\omega }_{raw}+{\omega }_{trans}$ (blue bar) relative to low-brightness conditions, and (2) blurred images exhibit consistently higher $|{x}_{trans}|+|{y}_{trans}|$ (green bar) compared to sharp counterparts. These correlations validate OmniMap's effectiveness in camera modeling.
    }
	\label{camera_modeling} 
\end{figure}

\subsection{Camera Modeling}

In image-based multi-view dense mapping, the camera model plays a critical role in determining reconstruction quality. The fundamental camera parameters consist of intrinsic and extrinsic parameters. During incremental 3DGS mapping, intrinsic parameters $\textbf{K}$ remain fixed (as they are camera-specific), while extrinsic parameters are derived from the camera pose $\mathcal{P}$, typically estimated by visual SLAM systems. However, two pervasive challenges extend beyond these basic parameters: motion blur and exposure inconsistency.
Motion blur arises from relative camera-scene motion during exposure, resulting in smeared trajectories in captured images. Exposure inconsistency occurs when varying lighting conditions or view-dependent reflectivity cause divergent brightness levels across different viewpoints, more precisely, commonly perceived as variations in light intensity.

The physical process of camera motion blur can be mathematically modeled as a temporal integration of latent sharp images along the camera's trajectory during exposure:
\begin{equation}
\begin{aligned}
    \label{cam_blur}
    {{\textbf{C}}_{t}}=\phi \int_{t-\tau}^{t}{{\tilde{\textbf{C}}}}\text{dt}
\end{aligned}
\end{equation}
where ${{\textbf{C}}_{t}}$ denotes the real captured motion-blurred image at time $t$, $\phi$ serves as a normalization factor, $\tilde{\textbf{C}}$ is the virtual latent sharp image captured at timestamp $[t-\tau, t]$ within the exposure time.

To simplify the solution, OmniMap discretely approximates the model in Eq. \eqref{cam_blur} as a weighted average of the exposure start and end moments:
\begin{equation}
\begin{aligned}
    \label{cam_blur_sim}
   {\textbf{C}_{t}} \approx {{\omega }_{trans}}{{\bar{\textbf{C}}}_{t-\tau }}+{{\omega }_{raw}}{{\bar{\textbf{C}}}_{t}}
\end{aligned}
\end{equation}
where ${\omega }_{trans}$ and ${\omega }_{raw}$ are two control parameters in the range $[0,1]$. With this setting, the camera model in Eq. \eqref{cam_blur_sim} can effectively simulate exposure intensity: when ${\omega }_{trans} + {\omega }_{raw} > 1$, the image appears brighter; and when ${\omega }_{trans} + {\omega }_{raw} < 1$, the image appears darker.

${{\bar{\textbf{C}}}_{t-\tau }}$ represents a simulated sharp image at a discrete timestep within the current frame's exposure period. For small time intervals, minor camera translation can be approximated as a global image shift, neglecting depth variations and rotational effects. We model ${{\bar{\textbf{C}}}_{t-\tau }}$ using the image transformation parameters $x_{trans}$ and $y_{trans}$:
\begin{equation}
\begin{aligned}
   {{\bar{\textbf{C}}}_{t-\tau }} \approx {{T}_{{{x}_{trans,}}{{{y}_{trans}}}}}\left({{\bar{\textbf{C}}}_{t}}\right)
\end{aligned}
\end{equation}
where ${{T}_{{{x}_{trans,}}{{{y}_{trans}}}}}(\cdot)$ denotes a translation transformation $[{{x}_{trans,}}{{{y}_{trans}}}]$ for the image. Thus the rendered image is modeled as four parameters:
\begin{equation}
\begin{aligned}
    \label{deblur}
   {\textbf{C}_{t}} \approx  {{\omega }_{trans}}{{T}_{{{x}_{trans,}}{{{y}_{trans}}}}}\left({{\bar{\textbf{C}}}_{t}}\right)+{{\omega }_{raw}}{{\bar{\textbf{C}}}_{t}}
\end{aligned}
\end{equation}

Similar to the Gaussian primitives, these four parameters $[{\omega }_{raw}, {\omega }_{trans},  {{x}_{trans}}, {{{y}_{trans}}}]$ are jointly optimized within the 3DGS pipeline, which are initialized to $[0.5,0.5,0,0]$.
The incorporation of camera modeling significantly enhances scene reconstruction quality while mitigating the impact of motion blur and exposure variations on Gaussian field optimization. 
As shown in Fig. \ref{camera_modeling}, our camera modeling approach is presented alongside experimental results that verify the effectiveness of the proposed blur and exposure modeling strategy.

\subsection{Loss Function}

To ensure Gaussian primitives faithfully reconstruct the real-world scene and the camera model accurately represent the imaging process, OmniMap employs multiple loss functions to optimize these parameters.

\textbf{RGB Loss:} For keyframe with time $t'$, we employ Eq. \eqref{color_render} to render the raw color image $\hat{\textbf{C}}_{t'}$. To ensure sharp reconstruction, we transform the rendered image $\hat{\textbf{C}}_{t'}$ using Eq. \eqref{deblur} with the corresponding camera model parameters for current keyframe, producing $\hat{\textbf{C}}^{cam}_{t'}$, which is then supervised against the real observed image $\textbf{C}_{t'}$. This supervision is implemented through a combination of L1 loss $\mathcal{L}_{rgb}$ and SSIM loss $\mathcal{L}_{ssim}$:
\begin{equation}
\begin{aligned}
   \mathcal{L}_{rgb} = {{\left\| \hat{\textbf{C}}^{cam}_{t'}-\textbf{C}_{t'} \right\|}_{1}}
\end{aligned}
\end{equation}
\begin{equation}
\begin{aligned}
   \mathcal{L}_{ssim} = \operatorname{SSIM}(\hat{\textbf{C}}^{cam}_{t'},\textbf{C}_{t'})
\end{aligned}
\end{equation}

\begin{figure}[!t]\centering
	\includegraphics[width=8.2cm]{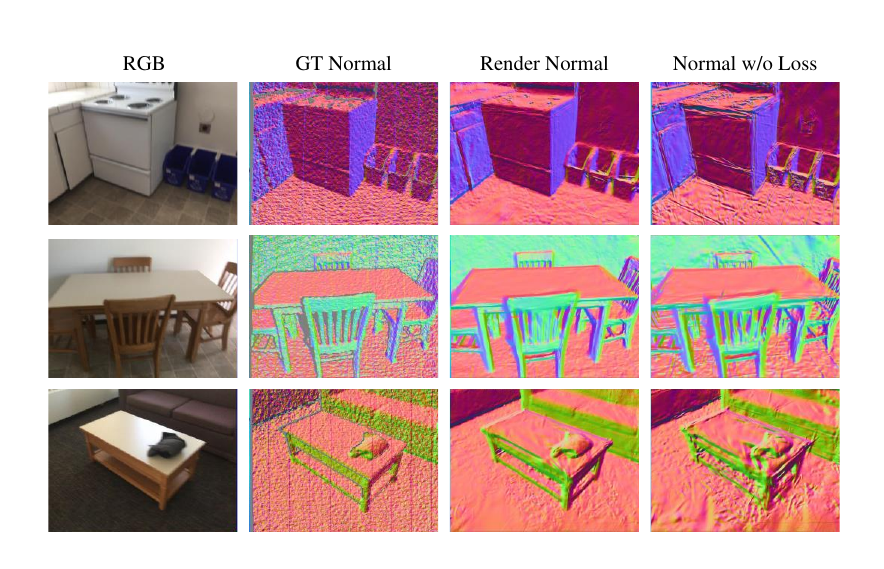}
	\caption{OmniMap's rendering produces smooth and geometrically consistent normal images despite being trained with noisy supervision, demonstrating significantly better reconstruction quality than approaches without $\mathcal{L}_{normal}$.}
	\label{normal} 
\end{figure}

\textbf{Depth Loss:} Relying solely on RGB supervision may lead to overfitting, significantly degrading novel view synthesis quality. To address this, we incorporate depth supervision to guide Gaussian primitives in accurately conforming to object surfaces, thereby improving geometric fidelity. The depth supervision is implemented using an L1 loss $\mathcal{L}_{depth}$ between the rendered depth $\hat{\textbf{D}}_{t'}$ and ground truth depth $\textbf{D}_{t'}$:
\begin{equation}
\begin{aligned}
   \mathcal{L}_{depth} = {{\left\| \hat{\textbf{D}}_{t'}-\textbf{D}_{t'} \right\|}_{1}}
\end{aligned}
\end{equation}

\textbf{Normal Loss:} Compared to depth, normals exhibit stronger local geometric sensitivity and multi-view consistency. Departing from monocular estimation approaches \cite{hi-slam2}, we directly compute normal images from depth images. To enforce geometric accuracy, we employ a cosine similarity loss $\mathcal{L}_{normal}$ between the rendered normal image $\hat{\textbf{N}}_{t'}$ (derived from $\hat{\textbf{D}}_{t'}$ ) and the ground truth normal image $\textbf{N}_{t'}$ (derived from $\textbf{D}_{t'}$:
\begin{equation}
\begin{aligned}
   \mathcal{L}_{normal} = 1-\left\langle \hat{\textbf{N}}_{t'},\textbf{N}_{t'} \right\rangle
\end{aligned}
\end{equation}
where $\left\langle \cdot \right\rangle$ denotes the inner product operation.
As illustrated in Fig. \ref{normal}, the optimized rendered normal image $\hat{\textbf{N}}_{t'}$ demonstrates superior geometric fidelity after optimization, despite noise-induced artifacts in the ground truth normal image $\textbf{N}_{t'}$ derived from depth measurements. This improvement is particularly evident when compared to reconstruction results without normal supervision, where fine geometric details are significantly harder to recover.

By integrating the aforementioned losses through a weighted combination, OmniMap progressively optimizes both the Gaussian primitives and camera parameters. This optimization framework enables incremental reconstruction of scene appearance and geometry, facilitated by the dynamically expanding keyframe collection.

\begin{table}[!t]
\scriptsize
\centering
\caption{Rendering quality evaluations on Replica dataset.}
\label{replica_render_table}
\renewcommand\arraystretch{1.1}
\setlength{\tabcolsep}{0.8mm}
{
\begin{tabular}{ccccccccccc}
\toprule
\textbf{Method} & \textbf{Metric} & \textbf{ro-0} & \textbf{ro-1} & \textbf{ro-2} & \textbf{of-0} & \textbf{of-1} & \textbf{of-2} & \textbf{of-3} & \textbf{of-4} & \textbf{Avg.} \\
 \midrule
 & \textbf{PSNR↑} & 26.14 & 30.64 & 31.56 & 37.32 & 37.43 & 30.65 & 30.54 & 33.37 & 32.21 \\
 & \textbf{SSIM ↑} & 0.905 & 0.947 & 0.956 & \cellcolor[HTML]{FFFFB2}0.985 & \cellcolor[HTML]{FFD8B2}0.986 & 0.968 & \cellcolor[HTML]{FFFFB2}0.970 & \cellcolor[HTML]{FFFFB2}0.974 & 0.961 \\
\multirow{-3}{*}{\textbf{\begin{tabular}[c]{@{}c@{}}RTG-\\ SLAM\end{tabular}}} & \textbf{LPIPS ↓} & 0.281 & 0.194 & 0.188 & 0.094 & 0.112 & 0.180 & 0.167 & 0.161 & 0.172 \\
 \midrule
 & \textbf{PSNR↑} & 32.34 & 33.08 & 35.08 & \cellcolor[HTML]{FFD8B2}41.16 & \cellcolor[HTML]{FFB2B2}42.73 & 34.28 & \cellcolor[HTML]{FFFFB2}35.76 & 36.55 & 36.37 \\
 & \textbf{SSIM ↑} & 0.913 & 0.921 & 0.953 & 0.980 & 0.982 & 0.951 & 0.963 & 0.967 & 0.954 \\
\multirow{-3}{*}{\textbf{\begin{tabular}[c]{@{}c@{}}Gaussian-\\ SLAM\end{tabular}}} & \textbf{LPIPS ↓} & 0.100 & 0.107 & 0.063 & \cellcolor[HTML]{FFFFB2}0.025 & \cellcolor[HTML]{FFFFB2}0.026 & 0.079 & 0.048 & 0.056 & 0.063 \\
 \midrule
 & \textbf{PSNR↑} & 33.20 & 34.80 & 36.41 & 39.52 & 40.73 & \cellcolor[HTML]{FFD8B2}35.91 & \cellcolor[HTML]{FFD8B2}36.04 & 33.23 & 36.23 \\
 & \textbf{SSIM ↑} & 0.939 & 0.942 & 0.964 & 0.969 & 0.976 & 0.966 & 0.968 & 0.938 & 0.958 \\
\multirow{-3}{*}{\textbf{MonoGS}} & \textbf{LPIPS ↓} & 0.073 & 0.060 & \cellcolor[HTML]{FFFFB2}0.041 & 0.034 & 0.031 & \cellcolor[HTML]{FFD8B2}0.035 & \cellcolor[HTML]{FFD8B2}0.028 & 0.077 & 0.047 \\
 \midrule
 & \textbf{PSNR↑} & \cellcolor[HTML]{FFFFB2}33.42 & 33.86 & 35.32 & 39.57 & 40.56 & 32.15 & 30.88 & 32.71 & 34.81 \\
 & \textbf{SSIM ↑} & \cellcolor[HTML]{FFB2B2}0.982 & \cellcolor[HTML]{FFD8B2}0.974 & \cellcolor[HTML]{FFB2B2}0.988 & \cellcolor[HTML]{FFD8B2}0.989 & \cellcolor[HTML]{FFB2B2}0.987 & \cellcolor[HTML]{FFD8B2}0.971 & 0.961 & 0.958 & \cellcolor[HTML]{FFD8B2}0.976 \\
\multirow{-3}{*}{\textbf{SplaTAM}} & \textbf{LPIPS ↓} & 0.059 & 0.067 & 0.045 & 0.040 & 0.052 & 0.074 & 0.086 & 0.105 & 0.066 \\
 \midrule
 & \textbf{PSNR↑} & 29.797 & 31.81 & 34.83 & 38.04 & 38.05 & 31.33 & 32.58 & 36.28 & 34.70 \\
 & \textbf{SSIM ↑} & 0.929 & 0.948 & 0.965 & 0.970 & 0.957 & 0.953 & 0.960 & 0.972 & 0.961 \\
\multirow{-3}{*}{\textbf{GSFusion}} & \textbf{LPIPS ↓} & 0.063 & 0.076 & 0.055 & 0.043 & 0.079 & 0.086 & 0.042 & 0.044 & 0.061 \\
 \midrule
 & \textbf{PSNR↑} & \cellcolor[HTML]{FFD8B2}35.24 & \cellcolor[HTML]{FFD8B2}36.24 & \cellcolor[HTML]{FFFFB2}36.94 & \cellcolor[HTML]{FFFFB2}41.06 & \cellcolor[HTML]{FFFFB2}41.15 & 34.72 & 34.94 & \cellcolor[HTML]{FFD8B2}39.52 & \cellcolor[HTML]{FFD8B2}37.47 \\
 & \textbf{SSIM ↑} & \cellcolor[HTML]{FFFFB2}0.962 & \cellcolor[HTML]{FFFFB2}0.961 & \cellcolor[HTML]{FFFFB2}0.971 & 0.985 & 0.982 & \cellcolor[HTML]{FFFFB2}0.969 & \cellcolor[HTML]{FFD8B2}0.972 & \cellcolor[HTML]{FFD8B2}0.979 & \cellcolor[HTML]{FFFFB2}0.973 \\
\multirow{-3}{*}{\textbf{\begin{tabular}[c]{@{}c@{}}GS-ICP-\\ SLAM\end{tabular}}} & \textbf{LPIPS ↓} & \cellcolor[HTML]{FFD8B2}0.035 & \cellcolor[HTML]{FFD8B2}0.049 & 0.041 & \cellcolor[HTML]{FFD8B2}0.020 & 0.028 & 0.055 & \cellcolor[HTML]{FFFFB2}0.042 & \cellcolor[HTML]{FFD8B2}0.028 & \cellcolor[HTML]{FFD8B2}0.037 \\
 \midrule
 & \textbf{PSNR↑} & 31.92 & 31.84 & \cellcolor[HTML]{FFD8B2}37.59 & 36.00 & 37.28 & 33.43 & 34.27 & 36.75 & 34.88 \\
 & \textbf{SSIM ↑} & 0.922 & 0.924 & 0.970 & 0.954 & 0.956 & 0.943 & 0.946 & 0.961 & 0.947 \\
\multirow{-3}{*}{\textbf{CaRtGS}} & \textbf{LPIPS ↓} & 0.058 & 0.077 & \cellcolor[HTML]{FFD8B2}0.029 & 0.060 & 0.038 & 0.070 & 0.059 & \cellcolor[HTML]{FFFFB2}0.043 & 0.054 \\
 \midrule
 & \textbf{PSNR↑} & 33.37 & \cellcolor[HTML]{FFFFB2}34.96 & 36.17 & 40.14 & 41.11 & \cellcolor[HTML]{FFFFB2}35.23 & 34.93 & \cellcolor[HTML]{FFFFB2}36.76 & \cellcolor[HTML]{FFFFB2}36.59 \\
 & \textbf{SSIM ↑} & 0.945 & 0.957 & 0.966 & 0.979 & 0.981 & 0.962 & 0.962 & 0.965 & 0.965 \\
\multirow{-3}{*}{\textbf{\begin{tabular}[c]{@{}c@{}}HI-\\ SLAM2\end{tabular}}} & \textbf{LPIPS ↓} & \cellcolor[HTML]{FFFFB2}0.054 & \cellcolor[HTML]{FFFFB2}0.052 & 0.043 & 0.030 & \cellcolor[HTML]{FFD8B2}0.025 & \cellcolor[HTML]{FFFFB2}0.053 & 0.046 & 0.053 & \cellcolor[HTML]{FFFFB2}0.044 \\
 \midrule
 & \textbf{PSNR↑} & \cellcolor[HTML]{FFB2B2}37.93 & \cellcolor[HTML]{FFB2B2}38.27 & \cellcolor[HTML]{FFB2B2}38.68 & \cellcolor[HTML]{FFB2B2}42.01 & \cellcolor[HTML]{FFD8B2}42.06 & \cellcolor[HTML]{FFB2B2}38.23 & \cellcolor[HTML]{FFB2B2}38.29 & \cellcolor[HTML]{FFB2B2}40.38 & \cellcolor[HTML]{FFB2B2}39.48 \\
 & \textbf{SSIM ↑} & \cellcolor[HTML]{FFD8B2}0.979 & \cellcolor[HTML]{FFB2B2}0.978 & \cellcolor[HTML]{FFD8B2}0.979 & \cellcolor[HTML]{FFB2B2}0.989 & \cellcolor[HTML]{FFFFB2}0.983 & \cellcolor[HTML]{FFB2B2}0.980 & \cellcolor[HTML]{FFB2B2}0.981 & \cellcolor[HTML]{FFB2B2}0.985 & \cellcolor[HTML]{FFB2B2}0.982 \\
\multirow{-3}{*}{\textbf{Ours}} & \textbf{LPIPS ↓} & \cellcolor[HTML]{FFB2B2}0.020 & \cellcolor[HTML]{FFB2B2}0.027 & \cellcolor[HTML]{FFB2B2}0.028 & \cellcolor[HTML]{FFB2B2}0.013 & \cellcolor[HTML]{FFB2B2}0.024 & \cellcolor[HTML]{FFB2B2}0.021 & \cellcolor[HTML]{FFB2B2}0.018 & \cellcolor[HTML]{FFB2B2}0.016 & \cellcolor[HTML]{FFB2B2}0.021 \\ \bottomrule
\end{tabular}
}
\end{table}

\section{Benchmark Results} \label{BR}

% We comprehensively evaluate OmniMap across multiple benchmarks encompassing optical, geometric, and semantic performance. To demonstrate the framework's practical utility, we present the frame rate during map construction and final model size. Furthermore, we perform systematic ablation studies to verify the effectiveness of our design choices.

\begin{figure*}[!t]\centering
	\includegraphics[width=16cm]{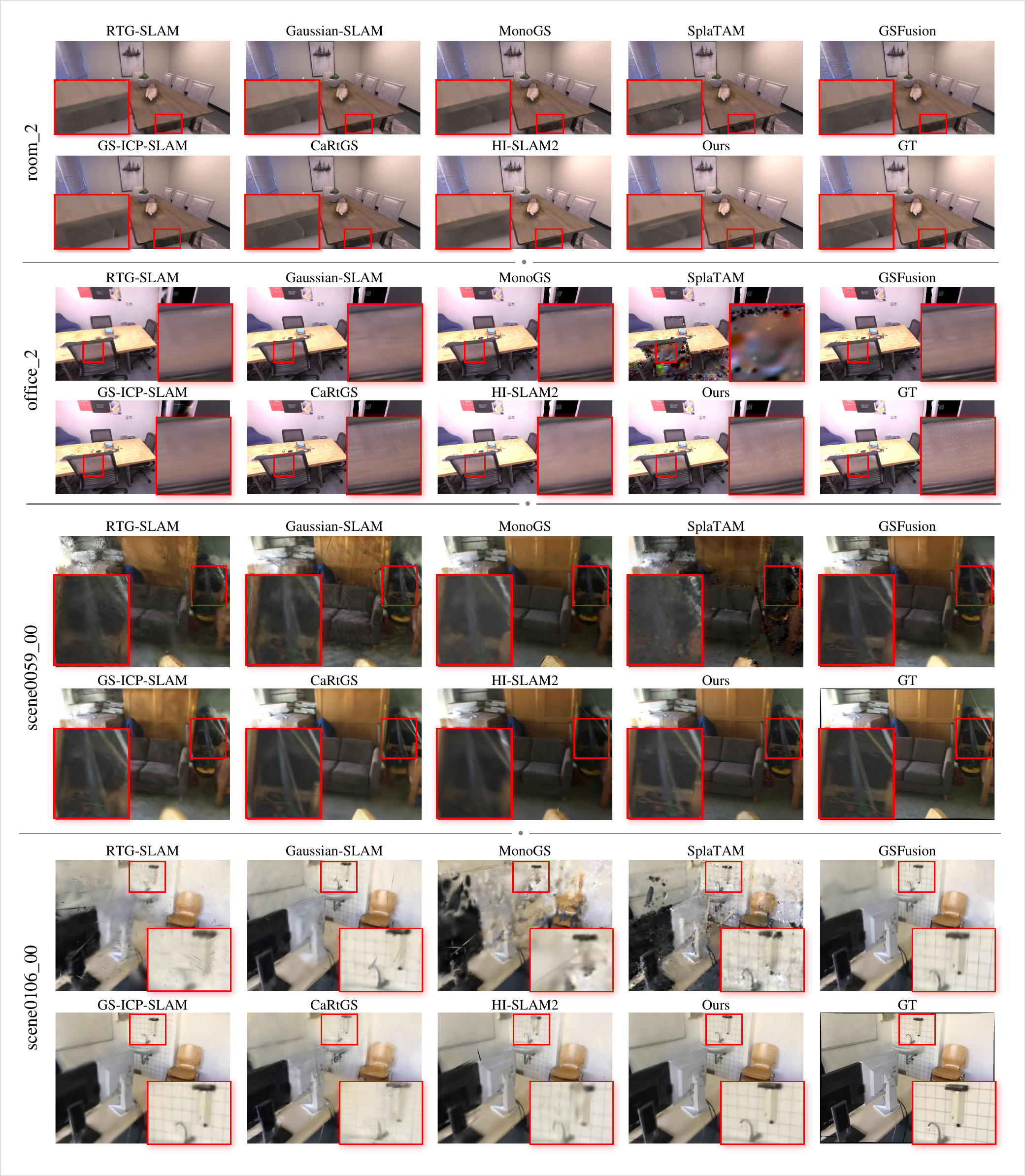}
	\caption{The rendering comparisons across diverse scenes demonstrate OmniMap's superior visual quality, achieving both higher overall fidelity and more accurate texture details than all baseline methods.}
	\label{render} 
\end{figure*}

\subsection{Experimental Setup}

\textbf{Implementation Details:} 
Our implementation primarily utilizes the PyTorch framework and is tested on a single RTX 4090 GPU. In all experiments, we set the resolution of the voxels to 0.03m to balance precision and memory. 
For the initial frame, we execute 1000 warm-up iterations. For subsequent frames, we process each input frame once, along with 19 randomly selected keyframes from the buffer. No Gaussian densification or pruning is performed beyond the voxel-based frame initialization.

\textbf{Baseline:} We benchmark OmniMap against SOTA methods in both 3DGS incremental mapping and open-vocabulary understanding. For rendering and reconstruction evaluations, we compare it with 8 3DGS incremental mapping methods (\underline{RTG-SLAM} \cite{Rtg-slam}, \underline{Gaussian-SLAM} \cite{Gaussian-slam}, \underline{MonoGS} \cite{monogs}, \underline{SplaTAM} \cite{Splatam}, \underline{GSFusion} \cite{Gsfusion}, \underline{GS-ICP-SLAM} \cite{gs-icp-slam}, \underline{CaRtGS} \cite{CaRtGS}, and \underline{HI-SLAM2} \cite{hi-slam2}) providing all methods with ground-truth poses $\mathcal{P}$ for fair comparison. For open-vocabulary semantic segmentation, we evaluate it against 3 3DGS-based approaches (\underline{OpenGaussian} \cite{opengaussian}, \underline{LangSplat} \cite{langsplat}, and \underline{GraspSplats} \cite{Graspsplats}) and 4 incremental methods (\underline{ConceptFusion} \cite{conceptfusion}, \underline{ConceptGraph} \cite{conceptgraphs}, \underline{OpenFusion} \cite{open-fusion}, and \underline{HOV-SG} \cite{hov-sg}). For fair comparison, we reproduced all baseline methods using their official implementations in our local environment.

\textbf{Dataset and Metrics:}
We select eight scenes from the synthetic \underline{Replica} \cite{replica} dataset and six scenes from the real-world \underline{ScanNet} \cite{scannet} dataset to represent a diverse set of environments. 
For rendering quality evaluation, we employ standard metrics: \underline{PSNR}, \underline{SSIM}, and \underline{LPIPS}. Geometric reconstruction is assessed using Accuracy (\underline{Acc.}), completeness (\underline{Comp.}), completion rate (\underline{Comp. Rat.}), and \underline{F-score}. For open-vocabulary semantic understanding, we measured: mean IoU (\underline{mIoU}), frequency-weighted IoU (\underline{fIoU}), mean accuracy (\underline{mAcc}), frequency-weighted accuracy (\underline{fAcc}).

\begin{table}[!t]
\scriptsize
\centering
\caption{Rendering quality evaluations on ScanNet dataset.}
\label{scannet_render_table}
\renewcommand\arraystretch{1.1}
\setlength{\tabcolsep}{1.3mm}
{
\begin{tabular}{ccccccccc}
\toprule
\textbf{Method} & \textbf{Metric} & \textbf{0000} & \textbf{0059} & \textbf{0106} & \textbf{0169} & \textbf{0181} & \textbf{0207} & \textbf{Avg.} \\  \midrule
 & \textbf{PSNR↑} & 19.07 & 17.45 & 15.83 & 18.88 & 18.17 & 19.27 & 18.11 \\
 & \textbf{SSIM ↑} & \cellcolor[HTML]{FFB2B2}0.820 & 0.747 & 0.712 & 0.792 & 0.776 & \cellcolor[HTML]{FFD8B2}0.790 & 0.773 \\
\multirow{-3}{*}{\textbf{\begin{tabular}[c]{@{}c@{}}RTG-\\ SLAM\end{tabular}}} & \textbf{LPIPS ↓} & 0.469 & 0.500 & 0.548 & 0.471 & 0.551 & 0.488 & 0.504 \\  \midrule
 & \textbf{PSNR↑} & \cellcolor[HTML]{FFFFB2}23.74 & \cellcolor[HTML]{FFD8B2}21.51 & \cellcolor[HTML]{FFFFB2}22.09 & \cellcolor[HTML]{FFFFB2}25.03 & \cellcolor[HTML]{FFD8B2}22.95 & \cellcolor[HTML]{FFD8B2}24.75 & \cellcolor[HTML]{FFFFB2}23.35 \\
 & \textbf{SSIM ↑} & 0.754 & 0.739 & 0.785 & 0.789 & 0.811 & \cellcolor[HTML]{FFFFB2}0.785 & 0.777 \\
\multirow{-3}{*}{\textbf{\begin{tabular}[c]{@{}c@{}}Gaussian-\\ SLAM\end{tabular}}} & \textbf{LPIPS ↓} & 0.475 & 0.430 & 0.418 & 0.401 & 0.519 & 0.450 & 0.449 \\  \midrule
 & \textbf{PSNR↑} & 19.60 & 18.13 & 17.73 & 20.42 & 17.93 & 19.48 & 18.88 \\
 & \textbf{SSIM ↑} & 0.696 & 0.663 & 0.687 & 0.726 & 0.743 & 0.710 & 00.70 \\
\multirow{-3}{*}{\textbf{MonoGS}} & \textbf{LPIPS ↓} & 0.499 & 0.454 & 0.527 & 0.455 & 0.519 & 0.477 & 00.49 \\  \midrule
 & \textbf{PSNR↑} & 19.19 & 19.09 & 17.99 & 22.21 & 16.72 & 19.31 & 19.09 \\
 & \textbf{SSIM ↑} & 0.655 & 0.773 & 0.700 & 0.789 & 0.670 & 0.685 & 0.712 \\
\multirow{-3}{*}{\textbf{SplaTAM}} & \textbf{LPIPS ↓} & 0.427 & \cellcolor[HTML]{FFD8B2}0.290 & \cellcolor[HTML]{FFFFB2}0.350 & \cellcolor[HTML]{FFD8B2}0.285 & \cellcolor[HTML]{FFD8B2}0.416 & \cellcolor[HTML]{FFB2B2}0.349 & \cellcolor[HTML]{FFD8B2}0.353 \\  \midrule
 & \textbf{PSNR↑} & 18.98 & 18.86 & 19.22 & 19.29 & 17.79 & 20.04 & 19.03 \\
 & \textbf{SSIM ↑} & 0.733 & 0.764 & 0.811 & 0.708 & 0.745 & 0.769 & 0.755 \\
\multirow{-3}{*}{\textbf{GSFusion}} & \textbf{LPIPS ↓} & 0.419 & 0.355 & 0.353 & 0.389 & 0.491 & 0.437 & 0.407 \\  \midrule
 & \textbf{PSNR↑} & 21.16 & 19.53 & 18.19 & 20.57 & 21.02 & 21.57 & 20.34 \\
 & \textbf{SSIM ↑} & 0.762 & 0.759 & 0.782 & 0.784 & \cellcolor[HTML]{FFD8B2}0.830 & 0.772 & \cellcolor[HTML]{FFFFB2}0.782 \\
\multirow{-3}{*}{\textbf{\begin{tabular}[c]{@{}c@{}}GS-ICP-\\ SLAM\end{tabular}}} & \textbf{LPIPS ↓} & 0.438 & 0.429 & 0.440 & 0.402 & 0.493 & 0.477 & 0.447 \\  \midrule
 & \textbf{PSNR↑} & 19.57 & 20.00 & 19.24 & 20.31 & 17.53 & 18.04 & 19.12 \\
 & \textbf{SSIM ↑} & 0.759 & \cellcolor[HTML]{FFD8B2}0.812 & \cellcolor[HTML]{FFFFB2}0.822 & \cellcolor[HTML]{FFFFB2}0.796 & 0.792 & 0.692 & 0.779 \\
\multirow{-3}{*}{\textbf{CaRtGS}} & \textbf{LPIPS ↓} & \cellcolor[HTML]{FFFFB2}0.397 & 0.353 & 0.364 & \cellcolor[HTML]{FFFFB2}0.362 & 0.590 & 0.535 & 0.433 \\  \midrule
 & \textbf{PSNR↑} & \cellcolor[HTML]{FFD8B2}24.10 & \cellcolor[HTML]{FFFFB2}21.33 & \cellcolor[HTML]{FFD8B2}22.40 & \cellcolor[HTML]{FFD8B2}26.53 & \cellcolor[HTML]{FFFFB2}22.23 & \cellcolor[HTML]{FFFFB2}23.65 & \cellcolor[HTML]{FFD8B2}23.37 \\
 & \textbf{SSIM ↑} & \cellcolor[HTML]{FFFFB2}0.785 & \cellcolor[HTML]{FFFFB2}0.793 & \cellcolor[HTML]{FFD8B2}0.825 & \cellcolor[HTML]{FFD8B2}0.805 & \cellcolor[HTML]{FFFFB2}0.829 & 0.774 & \cellcolor[HTML]{FFD8B2}0.802 \\
\multirow{-3}{*}{\textbf{\begin{tabular}[c]{@{}c@{}}HI-\\ SLAM2\end{tabular}}} & \textbf{LPIPS ↓} & \cellcolor[HTML]{FFD8B2}0.368 & \cellcolor[HTML]{FFFFB2}0.341 & \cellcolor[HTML]{FFD8B2}0.331 & 0.372 & \cellcolor[HTML]{FFFFB2}0.424 & \cellcolor[HTML]{FFFFB2}0.429 & \cellcolor[HTML]{FFFFB2}0.378 \\  \midrule
 & \textbf{PSNR↑} & \cellcolor[HTML]{FFB2B2}26.61 & \cellcolor[HTML]{FFB2B2}24.30 & \cellcolor[HTML]{FFB2B2}25.05 & \cellcolor[HTML]{FFB2B2}27.05 & \cellcolor[HTML]{FFB2B2}26.33 & \cellcolor[HTML]{FFB2B2}26.90 & \cellcolor[HTML]{FFB2B2}26.04 \\
 & \textbf{SSIM ↑} & \cellcolor[HTML]{FFD8B2}0.813 & \cellcolor[HTML]{FFB2B2}0.831 & \cellcolor[HTML]{FFB2B2}0.860 & \cellcolor[HTML]{FFB2B2}0.836 & \cellcolor[HTML]{FFB2B2}0.873 & \cellcolor[HTML]{FFB2B2}0.812 & \cellcolor[HTML]{FFB2B2}0.838 \\
\multirow{-3}{*}{\textbf{Ours}} & \textbf{LPIPS ↓} & \cellcolor[HTML]{FFB2B2}0.326 & \cellcolor[HTML]{FFB2B2}0.284 & \cellcolor[HTML]{FFB2B2}0.284 & \cellcolor[HTML]{FFB2B2}0.278 & \cellcolor[HTML]{FFB2B2}0.322 & \cellcolor[HTML]{FFD8B2}0.353 & \cellcolor[HTML]{FFB2B2}0.308
\\
\bottomrule
\end{tabular}
}
\end{table}

\subsection{Optics: Image Rendering}

To evaluate the optical reconstruction capability of OmniMap, we rendered novel views using camera trajectories $\mathcal{P}$ from the reconstructed 3DGS field $\mathcal{G}$ representation and quantitatively compared them with GT RGB images $\mathcal{C}$.
We present quantitative rendering results for the Replica and ScanNet datasets in Tab. \ref{replica_render_table} and Tab. \ref{scannet_render_table}, respectively. OmniMap achieves the best performance across almost all synthetic and real scenes, owing to several key design features: voxel-based Gaussian initialization, which ensures ample optimization potential in any region; a small number of iterations over multiple keyframes, enabling collaborative supervision from various viewpoints; and modeling of camera parameters to address motion blur and exposure inconsistencies across frames. 
The performance gain is particularly significant on ScanNet's challenging real-world scenes with frequent blurring and exposure issues. 
While alternative methods show competitive performance in specific scenarios, OmniMap consistently outperforms across metrics.

Qualitative results for representative scenes are shown in Fig. \ref{render}. In the noise-free Replica dataset (top four rows), all methods are nearly indistinguishable from the true images. However, when zooming in on details such as the stripes on the floor stand and the mesh chair backrest, OmniMap exhibits higher fidelity in color, indicating superior detailing capabilities in optical reconstruction. The accurate reproduction of fine textures further emphasizes OmniMap’s ability to capture subtle features, which may be missed by competing methods.
The ScanNet comparisons (bottom rows) provide more compelling evidence, with OmniMap accurately reconstructing challenging elements (long sticks, wall meshes) that other methods either distort or fail to capture completely. Notably, OmniMap excels at reconstructing long, thin objects and intricate details that often suffer from noise or occlusion in real-world datasets. This highlights the method's robustness and flexibility in handling diverse scene geometries and textures.
OmniMap even demonstrates superior reconstruction quality compared to GT in certain scenarios, as evidenced by scene0059\_00 where our method recovers clearer rope textures that appear blurred in the original color observation.

\begin{table}[!t]
\scriptsize
\centering
\caption{Reconstruction quality evaluations on ScanNet dataset.}
\label{scannet_mesh_table}
\renewcommand\arraystretch{1.1}
\setlength{\tabcolsep}{1.15mm}
{
\begin{tabular}{ccccccccc}
\toprule
\textbf{Method} & \textbf{Metric} & \textbf{0000} & \textbf{0059} & \textbf{0106} & \textbf{0169} & \textbf{0181} & \textbf{0207} & \textbf{Avg.} \\
\midrule
 & \textbf{Acc. ↓} & \cellcolor[HTML]{FFFFB2}03.98 & 13.67 & \cellcolor[HTML]{FFB2B2}22.84 & \cellcolor[HTML]{FFB2B2}05.23 & 08.83 & \cellcolor[HTML]{FFB2B2}08.51 & 10.51 \\
 & \textbf{Comp.  ↓} & \cellcolor[HTML]{FFD8B2}02.54 & 04.00 & 04.57 & \cellcolor[HTML]{FFD8B2}02.86 & 05.69 & \cellcolor[HTML]{FFFFB2}02.70 & 03.73 \\
 & \textbf{Comp. Rat. ↑} & \cellcolor[HTML]{FFD8B2}94.26 & 82.70 & 76.22 & \cellcolor[HTML]{FFD8B2}90.19 & 61.94 & \cellcolor[HTML]{FFFFB2}90.06 & 82.56 \\
\multirow{-4}{*}{\textbf{\begin{tabular}[c]{@{}c@{}}RTG-\\ SLAM\end{tabular}}} & \textbf{F-Score ↑} & \cellcolor[HTML]{FFFFB2}85.86 & 61.72 & \cellcolor[HTML]{FFFFB2}54.18 & \cellcolor[HTML]{FFD8B2}78.40 & 48.82 & \cellcolor[HTML]{FFFFB2}78.59 & 67.93 \\
\midrule
 & \textbf{Acc. ↓} & 04.21 & 13.54 & 25.98 & 07.86 & \cellcolor[HTML]{FFFFB2}05.12 & 09.90 & 11.10 \\
 & \textbf{Comp.  ↓} & 03.93 & 05.35 & \cellcolor[HTML]{FFFFB2}04.34 & 04.94 & 04.52 & 03.86 & 04.49 \\
 & \textbf{Comp. Rat. ↑} & 75.70 & 57.83 & 68.07 & 52.57 & 61.62 & 79.72 & 65.92 \\
\multirow{-4}{*}{\textbf{\begin{tabular}[c]{@{}c@{}}Gaussian-\\ SLAM\end{tabular}}} & \textbf{F-Score ↑} & 75.46 & 51.07 & 50.31 & 50.14 & 61.58 & 72.87 & 60.24 \\
\midrule
 & \textbf{Acc. ↓} & 04.78 & \cellcolor[HTML]{FFFFB2}11.02 & 25.33 & \cellcolor[HTML]{FFD8B2}06.91 & 05.82 & \cellcolor[HTML]{FFFFB2}08.97 & \cellcolor[HTML]{FFFFB2}10.47 \\
 & \textbf{Comp.  ↓} & 03.91 & \cellcolor[HTML]{FFFFB2}03.66 & 04.45 & 03.05 & \cellcolor[HTML]{FFFFB2}04.20 & 02.76 & \cellcolor[HTML]{FFFFB2}03.67 \\
 & \textbf{Comp. Rat. ↑} & 86.37 & \cellcolor[HTML]{FFFFB2}84.45 & 76.76 & \cellcolor[HTML]{FFFFB2}86.99 & \cellcolor[HTML]{FFFFB2}75.19 & 89.38 & \cellcolor[HTML]{FFFFB2}83.19 \\
\multirow{-4}{*}{\textbf{MonoGS}} & \textbf{F-Score ↑} & 82.50 & \cellcolor[HTML]{FFFFB2}70.09 & 53.53 & \cellcolor[HTML]{FFFFB2}77.80 & \cellcolor[HTML]{FFFFB2}71.66 & 77.62 & \cellcolor[HTML]{FFFFB2}72.20 \\
\midrule
 & \textbf{Acc. ↓} & \cellcolor[HTML]{FFD8B2}03.02 & 12.73 & 26.11 & 09.67 & \cellcolor[HTML]{FFB2B2}03.96 & 09.61 & 10.85 \\
 & \textbf{Comp.  ↓} & \cellcolor[HTML]{FFFFB2}02.55 & \cellcolor[HTML]{FFD8B2}03.15 & \cellcolor[HTML]{FFD8B2}03.74 & 03.33 & \cellcolor[HTML]{FFD8B2}03.04 & \cellcolor[HTML]{FFD8B2}02.62 & \cellcolor[HTML]{FFD8B2}03.07 \\
 & \textbf{Comp. Rat. ↑} & \cellcolor[HTML]{FFFFB2}91.04 & \cellcolor[HTML]{FFD8B2}84.95 & \cellcolor[HTML]{FFD8B2}79.68 & 84.41 & \cellcolor[HTML]{FFD8B2}86.62 & \cellcolor[HTML]{FFD8B2}91.21 & \cellcolor[HTML]{FFD8B2}86.32 \\
\multirow{-4}{*}{\textbf{SplaTAM}} & \textbf{F-Score ↑} & \cellcolor[HTML]{FFD8B2}89.77 & \cellcolor[HTML]{FFB2B2}71.28 & \cellcolor[HTML]{FFD8B2}58.29 & 73.17 & \cellcolor[HTML]{FFD8B2}84.66 & \cellcolor[HTML]{FFD8B2}81.37 & \cellcolor[HTML]{FFD8B2}76.42 \\
\midrule
 & \textbf{Acc. ↓} & 07.45 & 12.35 & 26.97 & 12.24 & 10.50 & 11.46 & 13.50 \\
 & \textbf{Comp.  ↓} & 04.80 & 04.24 & 07.89 & 07.40 & 07.67 & 05.82 & 06.30 \\
 & \textbf{Comp. Rat. ↑} & 68.39 & 72.91 & 47.62 & 43.11 & 44.32 & 60.94 & 56.22 \\
\multirow{-4}{*}{\textbf{GSFusion}} & \textbf{F-Score ↑} & 62.16 & 57.07 & 33.65 & 36.32 & 39.67 & 52.82 & 46.95 \\
\midrule
 & \textbf{Acc. ↓} & 04.80 & \cellcolor[HTML]{FFD8B2}10.96 & \cellcolor[HTML]{FFD8B2}25.10 & \cellcolor[HTML]{FFFFB2}06.93 & 05.83 & 09.00 & \cellcolor[HTML]{FFD8B2}10.44 \\
 & \textbf{Comp.  ↓} & 03.91 & 03.72 & 04.42 & \cellcolor[HTML]{FFFFB2}03.05 & 04.21 & 02.78 & 03.68 \\
 & \textbf{Comp. Rat. ↑} & 86.31 & 83.73 & \cellcolor[HTML]{FFFFB2}76.94 & 86.99 & 75.04 & 89.11 & 83.02 \\
\multirow{-4}{*}{\textbf{\begin{tabular}[c]{@{}c@{}}GS-ICP-\\ SLAM\end{tabular}}} & \textbf{F-Score ↑} & 82.42 & 69.43 & 53.63 & 77.68 & 71.45 & 77.65 & 72.04 \\
\midrule
 & \textbf{Acc. ↓} & 09.19 & 13.78 & 27.49 & 11.85 & 19.65 & 14.62 & 16.10 \\
 & \textbf{Comp.  ↓} & 05.73 & 08.92 & 08.65 & 08.22 & 15.19 & 08.81 & 09.25 \\
 & \textbf{Comp. Rat. ↑} & 63.84 & 46.23 & 45.29 & 45.71 & 26.31 & 47.55 & 45.82 \\
\multirow{-4}{*}{\textbf{CaRtGS}} & \textbf{F-Score ↑} & 51.19 & 37.47 & 31.18 & 37.24 & 22.55 & 39.02 & 36.44 \\
\midrule
 & \textbf{Acc. ↓} & \cellcolor[HTML]{FFB2B2}02.70 & \cellcolor[HTML]{FFB2B2}10.14 & \cellcolor[HTML]{FFFFB2}25.25 & 07.29 & \cellcolor[HTML]{FFD8B2}04.01 & \cellcolor[HTML]{FFD8B2}08.80 & \cellcolor[HTML]{FFB2B2}09.70 \\
 & \textbf{Comp.  ↓} & \cellcolor[HTML]{FFB2B2}02.11 & \cellcolor[HTML]{FFB2B2}02.88 & \cellcolor[HTML]{FFB2B2}03.03 & \cellcolor[HTML]{FFB2B2}02.31 & \cellcolor[HTML]{FFB2B2}02.22 & \cellcolor[HTML]{FFB2B2}02.28 & \cellcolor[HTML]{FFB2B2}02.47 \\
 & \textbf{Comp. Rat. ↑} & \cellcolor[HTML]{FFB2B2}95.71 & \cellcolor[HTML]{FFB2B2}87.70 & \cellcolor[HTML]{FFB2B2}87.56 & \cellcolor[HTML]{FFB2B2}91.33 & \cellcolor[HTML]{FFB2B2}92.42 & \cellcolor[HTML]{FFB2B2}92.71 & \cellcolor[HTML]{FFB2B2}91.24 \\
\multirow{-4}{*}{\textbf{Ours}} & \textbf{F-Score ↑} & \cellcolor[HTML]{FFB2B2}93.94 & \cellcolor[HTML]{FFD8B2}70.26 & \cellcolor[HTML]{FFB2B2}60.16 & \cellcolor[HTML]{FFB2B2}79.33 & \cellcolor[HTML]{FFB2B2}88.16 & \cellcolor[HTML]{FFB2B2}82.75 & \cellcolor[HTML]{FFB2B2}79.10 \\
\bottomrule
\end{tabular}
}
\end{table}

\begin{figure*}[!t]\centering
	\includegraphics[width=16.5cm]{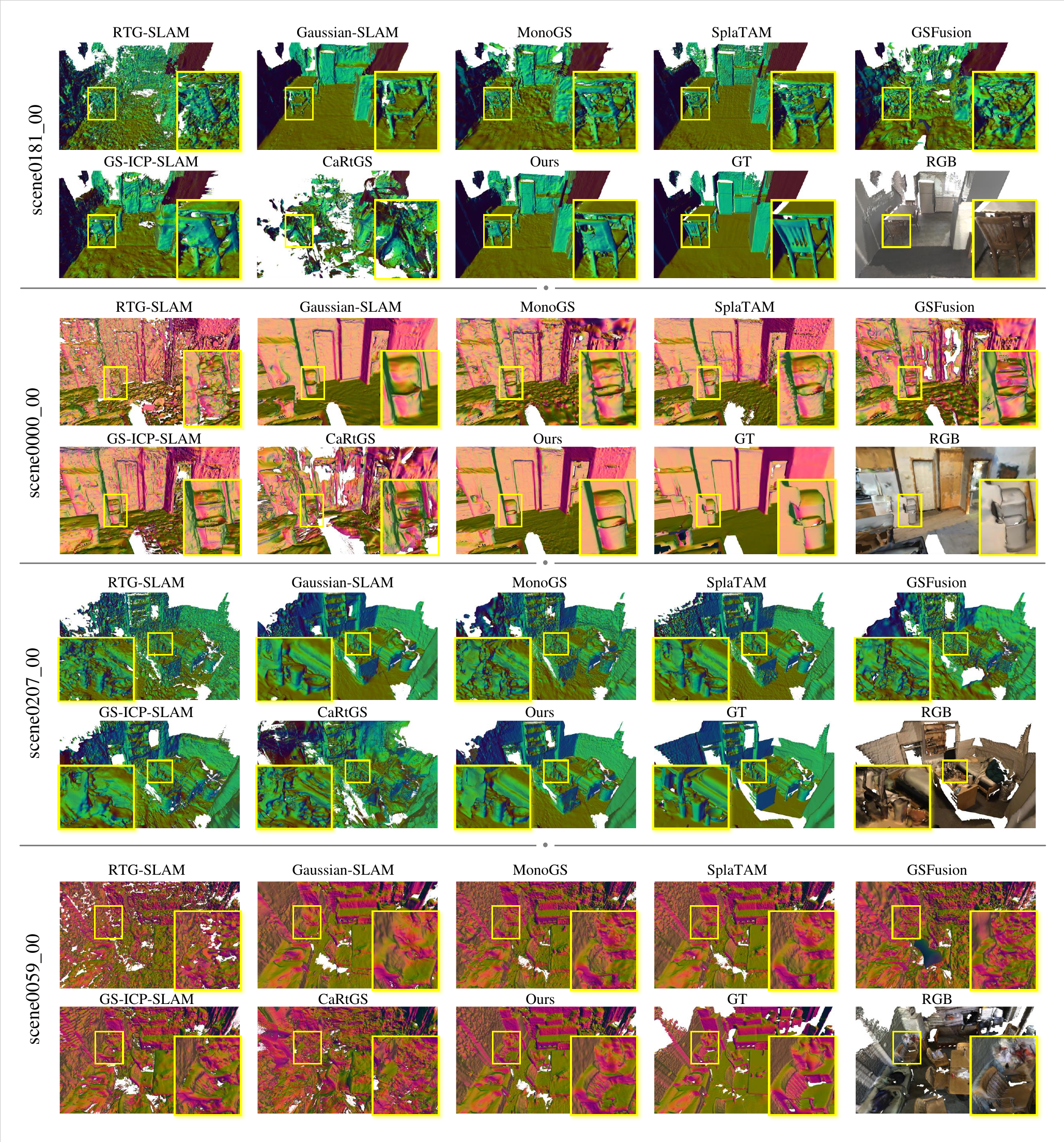}
	\caption{Mesh reconstruction quality comparison. OmniMap generates notably smoother surfaces and more geometrically precise reconstructions, particularly excelling in both continuous surface representation and fine structural detail recovery.}
	\label{mesh} 
\end{figure*}

\subsection{Geometry: Mesh Reconstruction}

To obtain the meshes, we rendered depth images from 3DGS and performed TSDF-Fusion at a consistent resolution of 0.01m for all methods. While HI-SLAM2 incorporates monocular depth estimation, as a monocular 3DGS method, it is excluded from geometric evaluation.
Furthermore, we conduct validation exclusively on the ScanNet dataset, as all methods achieve near-optimal performance on Replica.

The quantitative results are summarized in Tab. \ref{scannet_mesh_table}. All methods demonstrate generally worse Acc. than Comp., attributable to ScanNet's conservative ground truth mesh that excludes certain regions.
Despite this, it maintains strong performance through its canonical Gaussian initialization and higher-order geometric signal supervision. In contrast, GSFusion and CaRtGS produce the poorest results, as they prioritize rendering quality over effective depth image supervision, leading to suboptimal mesh reconstruction.

Qualitative results are displayed in Fig. \ref{mesh}. All methods except OmniMap show misaligned artifacts and uneven surfaces, primarily caused by noisy depth images and inter-frame depth inconsistencies. OmniMap inherits the effect of normal consistency across viewpoints, producing smoother meshes. The close-up visualization further illustrates OmniMap's advantage in reconstructing thin structures, such as chair legs, and continuous geometries, like garbage cans. Notably, OmniMap surpasses GT mesh quality in certain cases (e.g., the display screen in scene0207\_00) by inferring depth values from multi-view RGB observations that appear missing in the original depth observations.

\begin{table}[!t]
\scriptsize
\centering
\caption{Zero-shot semantic evaluations on Replica dataset.}
\label{replica_semantic_table}
\renewcommand\arraystretch{1.1}
\setlength{\tabcolsep}{0.8mm}
{
\begin{tabular}{ccccccccccc}
\toprule
\textbf{Method} & \textbf{Metric} & \textbf{ro-0} & \textbf{ro-1} & \textbf{ro-2} & \textbf{of-0} & \textbf{of-1} & \textbf{of-2} & \textbf{of-3} & \textbf{of-4} & \textbf{Avg.} \\
\midrule
 & \textbf{mIoU↑} & 11.64 & 03.82 & 09.04 & 06.06 & 04.97 & 02.10 & 05.55 & 11.36 & 06.82 \\
 & \textbf{fIoU↑} & 31.07 & 10.44 & 22.17 & 10.14 & 15.32 & 03.15 & 07.00 & 24.00 & 15.41 \\
 & \textbf{mAcc ↑} & 29.39 & 11.13 & 19.67 & 10.07 & 20.36 & 07.01 & 11.45 & 24.16 & 16.66 \\
\multirow{-4}{*}{\textbf{\begin{tabular}[c]{@{}c@{}}Open\\ Gaussian\end{tabular}}} & \textbf{fAcc ↑} & 05.96 & 14.35 & 28.09 & 12.24 & 16.90 & 04.92 & 07.90 & 24.31 & 18.08 \\
\midrule
 & \textbf{mIoU↑} & 12.62 & 11.30 & 10.48 & 10.40 & 05.38 & 10.33 & 09.55 & 09.91 & 10.00 \\
 & \textbf{fIoU↑} & 47.83 & \cellcolor[HTML]{FFFFB2}47.74 & 41.56 & 29.69 & 24.49 & \cellcolor[HTML]{FFD8B2}54.13 & 44.79 & 27.29 & 39.69 \\
 & \textbf{mAcc ↑} & 26.76 & 28.71 & 25.09 & 21.85 & 14.73 & 22.71 & 19.90 & 23.68 & 22.93 \\
\multirow{-4}{*}{\textbf{\begin{tabular}[c]{@{}c@{}}Lang\\ Splat\end{tabular}}} & \textbf{fAcc ↑} & 53.77 & 53.70 & 50.51 & 31.94 & 28.57 & \cellcolor[HTML]{FFD8B2}56.78 & 50.24 & 27.77 & 44.16 \\
\midrule
 & \textbf{mIoU↑} & 16.74 & 08.25 & 14.13 & 11.35 & 02.77 & 11.13 & 14.79 & 04.17 & 10.42 \\
 & \textbf{fIoU↑} & 55.05 & 30.75 & 48.56 & \cellcolor[HTML]{FFD8B2}51.38 & 14.12 & 45.22 & \cellcolor[HTML]{FFFFB2}45.37 & \cellcolor[HTML]{FFFFB2}50.88 & 42.67 \\
 & \textbf{mAcc ↑} & 32.92 & 21.74 & 32.45 & \cellcolor[HTML]{FFD8B2}33.51 & 08.89 & 22.12 & 27.92 & 10.80 & 23.79 \\
\multirow{-4}{*}{\textbf{\begin{tabular}[c]{@{}c@{}}Grasp\\ Splats\end{tabular}}} & \textbf{fAcc ↑} & 66.82 & 45.24 & 65.38 & \cellcolor[HTML]{FFFFB2}59.56 & 22.42 & \cellcolor[HTML]{FFFFB2}52.92 & \cellcolor[HTML]{FFFFB2}54.60 & \cellcolor[HTML]{FFFFB2}52.13 & 52.39 \\
\midrule
 & \textbf{mIoU↑} & 07.96 & 08.86 & 02.57 & 04.66 & 03.82 & 02.93 & 03.53 & 03.68 & 04.75 \\
 & \textbf{fIoU↑} & 32.28 & 34.86 & 31.65 & 23.45 & 20.35 & 23.35 & 19.15 & 17.29 & 25.30 \\
 & \textbf{mAcc ↑} & 25.41 & 31.04 & 07.59 & 17.42 & 23.38 & 12.13 & 16.65 & 20.71 & 19.29 \\
\multirow{-4}{*}{\textbf{\begin{tabular}[c]{@{}c@{}}Concept\\ Fusion\end{tabular}}} & \textbf{fAcc ↑} & 38.11 & 42.19 & 37.44 & 24.86 & 22.82 & 24.43 & 21.85 & 20.23 & 28.99 \\
\midrule
 & \textbf{mIoU↑} & \cellcolor[HTML]{FFFFB2}22.53 & \cellcolor[HTML]{FFFFB2}18.74 & 14.66 & \cellcolor[HTML]{FFD8B2}19.36 & 11.22 & \cellcolor[HTML]{FFFFB2}15.75 & 11.66 & 17.74 & \cellcolor[HTML]{FFFFB2}16.46 \\
 & \textbf{fIoU↑} & 50.28 & 46.21 & 54.52 & 43.87 & 22.00 & 23.25 & 11.32 & 34.07 & 35.69 \\
 & \textbf{mAcc ↑} & 38.88 & 36.57 & 25.10 & 29.32 & 22.54 & \cellcolor[HTML]{FFFFB2}33.65 & 28.43 & 37.56 & 31.51 \\
\multirow{-4}{*}{\textbf{\begin{tabular}[c]{@{}c@{}}Concept\\ Graph\end{tabular}}} & \textbf{fAcc ↑} & 59.65 & \cellcolor[HTML]{FFFFB2}57.93 & 62.64 & 50.21 & 25.64 & 27.78 & 16.05 & 39.63 & 42.44 \\
\midrule
 & \textbf{mIoU↑} & \cellcolor[HTML]{FFD8B2}36.26 & \cellcolor[HTML]{FFD8B2}29.00 & \cellcolor[HTML]{FFD8B2}24.13 & \cellcolor[HTML]{FFFFB2}19.14 & \cellcolor[HTML]{FFB2B2}22.69 & \cellcolor[HTML]{FFD8B2}23.88 & \cellcolor[HTML]{FFFFB2}15.28 & \cellcolor[HTML]{FFD8B2}19.89 & \cellcolor[HTML]{FFD8B2}23.79 \\
 & \textbf{fIoU↑} & \cellcolor[HTML]{FFB2B2}71.24 & \cellcolor[HTML]{FFD8B2}51.95 & \cellcolor[HTML]{FFD8B2}62.52 & 44.05 & \cellcolor[HTML]{FFD8B2}49.50 & 46.28 & 20.95 & 44.42 & \cellcolor[HTML]{FFFFB2}48.86 \\
 & \textbf{mAcc ↑} & \cellcolor[HTML]{FFD8B2}49.68 & \cellcolor[HTML]{FFD8B2}52.96 & \cellcolor[HTML]{FFFFB2}39.79 & \cellcolor[HTML]{FFFFB2}30.60 & \cellcolor[HTML]{FFB2B2}40.28 & \cellcolor[HTML]{FFD8B2}35.16 & \cellcolor[HTML]{FFFFB2}28.86 & \cellcolor[HTML]{FFFFB2}39.36 & \cellcolor[HTML]{FFD8B2}39.59 \\
\multirow{-4}{*}{\textbf{HOV-SG}} & \textbf{fAcc ↑} & \cellcolor[HTML]{FFD8B2}81.69 & \cellcolor[HTML]{FFD8B2}61.36 & \cellcolor[HTML]{FFFFB2}72.31 & 46.72 & \cellcolor[HTML]{FFB2B2}59.72 & 47.81 & 24.90 & 46.71 & \cellcolor[HTML]{FFFFB2}55.15 \\
\midrule
 & \textbf{mIoU↑} & 22.17 & 16.64 & \cellcolor[HTML]{FFFFB2}21.94 & 08.95 & \cellcolor[HTML]{FFFFB2}12.78 & 12.36 & \cellcolor[HTML]{FFD8B2}18.10 & \cellcolor[HTML]{FFFFB2}18.00 & 16.37 \\
 & \textbf{fIoU↑} & \cellcolor[HTML]{FFFFB2}60.17 & 41.62 & \cellcolor[HTML]{FFFFB2}61.72 & \cellcolor[HTML]{FFFFB2}49.64 & \cellcolor[HTML]{FFFFB2}33.86 & \cellcolor[HTML]{FFFFB2}48.22 & \cellcolor[HTML]{FFD8B2}53.00 & \cellcolor[HTML]{FFD8B2}64.93 & \cellcolor[HTML]{FFD8B2}51.65 \\
 & \textbf{mAcc ↑} & \cellcolor[HTML]{FFFFB2}41.65 & \cellcolor[HTML]{FFFFB2}41.39 & \cellcolor[HTML]{FFB2B2}43.03 & 25.61 & \cellcolor[HTML]{FFFFB2}30.01 & 26.99 & \cellcolor[HTML]{FFD8B2}32.14 & \cellcolor[HTML]{FFD8B2}40.40 & \cellcolor[HTML]{FFFFB2}35.15 \\
\multirow{-4}{*}{\textbf{\begin{tabular}[c]{@{}c@{}}Open\\ Fusion\end{tabular}}} & \textbf{fAcc ↑} & \cellcolor[HTML]{FFFFB2}72.78 & 54.68 & \cellcolor[HTML]{FFD8B2}73.71 & \cellcolor[HTML]{FFD8B2}60.10 & \cellcolor[HTML]{FFFFB2}37.01 & 52.82 & \cellcolor[HTML]{FFD8B2}62.53 & \cellcolor[HTML]{FFD8B2}69.33 & \cellcolor[HTML]{FFD8B2}60.37 \\
\midrule
 & \textbf{mIoU↑} & \cellcolor[HTML]{FFB2B2}49.57 & \cellcolor[HTML]{FFB2B2}42.63 & \cellcolor[HTML]{FFB2B2}26.50 & \cellcolor[HTML]{FFB2B2}21.92 & \cellcolor[HTML]{FFD8B2}21.32 & \cellcolor[HTML]{FFB2B2}24.45 & \cellcolor[HTML]{FFB2B2}23.31 & \cellcolor[HTML]{FFB2B2}22.82 & \cellcolor[HTML]{FFB2B2}29.06 \\
 & \textbf{fIoU↑} & \cellcolor[HTML]{FFD8B2}70.73 & \cellcolor[HTML]{FFB2B2}57.89 & \cellcolor[HTML]{FFB2B2}65.56 & \cellcolor[HTML]{FFB2B2}62.37 & \cellcolor[HTML]{FFB2B2}52.14 & \cellcolor[HTML]{FFB2B2}76.53 & \cellcolor[HTML]{FFB2B2}62.30 & \cellcolor[HTML]{FFB2B2}67.85 & \cellcolor[HTML]{FFB2B2}64.42 \\
 & \textbf{mAcc ↑} & \cellcolor[HTML]{FFB2B2}62.94 & \cellcolor[HTML]{FFB2B2}61.68 & \cellcolor[HTML]{FFD8B2}41.04 & \cellcolor[HTML]{FFB2B2}38.28 & \cellcolor[HTML]{FFD8B2}32.76 & \cellcolor[HTML]{FFB2B2}39.42 & \cellcolor[HTML]{FFB2B2}37.17 & \cellcolor[HTML]{FFB2B2}43.03 & \cellcolor[HTML]{FFB2B2}44.54 \\
\multirow{-4}{*}{\textbf{Ours}} & \textbf{fAcc ↑} & \cellcolor[HTML]{FFB2B2}82.29 & \cellcolor[HTML]{FFB2B2}69.93 & \cellcolor[HTML]{FFB2B2}74.66 & \cellcolor[HTML]{FFB2B2}71.40 & \cellcolor[HTML]{FFD8B2}55.29 & \cellcolor[HTML]{FFB2B2}81.94 & \cellcolor[HTML]{FFB2B2}70.85 & \cellcolor[HTML]{FFB2B2}71.37 & \cellcolor[HTML]{FFB2B2}72.22 \\
\bottomrule
\end{tabular}
}
\end{table}

\begin{table}[!t]
\scriptsize
\centering
\caption{Zero-shot semantic evaluations on ScanNet dataset.}
\label{scannet_semantic_table}
\renewcommand\arraystretch{1.1}
\setlength{\tabcolsep}{1.4mm}
{
\begin{tabular}{ccccccccc}
\toprule
\textbf{Method} & \textbf{Metric} & \textbf{0000} & \textbf{0059} & \textbf{0106} & \textbf{0169} & \textbf{0181} & \textbf{0207} & \textbf{Avg.} \\
\midrule
 & \textbf{mIoU↑} & 10.84 & 05.83 & 09.08 & 09.29 & 11.38 & 05.44 & 08.64 \\
 & \textbf{fIoU↑} & 18.52 & 27.51 & 15.67 & 54.55 & 19.59 & 06.44 & 23.71 \\
 & \textbf{mAcc ↑} & 21.82 & 12.75 & 22.55 & 17.15 & 24.41 & 08.50 & 17.86 \\
\multirow{-4}{*}{\textbf{\begin{tabular}[c]{@{}c@{}}Open\\ Gaussian\end{tabular}}} & \textbf{fAcc ↑} & 22.68 & 30.08 & 17.92 & 57.10 & 23.59 & 07.25 & 26.44 \\
\midrule
 & \textbf{mIoU↑} & 06.72 & 07.41 & 08.55 & 06.02 & 11.13 & 03.48 & 07.22 \\
 & \textbf{fIoU↑} & 20.03 & \cellcolor[HTML]{FFFFB2}46.98 & 29.48 & 43.60 & 23.06 & 02.38 & 27.59 \\
 & \textbf{mAcc ↑} & 18.99 & 22.48 & 25.14 & 25.39 & 28.87 & 05.18 & 21.01 \\
\multirow{-4}{*}{\textbf{\begin{tabular}[c]{@{}c@{}}Lang\\ Splat\end{tabular}}} & \textbf{fAcc ↑} & 26.26 & 52.80 & 36.09 & 47.34 & 28.21 & 02.54 & 32.21 \\
\midrule
 & \textbf{mIoU↑} & 06.27 & 08.22 & 06.46 & 08.40 & 05.53 & 05.27 & 06.69 \\
 & \textbf{fIoU↑} & 08.39 & 25.88 & 19.26 & 30.77 & 07.58 & 07.82 & 16.62 \\
 & \textbf{mAcc ↑} & 12.27 & 20.24 & 09.46 & 26.28 & 10.80 & 10.29 & 14.89 \\
\multirow{-4}{*}{\textbf{\begin{tabular}[c]{@{}c@{}}Grasp\\ Splats\end{tabular}}} & \textbf{fAcc ↑} & 09.60 & 28.87 & 21.12 & 38.66 & 09.79 & 09.40 & 19.57 \\
\midrule
 & \textbf{mIoU↑} & 11.13 & 05.48 & 09.55 & 06.15 & 09.92 & 08.77 & 08.50 \\
 & \textbf{fIoU↑} & 23.70 & 32.18 & 37.65 & 42.38 & 29.15 & 15.24 & 30.05 \\
 & \textbf{mAcc ↑} & 34.92 & 28.69 & 37.16 & 30.35 & 35.65 & 24.06 & 31.81 \\
\multirow{-4}{*}{\textbf{\begin{tabular}[c]{@{}c@{}}Concept\\ Fusion\end{tabular}}} & \textbf{fAcc ↑} & 31.50 & 37.36 & 45.10 & 47.92 & 36.52 & 21.41 & 36.64 \\
\midrule
 & \textbf{mIoU↑} & 15.38 & 16.06 & 18.57 & 17.78 & \cellcolor[HTML]{FFFFB2}16.55 & 13.42 & 16.29 \\
 & \textbf{fIoU↑} & 16.17 & 42.06 & \cellcolor[HTML]{FFD8B2}39.11 & 52.52 & \cellcolor[HTML]{FFFFB2}30.56 & 19.31 & 33.29 \\
 & \textbf{mAcc ↑} & 35.69 & 32.87 & 40.52 & 37.67 & 32.09 & 25.60 & 34.07 \\
\multirow{-4}{*}{\textbf{\begin{tabular}[c]{@{}c@{}}Concept\\ Graph\end{tabular}}} & \textbf{fAcc ↑} & 21.84 & \cellcolor[HTML]{FFFFB2}53.36 & \cellcolor[HTML]{FFD8B2}47.93 & 64.08 & \cellcolor[HTML]{FFFFB2}38.27 & 24.15 & 41.60 \\
\midrule
 & \textbf{mIoU↑} & \cellcolor[HTML]{FFB2B2}24.57 & \cellcolor[HTML]{FFB2B2}18.19 & \cellcolor[HTML]{FFD8B2}22.90 & \cellcolor[HTML]{FFB2B2}25.49 & 16.52 & \cellcolor[HTML]{FFFFB2}16.92 & \cellcolor[HTML]{FFD8B2}20.76 \\
 & \textbf{fIoU↑} & \cellcolor[HTML]{FFD8B2}31.93 & 42.53 & \cellcolor[HTML]{FFFFB2}37.66 & \cellcolor[HTML]{FFD8B2}58.84 & 28.58 & \cellcolor[HTML]{FFFFB2}30.51 & \cellcolor[HTML]{FFFFB2}38.34 \\
 & \textbf{mAcc ↑} & \cellcolor[HTML]{FFFFB2}46.78 & \cellcolor[HTML]{FFFFB2}37.98 & \cellcolor[HTML]{FFD8B2}47.41 & \cellcolor[HTML]{FFB2B2}45.21 & 37.08 & \cellcolor[HTML]{FFFFB2}34.56 & \cellcolor[HTML]{FFFFB2}41.50 \\
\multirow{-4}{*}{\textbf{HOV-SG}} & \textbf{fAcc ↑} & \cellcolor[HTML]{FFFFB2}38.74 & 51.73 & 42.41 & \cellcolor[HTML]{FFB2B2}69.97 & 33.47 & \cellcolor[HTML]{FFFFB2}36.65 & \cellcolor[HTML]{FFFFB2}45.50 \\
\midrule
 & \textbf{mIoU↑} & \cellcolor[HTML]{FFFFB2}18.58 & 12.09 & 15.76 & \cellcolor[HTML]{FFFFB2}19.32 & \cellcolor[HTML]{FFD8B2}21.26 & \cellcolor[HTML]{FFD8B2}21.11 & \cellcolor[HTML]{FFFFB2}18.02 \\
 & \textbf{fIoU↑} & \cellcolor[HTML]{FFFFB2}28.56 & \cellcolor[HTML]{FFD8B2}48.04 & 36.22 & \cellcolor[HTML]{FFB2B2}59.03 & \cellcolor[HTML]{FFD8B2}50.44 & \cellcolor[HTML]{FFD8B2}40.60 & \cellcolor[HTML]{FFD8B2}43.82 \\
 & \textbf{mAcc ↑} & \cellcolor[HTML]{FFD8B2}47.85 & \cellcolor[HTML]{FFD8B2}44.21 & \cellcolor[HTML]{FFFFB2}42.95 & \cellcolor[HTML]{FFFFB2}39.06 & \cellcolor[HTML]{FFD8B2}43.27 & \cellcolor[HTML]{FFB2B2}48.52 & \cellcolor[HTML]{FFD8B2}44.31 \\
\multirow{-4}{*}{\textbf{\begin{tabular}[c]{@{}c@{}}Open\\ Fusion\end{tabular}}} & \textbf{fAcc ↑} & \cellcolor[HTML]{FFD8B2}38.84 & \cellcolor[HTML]{FFD8B2}56.89 & 39.51 & 62.63 & \cellcolor[HTML]{FFD8B2}57.10 & \cellcolor[HTML]{FFB2B2}51.56 & \cellcolor[HTML]{FFD8B2}51.09 \\
\midrule
 & \textbf{mIoU↑} & \cellcolor[HTML]{FFD8B2}20.25 & \cellcolor[HTML]{FFFFB2}14.81 & \cellcolor[HTML]{FFB2B2}38.44 & \cellcolor[HTML]{FFD8B2}24.41 & \cellcolor[HTML]{FFB2B2}33.04 & \cellcolor[HTML]{FFB2B2}21.59 & \cellcolor[HTML]{FFB2B2}25.42 \\
 & \textbf{fIoU↑} & \cellcolor[HTML]{FFB2B2}34.24 & \cellcolor[HTML]{FFB2B2}53.59 & \cellcolor[HTML]{FFB2B2}64.18 & \cellcolor[HTML]{FFFFB2}57.34 & \cellcolor[HTML]{FFB2B2}54.96 & \cellcolor[HTML]{FFB2B2}40.86 & \cellcolor[HTML]{FFB2B2}50.86 \\
 & \textbf{mAcc ↑} & \cellcolor[HTML]{FFB2B2}49.81 & \cellcolor[HTML]{FFB2B2}47.36 & \cellcolor[HTML]{FFB2B2}62.37 & \cellcolor[HTML]{FFD8B2}41.43 & \cellcolor[HTML]{FFB2B2}61.61 & \cellcolor[HTML]{FFD8B2}42.99 & \cellcolor[HTML]{FFB2B2}50.93 \\
\multirow{-4}{*}{\textbf{Ours}} & \textbf{fAcc ↑} & \cellcolor[HTML]{FFB2B2}40.95 & \cellcolor[HTML]{FFB2B2}61.26 & \cellcolor[HTML]{FFB2B2}71.72 & \cellcolor[HTML]{FFFFB2}62.77 & \cellcolor[HTML]{FFB2B2}59.90 & \cellcolor[HTML]{FFD8B2}45.68 & \cellcolor[HTML]{FFB2B2}57.05 \\
\bottomrule
\end{tabular}
}
\end{table}

\begin{table}[!t]
\scriptsize
\centering
\caption{Comparisons of mapping efficiency and model size.}
\label{eff_table}
\renewcommand\arraystretch{1.1}
\setlength{\tabcolsep}{0.5mm}
{
\begin{tabular}{c|ccc|cccc}
\toprule
\textbf{Dataset} & \multicolumn{3}{c|}{\textbf{Replica}} & \multicolumn{4}{c}{\textbf{Scannet}} \\ \midrule
\textbf{Metric} & \textbf{PSNR ↑} & \textbf{FPS ↑} & \textbf{Size ↓} & \textbf{PSNR ↑} & \textbf{F-Score ↑} & \textbf{FPS ↑} & \textbf{Size ↓} \\ \midrule
\textbf{RTG-SLAM} & 32.21 & 2.34 & 163.4 & 18.11 & 67.93 & 0.91 & 460.0 \\
\textbf{Gaussian-SLAM} & 36.37 & 1.93 & 81.4 & \cellcolor[HTML]{FFFFB2}23.35 & 60.24 & 2.35 & 59.8 \\
\textbf{MonoGS} & 36.23 & 0.73 & \cellcolor[HTML]{FFB2B2}10.5 & 18.88 & \cellcolor[HTML]{FFFFB2}72.20 & 0.90 & \cellcolor[HTML]{FFD8B2}15.2 \\
\textbf{SplaTAM} & 34.81 & 0.30 & 58.1 & 19.09 & \cellcolor[HTML]{FFD8B2}76.42 & 0.47 & 181.5 \\
\textbf{GSFusion} & 34.70 & 3.91 & 36.2 & 19.03 & 46.95 & 5.22 & 86.7 \\
\textbf{GS-ICP-SLAM} & \cellcolor[HTML]{FFD8B2}37.47 & \cellcolor[HTML]{FFB2B2}6.56 & 21.6 & 20.34 & 72.04 & \cellcolor[HTML]{FFD8B2}7.01 & 25.0 \\
\textbf{CaRtGS} & 34.88 & 5.24 & 23.7 & 19.12 & 36.44 & \cellcolor[HTML]{FFB2B2}9.98 & \cellcolor[HTML]{FFFFB2}18.2 \\
\textbf{HI-SLAM2} & \cellcolor[HTML]{FFFFB2}36.59 & \cellcolor[HTML]{FFD8B2}5.68 & \cellcolor[HTML]{FFD8B2}11.6 & \cellcolor[HTML]{FFD8B2}23.37 & - & 3.00 & \cellcolor[HTML]{FFB2B2}10.6 \\
\textbf{Ours} & \cellcolor[HTML]{FFB2B2}39.48 & \cellcolor[HTML]{FFFFB2}5.43 & \cellcolor[HTML]{FFFFB2}14.2 & \cellcolor[HTML]{FFB2B2}26.04 & \cellcolor[HTML]{FFB2B2}79.10 & \cellcolor[HTML]{FFFFB2}5.55 & 40.5 \\ \bottomrule
\end{tabular}
}
\end{table}

\subsection{Semantics: Zero-shot Segmentation} \label{szs}

Beyond ensuring high-fidelity geometry and appearance reconstruction, OmniMap preserves instance-level scene understanding through its voxel representations $\mathcal{V}$ and embedded codebook $\mathcal{B}$. 
We evaluate OmniMap against various offline and online open-vocabulary approaches by assessing their zero-shot semantic segmentation performance. Using dataset-provided semantic labels as query texts, we assign each map element the label with maximum embedding similarity, formulating the evaluation as a 3D multi-class classification task.

\begin{figure*}[!t]\centering
	\includegraphics[width=17cm]{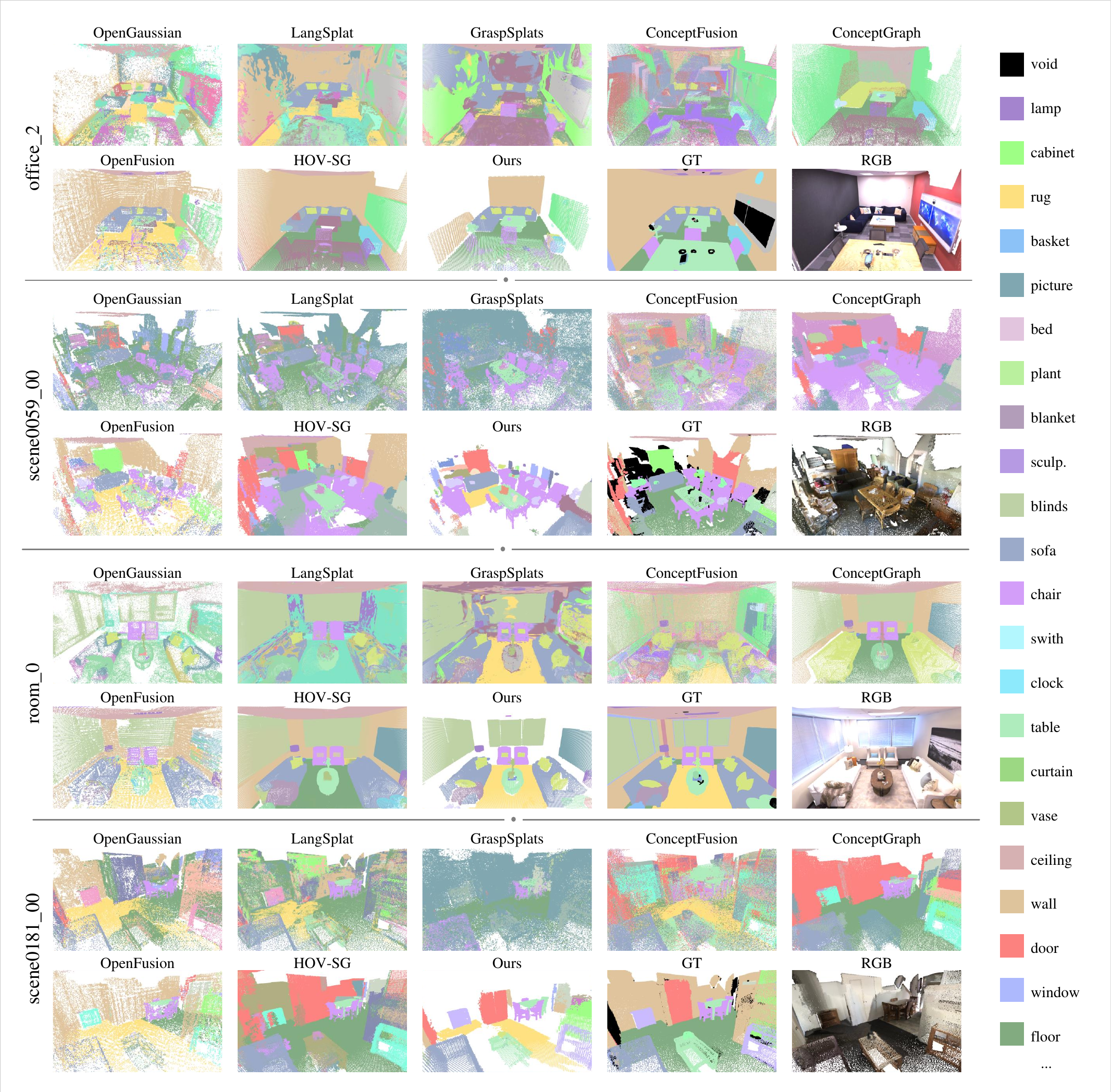}
	\caption{Zero-shot semantic segmentation results. OmniMap's probabilistic voxel representation demonstrates robustness to sensor noise while maintaining accurate object segmentation and scene understanding capabilities.}
	\label{semantic} 
\end{figure*}

Tab. \ref{replica_semantic_table} and Tab. \ref{scannet_semantic_table} summarize the qualitative results of semantic segmentation on the two datasets. OmniMap outperforms all baselines, owing to its unique probabilistic voxel representation, which enhances robustness to noise. 
For visualization purposes, we assign distinct colors to each semantic label and display the segmentation results in Fig. \ref{semantic}.
Built upon the YOLO-World model \cite{Yolo-world}, OmniMap specifically focuses on foreground object segmentation, which is particularly valuable for robotic applications.
The qualitative comparison highlights OmniMap's advantage in semantic understanding. Point-wise feature fields like LangSplat struggle to distinguish between instances, while ConceptGraph and HOV-SG, though incorporating instance notions, suffer from redundant under-segmentation due to simple point-cloud IoU-based instance associations. In contrast, OmniMap achieves nearly identical results to the GT, demonstrating competitive performance in both segmentation and classification tasks.

\subsection{Efficiency: Frame Rate and Model Size}

For real-time applications across diverse robotic platforms, frame rate and model size serve as critical metrics for evaluating algorithm performance. While theoretically any method could achieve improved results through excessive Gaussian primitives (increasing model size) or extended per-frame optimization (reducing frame rate), such approaches prove impractical for real-world deployment. To verify that OmniMap maintains efficient construction while delivering high-quality reconstructions, we measure both build frame rate and model size in Tab. \ref{eff_table}, alongside rendering and reconstruction metrics for reference. Experimental results demonstrate that OmniMap achieves an optimal balance between real-time performance (FPS), model compactness, and reconstruction quality compared to all baseline methods. This confirms OmniMap's ability to maintain high mapping performance without compromising efficiency.

\subsection{Ablation Study}

To evaluate the effectiveness of the proposed components, we conducted an ablation study for ScanNet's scene0181\_00 and report the results in Tab. \ref{ablation_table}.
The significance of probabilistic voxel representation for open-vocabulary semantics has been established in Subsection \ref{szs}. We further evaluate three additional key modules' impact on rendering and reconstruction.  Our quantitative analysis demonstrates that: (1) removal of the camera parameter modeling module leads to significant degradation in rendering quality, particularly in uncertain real-world environments; (2) while normal supervision removal has minimal impact on quantitative rendering metrics, it substantially reduces mesh accuracy, with qualitative results showing pronounced deterioration (see Fig. \ref{normal}); and (3) elimination of the keyframe buffer results in noticeable declines in both optical and geometric performance due to catastrophic forgetting effects.

\begin{table}[!t]
\scriptsize
\centering
\caption{Results of ablation study.}
\label{ablation_table}
\renewcommand\arraystretch{1.1}
\setlength{\tabcolsep}{2.5mm}
{
\begin{tabular}{ccc}
\toprule
\textbf{Metric} & \textbf{PSNR ↑} & \textbf{F-Score ↑} \\ \midrule
\textbf{w/o camera modeling} & \cellcolor[HTML]{FFFFB2}25.07 & \cellcolor[HTML]{FFFFB2}86.81 \\
\textbf{w/o normal loss} & \cellcolor[HTML]{FFD8B2}26.29 & 84.56 \\
\textbf{w/o keyframe buffer} & 21.16 & \cellcolor[HTML]{FFD8B2}86.46 \\
\textbf{Full OmniMap} & \cellcolor[HTML]{FFB2B2}26.33 & \cellcolor[HTML]{FFB2B2}88.16 \\
\bottomrule
\end{tabular}
}
\end{table}

\begin{figure}[!t]\centering
	\includegraphics[width=8.8cm]{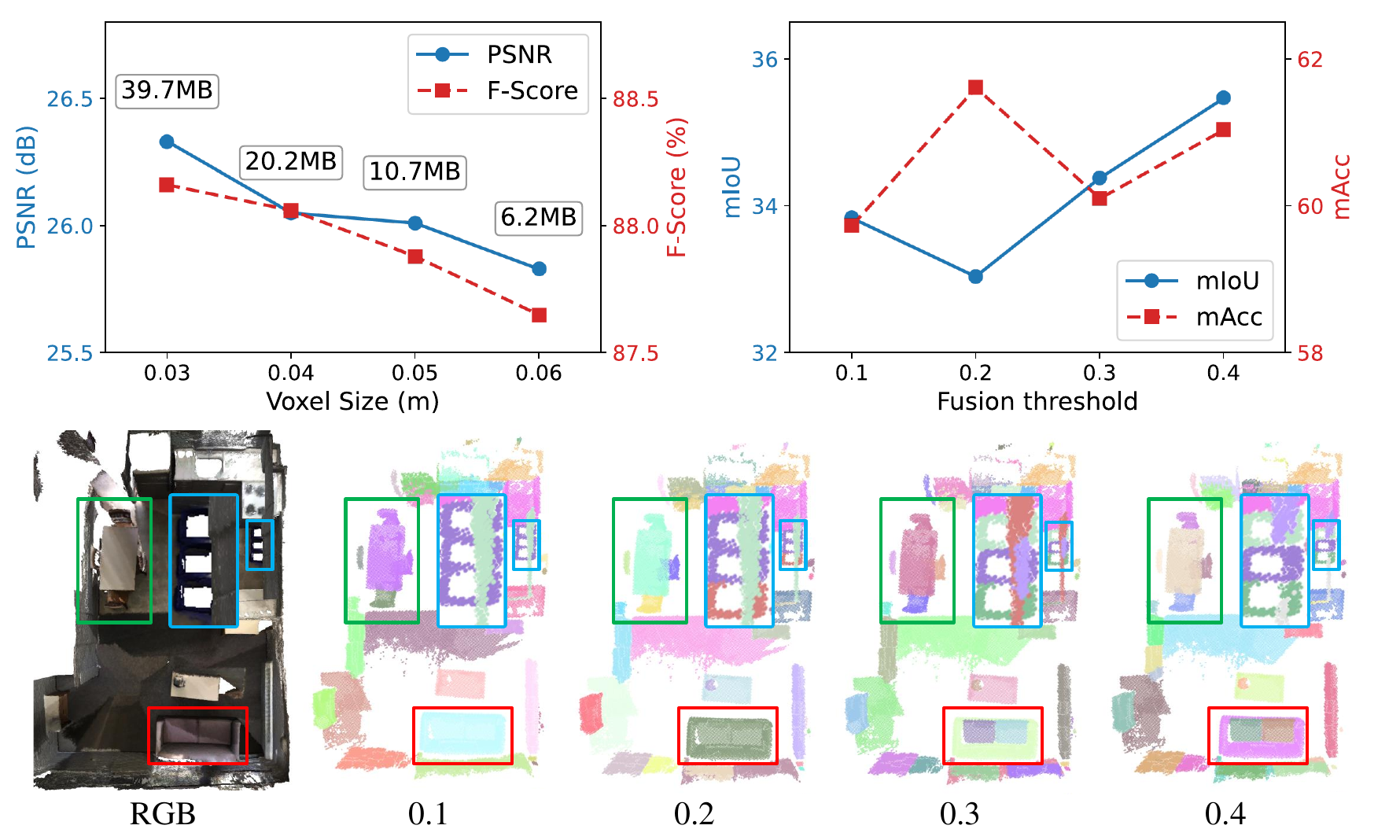}
	\caption{The impact of two key parameters on mapping performance: voxel resolution and instance fusion threshold. The bottom row shows the final instance-level maps for different fusion thresholds.}
	\label{params} 
\end{figure}

We further evaluate the impact of two key parameters on mapping performance: voxel resolution and instance fusion threshold. Voxel resolution directly determines the number of Gaussian primitives, influencing both rendering quality and mesh accuracy, while the instance fusion threshold governs instance segmentation granularity. As shown in Fig. \ref{params}, increasing the voxel size reduces model complexity (through fewer Gaussian primitives) while marginally decreasing optical and geometric accuracy, enabling a practical accuracy-efficiency trade-off. Notably, OmniMap with a voxel resolution of 0.06m achieves minimal model size (6.2 MB only), while still maintaining the highest rendering and mesh accuracy.
Although higher fusion thresholds affect semantic accuracy to some degree, the impact remains within acceptable bounds, demonstrating OmniMap's robustness to parameter variations in probabilistic modeling. The bottom row of Fig. \ref{params} illustrates fusion results at varying thresholds, highlighting differences in challenging cases such as box bases on couches, adjacent trash cans, and highly overlapping tables and chairs.

\section{Applications} \label{Ap}

OmniMap's unified representation of optical, geometric, and semantic information provides inherent advantages for supporting diverse downstream applications. This integrated framework naturally enables multiple use cases, including (1) Scene Q\&A, (2) Interactive Editing, (3) Perception-guided Manipulation, and (4) Map-assisted Navigation.

\begin{figure}[!t]\centering
	\includegraphics[width=8.3cm]{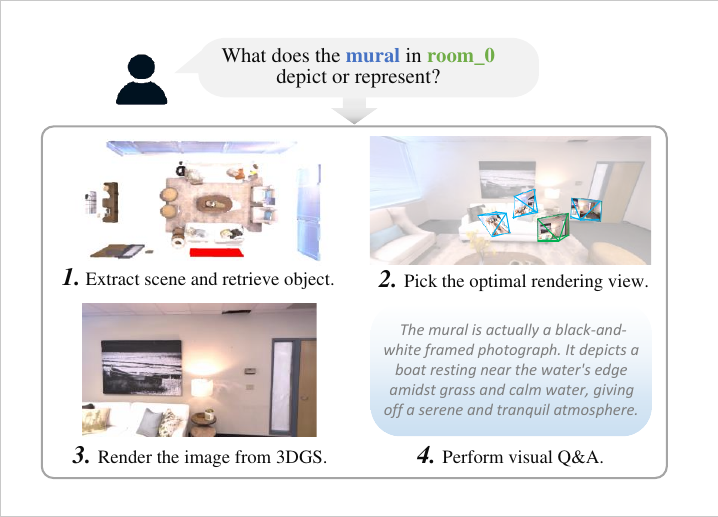}
	\caption{The process of 2D Scene Q\&A. OmniMap enables scene understanding and question answering through object retrieval and image rendering capabilities, augmented by VQA techniques.}
	\label{vqa} 
\end{figure}

\begin{figure}[!t]\centering
	\includegraphics[width=8.5cm]{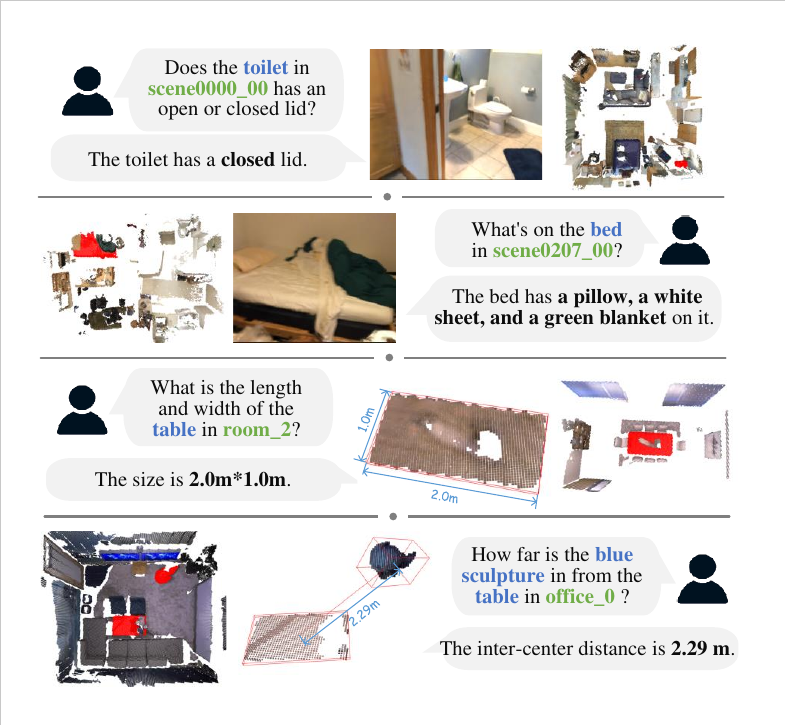}
	\caption{More demonstrations of Scene Q\&A across 2D and 3D modalities.}
	\label{more_sqa} 
\end{figure}

\subsection{Scene Q\&A}

After completing environmental reconstruction, scene quizzing serves as an effective tool for validating representation quality. User queries typically target specific object-level information, demanding rapid localization of relevant objects and extraction of fine-grained attributes like appearance. Fig. \ref{vqa} demonstrates OmniMap's Scene Q\&A pipeline with the following workflow:

\textbf{1. Object retrieval:} Leveraging the embedded codebook, we employ feature similarity matching to efficiently retrieve the most relevant object ID and localize its corresponding voxels.

\textbf{2. Viewpoint selection:} As visual perception dominates scene understanding, we project the identified voxels across multiple camera poses. By comparing projection depth with rendering depth, we select the optimal viewpoint maximizing valid voxel projections.

\textbf{3. Image rendering:} The 3DGS field is rendered at the chosen viewpoint to generate RGB and optional depth images (for 3D-related queries).

\textbf{4. Visual Q\&A:} Utilizing Multimodal Large Language Models (MLLMs) capabilities, we provide comprehensive responses to user queries. An optional feedback mechanism dynamically adjusts rendering viewpoints during Q\&A to enhance visual representation.

Beyond basic 2D scene Q\&A, OmniMap inherently supports 3D queries, enabling object dimension measurement and inter-object distance calculation. Fig. \ref{more_sqa} demonstrates additional scene query results encompassing both 2D and 3D capabilities.

\begin{figure}[!t]\centering
	\includegraphics[width=8.5cm]{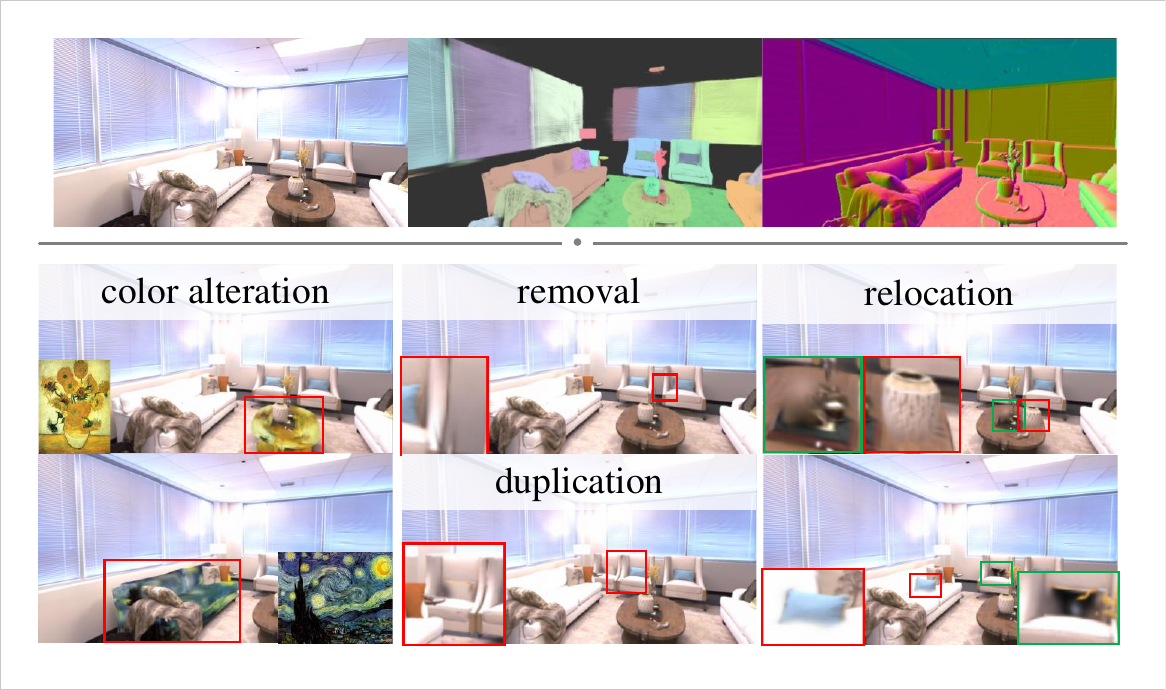}
	\caption{Scene editing demonstration. Through object querying and direct manipulation of Gaussian primitives, our framework enables flexible scene modification with multi-view rendering capabilities.}
	\label{edit} 
\end{figure}

\subsection{Interactive Editing}

Thanks to OmniMap’s instance-level scene decomposition and semantic understanding, we enable intuitive interactive scene editing with free-viewpoint rendering. Specifically, given an object-level editing command, we first retrieve the target object’s voxels. These voxels are then filtered by a distance threshold and passed to the 3D Gaussian Splatting (3DGS) field to extract the associated Gaussian primitives. By directly manipulating these primitives, the modified Gaussian field can be rendered from any viewpoint. Supported edits include color alteration, removal, duplication, and relocation, as illustrated in Fig. \ref{edit}.
This interactive editing capability allows OmniMap to efficiently generate diverse scene layouts and visual appearances while maintaining high-fidelity rendering quality, offering an efficient approach to address critical data augmentation challenges in embodied intelligence research.

\subsection{ Perception-guided Manipulation}

\begin{figure}[!t]\centering
	\includegraphics[width=8.8cm]{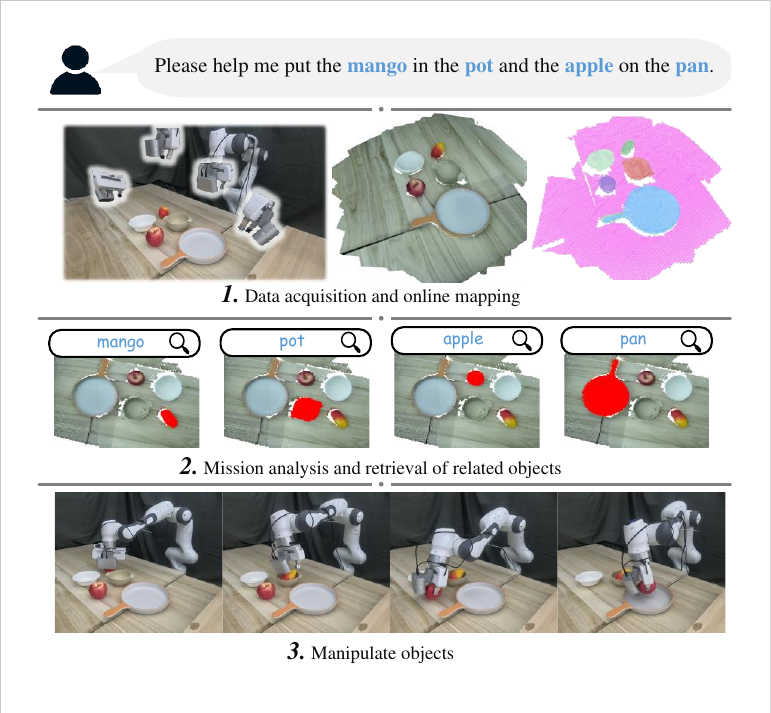}
	\caption{OmniMap supports robotic arm operations by reconstructing workspace optics, geometry, and semantics through end-effector-mounted camera scanning, enabling task completion.}
	\label{robot_arm} 
\end{figure}

Beyond interactive mapping applications, OmniMap demonstrates direct hardware control capabilities for robotic systems. We first validated this through robotic arm manipulation experiments. As illustrated in Fig. \ref{robot_arm}, an Orbbec Astra Pro camera was mounted on the end-effector of a Franka robotic arm. When presented with the user command `\textit{Please help me put the mango in the pot and the apple on the pan}', the system executed a comprehensive tabletop scanning procedure: the robotic arm maneuvered the camera around the workspace, enabling OmniMap to perform rapid incremental reconstruction. The camera pose was precisely determined through end-effector tracking combined with calibrated extrinsic parameters.

This process yielded not only a complete geometric map but also a semantic understanding of the objects within the scene. OmniMap then localized the task-relevant objects (mango, pot, apple, and pan) within the reconstructed environment. Leveraging LLM-based task decomposition, the system successfully guided the robotic arm through sequential operations to complete the requested tasks with precise object manipulation.

\subsection{Map-assisted Navigation}

We conducted real-world validation of OmniMap's effectiveness in an office environment using both a wheeled robot and a quadruped robotic platform. The wheeled robot, equipped with an Azure Kinect RGB-D camera and an on-board computing system (featuring an RTX 3090 GPU), utilized Livox MID-360 LiDAR SLAM along with multi-sensor calibration technology to generate images' poses while simultaneously streaming RGB-D data to OmniMap for processing. 

\begin{figure}[!t]\centering
	\includegraphics[width=8.5cm]{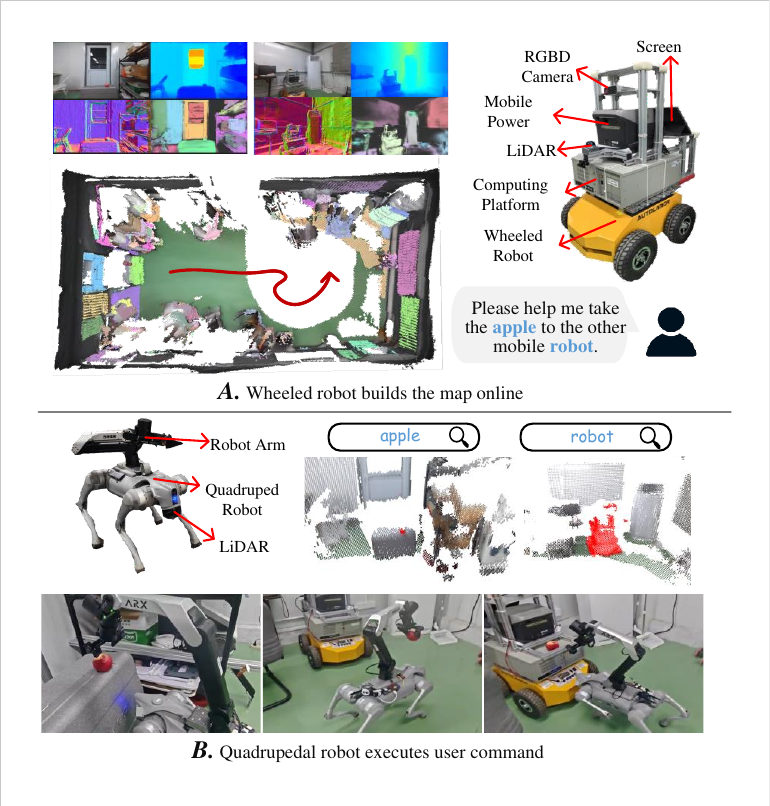}
	\caption{OmniMap enables mobile robot navigation experiments. A wheeled robot performs environmental scanning to generate comprehensive scene reconstruction and semantic understanding, while a quadruped robot with a manipulator utilizes this representation to complete user-specified tasks within the mapped environment.}
	\label{mobile_mapping} 
\end{figure}

OmniMap successfully performs online environment reconstruction as the robot traverses the scene. As illustrated in Fig. \ref{mobile_mapping}, the system demonstrates robust real-time mapping performance and produces high-quality reconstruction results. Notably, despite challenging conditions including ground surface reflections, non-uniform lighting distribution, and cluttered objects, OmniMap maintains remarkable reconstruction fidelity, with its real-time rendered output closely matching the actual RGB feed.
This achievement highlights the system's exceptional perceptual capabilities. Furthermore, the randomly color-coded instance mapping results provide compelling evidence of OmniMap's superior performance in real-world applications involving simultaneous instance detection, semantic segmentation, and environmental reconstruction tasks.

To validate the practical deployment of the reconstructed map, we implemented it on a quadruped robot equipped with a 6-DOF manipulator. As shown in Fig. \ref{mobile_mapping}, when given the natural language command `\textit{Please help me take the apple to the other mobile robot}', the system successfully identified and localized both target objects, the apple and the robot, through LLM-based semantic parsing. The robot then autonomously planned and executed a collision-free trajectory using the generated obstacle map, demonstrating precise navigation and manipulation capabilities as it grasped and transported the apple to the specified target location. This complete functional chain from perception to action conclusively verifies the operational readiness of our mapping system for real-world service robotics applications.

\section{Conclusion} \label{C}

In this work, we present OmniMap, a general framework for online mapping that simultaneously incorporates optical (high-fidelity rendering), geometric (fine-grained mesh), and semantic (open-vocabulary understanding) representations. The framework introduces several key innovations, including hybrid 3DGS-Voxel architecture, differentiable camera parameter modeling, voxel-based Gaussian initialization, higher-order geometric signal supervision, and probabilistic instance-level reconstruction. 
These technical advances enable OmniMap to achieve comprehensive optimization across all performance metrics in nearly all scenes. Efficiency evaluations demonstrate that OmniMap accomplishes these results without compromising computational efficiency or memory usage, with ablation studies confirming that even with significantly reduced model sizes, the system maintains high accuracy. Furthermore, the unified nature of OmniMap supports diverse downstream applications such as scene-based multi-domain question answering, interactive scene editing, robotic arm manipulation, and mobile robot navigation. 

However, OmniMap currently has limitations: its mapping quality depends on external positioning systems due to the lack of an integrated tracking module, and it cannot effectively handle dynamic scenes. Future work will focus on developing an integrated tracking module and implementing dynamic instance discrimination to transform OmniMap into a more versatile, plug-and-play scene representation solution.

\bibliographystyle{Bibliography/IEEEtran}
\bibliography{Bibliography/arxiv}

\begin{IEEEbiography}[{\includegraphics[width=1in,height=1.25in,clip,keepaspectratio]{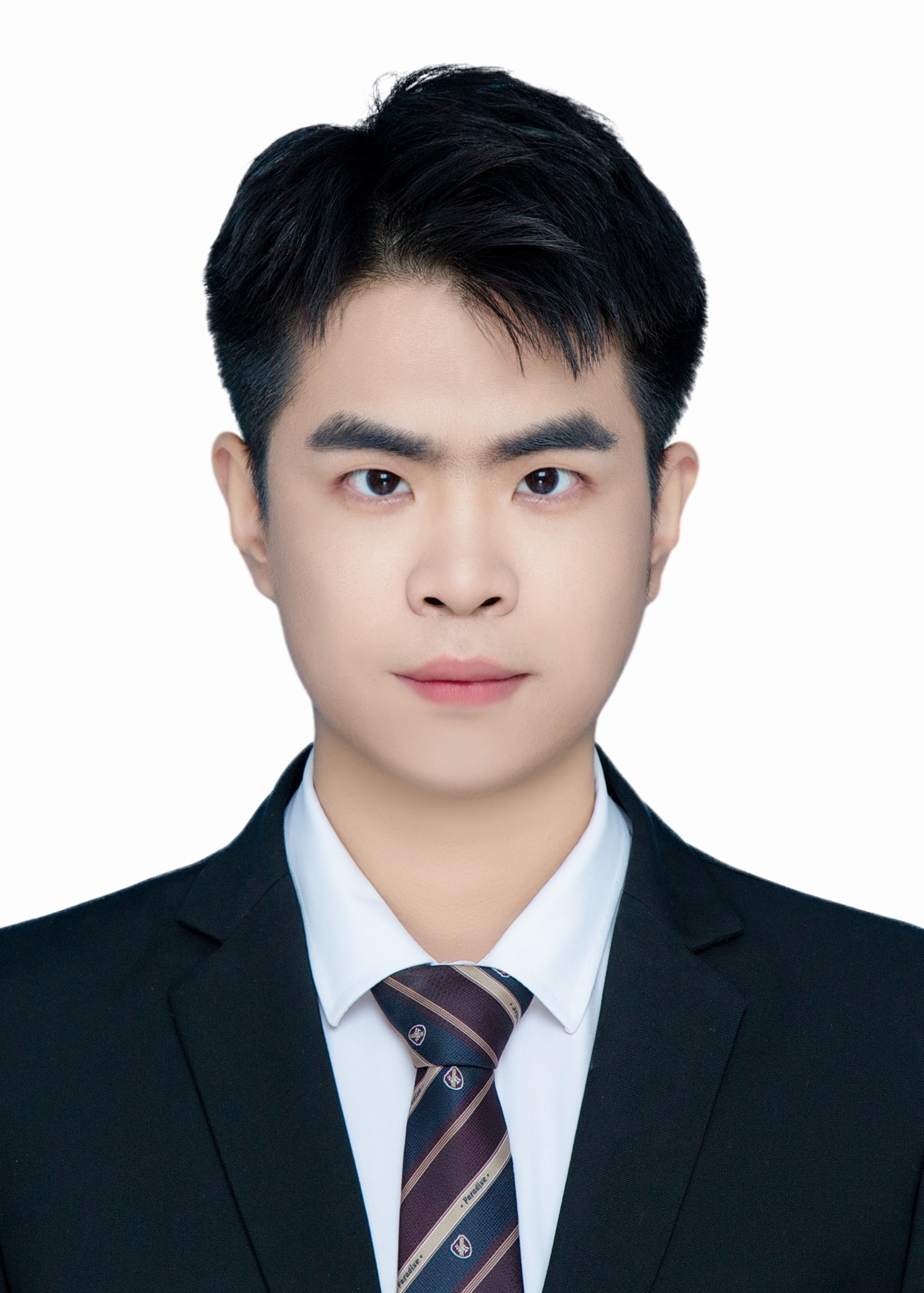}}]{Yinan Deng} (Student Member, IEEE) received the B.S. degree from Beijing Institute of Technology, Beijing, China, in 2021. He is currently pursuing his Ph.D. degree in Control Science and Engineering at the School of Automation, Beijing Institute of Technology, China. He has published 16 journal/conference papers, including IEEE TRO/TIE/RAL/IROS. His research interests include 3D vision, open vocabulary scene reconstruction, and embodied perception.
 \end{IEEEbiography}

\begin{IEEEbiography}[{\includegraphics[width=1in,height=1.25in,clip,keepaspectratio]{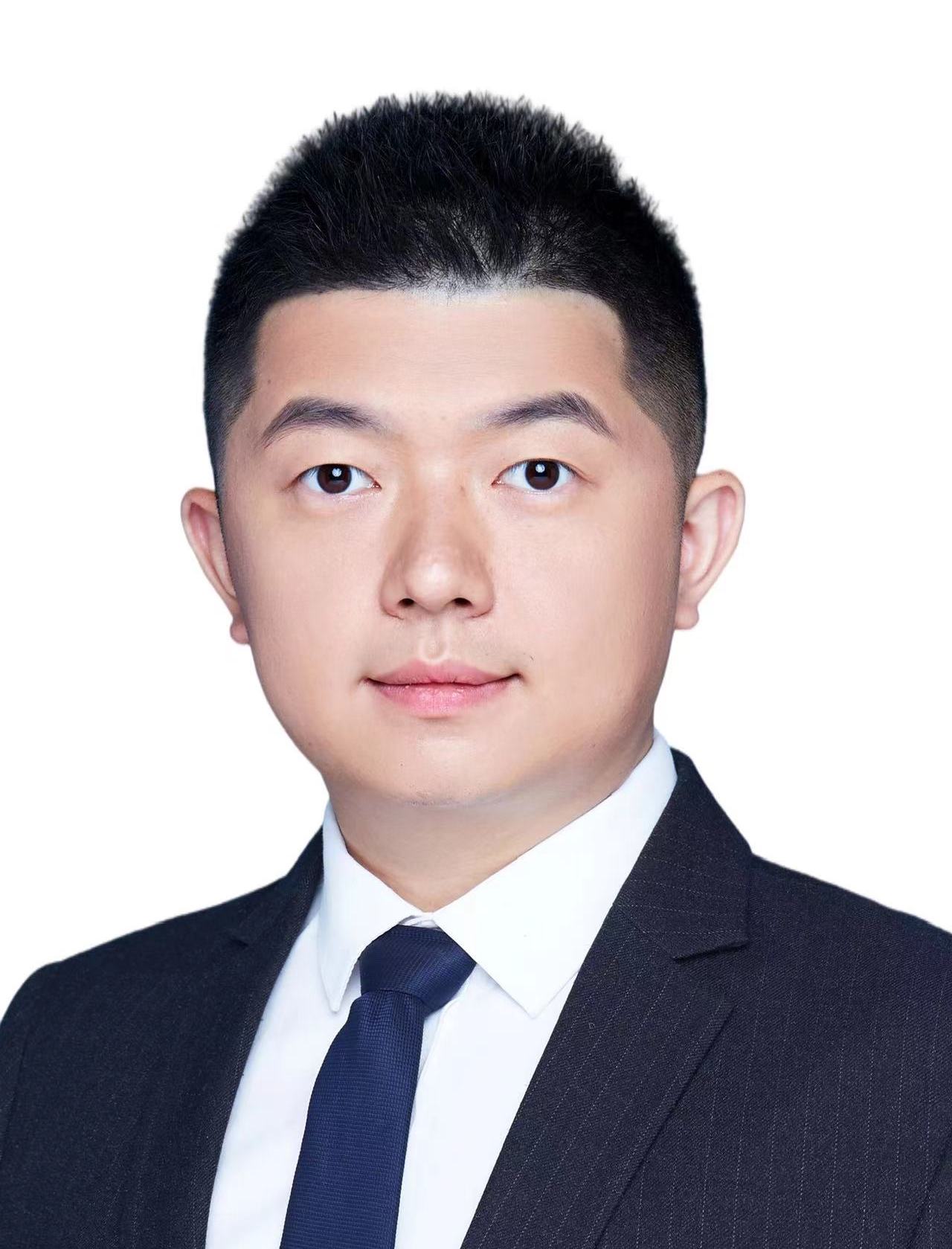}}]{Yufeng Yue}
(Member, IEEE) received the B.Eng. degree in automation from the Beijing Institute of Technology, Beijing, China, in 2014, and the Ph.D. degree in electrical and electronic engineering from Nanyang Technological University, Singapore, in 2019. 
He is currently a Professor with School of Automation, Beijing Institute of Technology. His research interests
include perception, mapping and navigation for autonomous robotics. He
has authored a book in Springer, and more than 100 journal/conference papers, including IJRR, IEEE TRO/TNNLS/TCSVT/TMech/TIE, and NeurIPS/CVPR/ICRA/IROS. He serves as an Associate Editor for  IEEE RAL/ICRA/IROS.
 \end{IEEEbiography}

\begin{IEEEbiography}[{\includegraphics[width=1in,height=1.25in,clip,keepaspectratio]{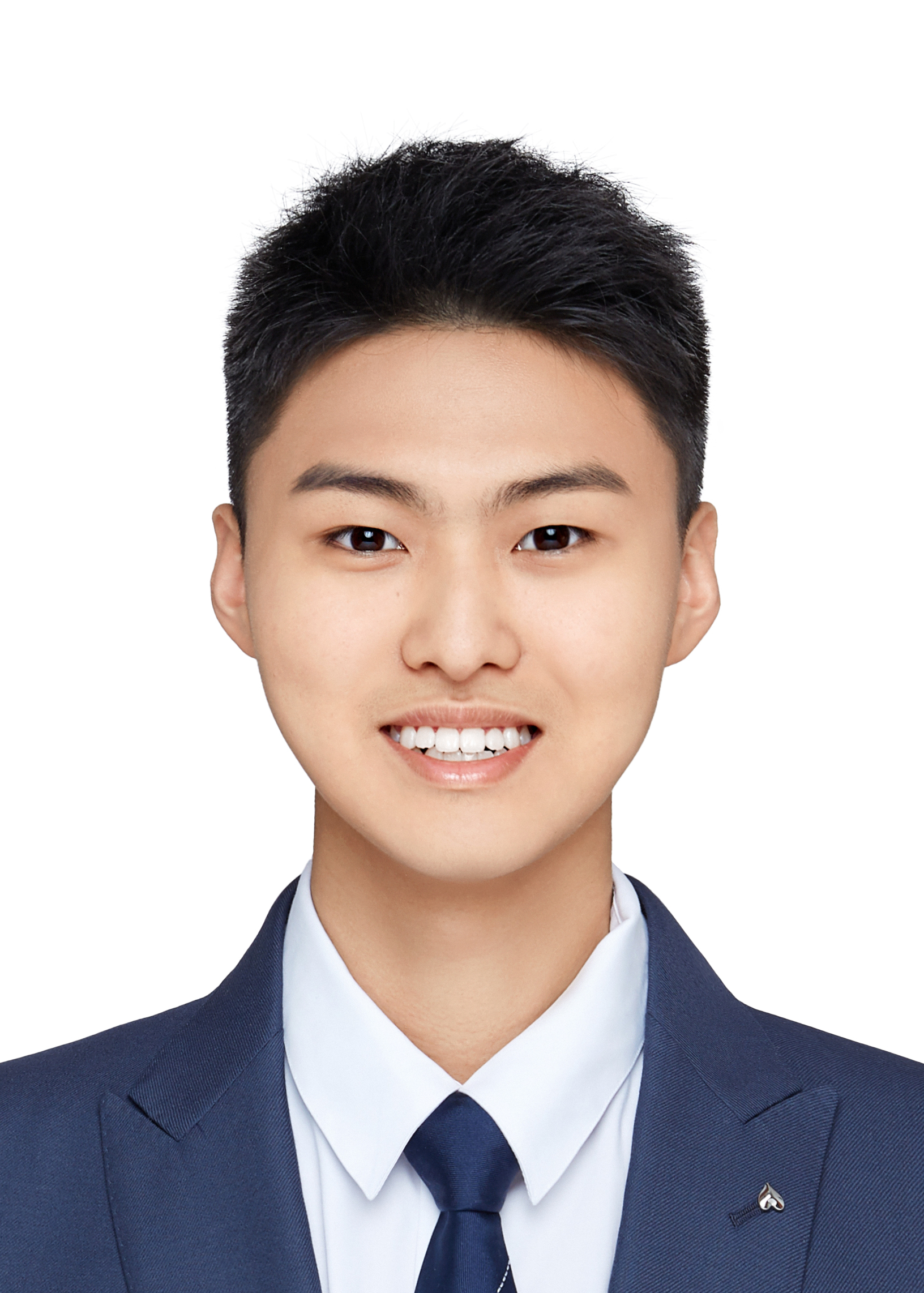}}]{Jianyu Dou} received the B.S. degree from Beijing Institute of Technology, Beijing, China, in 2025. He has just become a PhD student in Control Science and Engineering at the School of Automation, Beijing Institute of Technology, China.  He has published a journal paper on multi-agent collaborative mapping, and his current research interests focus on data generation
 \end{IEEEbiography}

\begin{IEEEbiography}[{\includegraphics[width=1in,height=1.25in,clip,keepaspectratio]{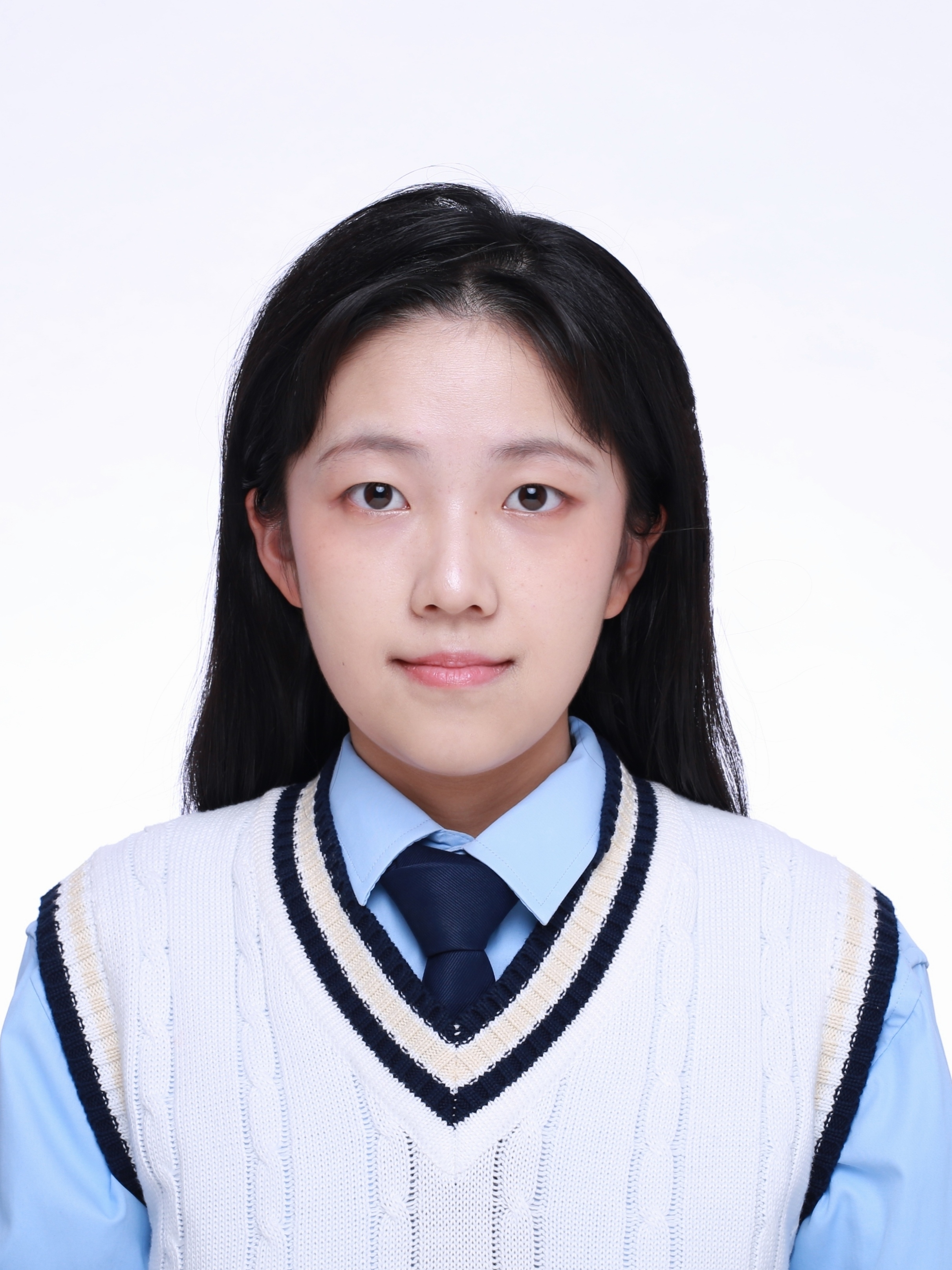}}]{Jingyu Zhao} received the bachelor's degree from the Beijing Institute of Technology, Beijing, China. She is currently pursuing a master's degree in control science and engineering at the Beijing Institute of Technology. Her research interests focus on 3D scene reconstruction and understanding.
 \end{IEEEbiography}

\begin{IEEEbiography}[{\includegraphics[width=1in,height=1.25in,clip,keepaspectratio]{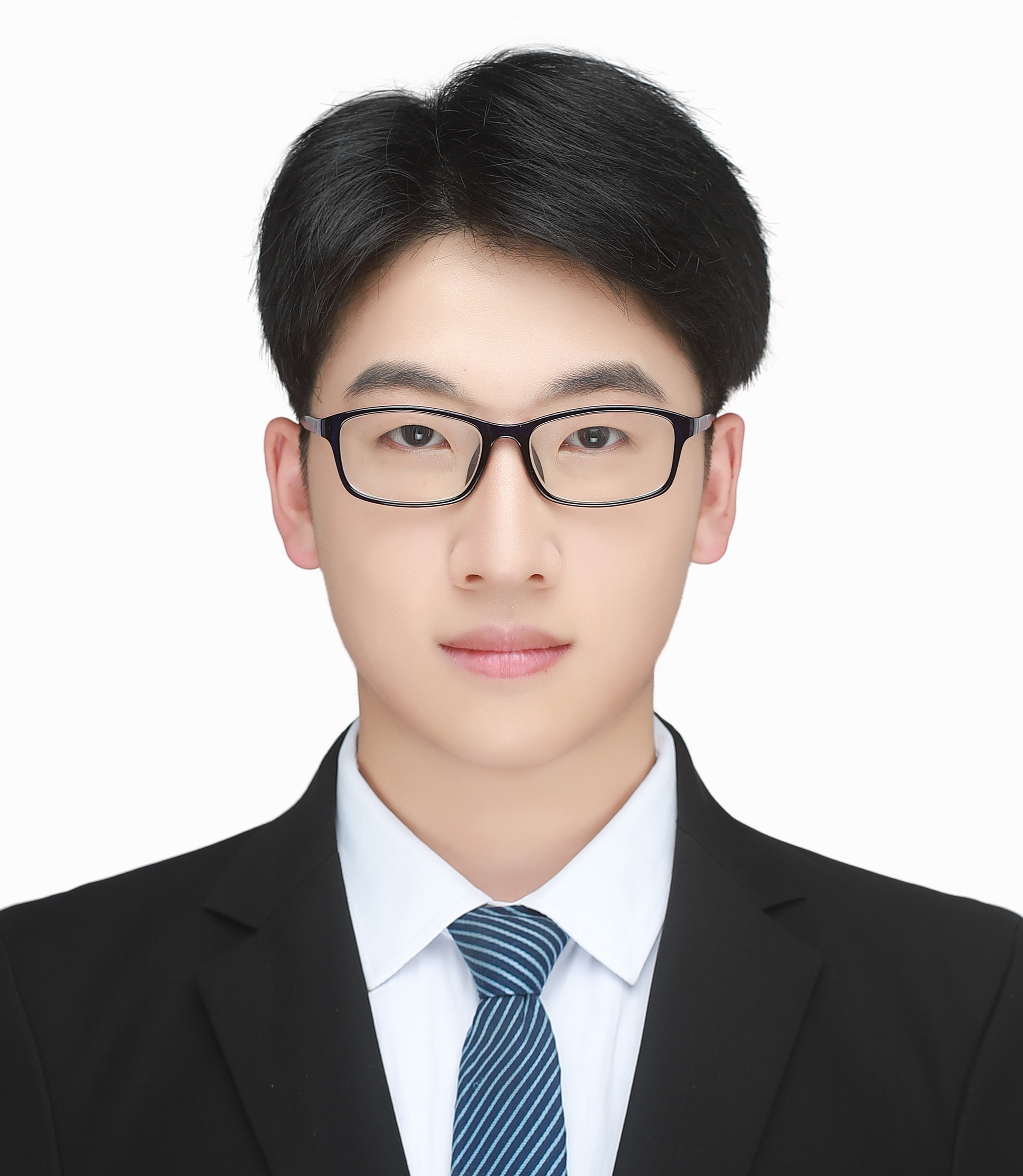}}]{Jiahui Wang} received the B.S. degree from Central South University, Changsha, China, in 2023. He is currently working toward the Ph.D. degree for the second year in cognitive space AI with School of Automation, Beijing Institute of Technology, Beijing, China. His research interests include multi-modal mapping and language embedded representation for robotic application.
 \end{IEEEbiography}

\begin{IEEEbiography}[{\includegraphics[width=1in,height=1.25in,clip,keepaspectratio]{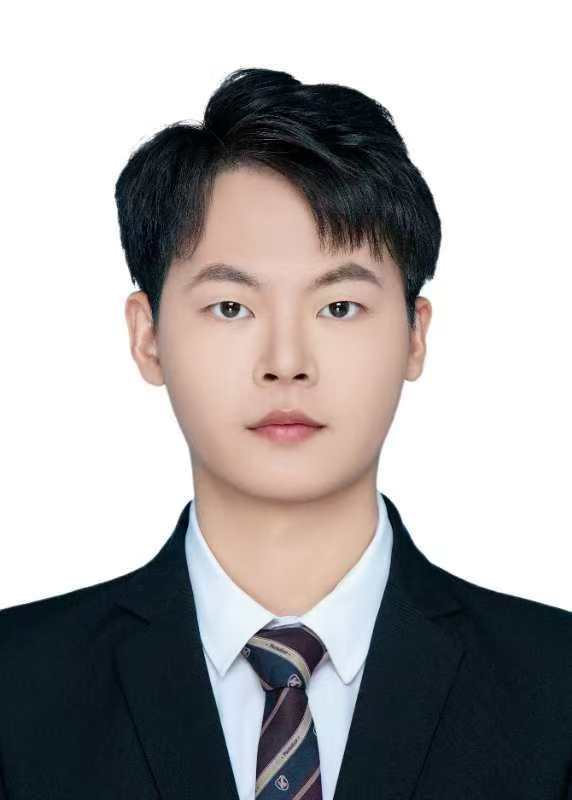}}]{Yujie Tang} (Student Member, IEEE) received the B.S. degree in automation from Beijing Institute of Technology, Beijing, China, in 2021. He is currently working toward the Ph.D. degree in intelligent navigation with School of Automation, Beijing Institute of Technology, Beijing, China. He has published 10 journal/conference papers, including TMech/RAL, and conferences like IROS. He has received 2023/2024 IEEE ICUS Best Paper Award. His research interests include robotics localization, loop closure detection, cognitive navigation.
 \end{IEEEbiography}

\begin{IEEEbiography}[{\includegraphics[width=1in,height=1.25in,clip,keepaspectratio]{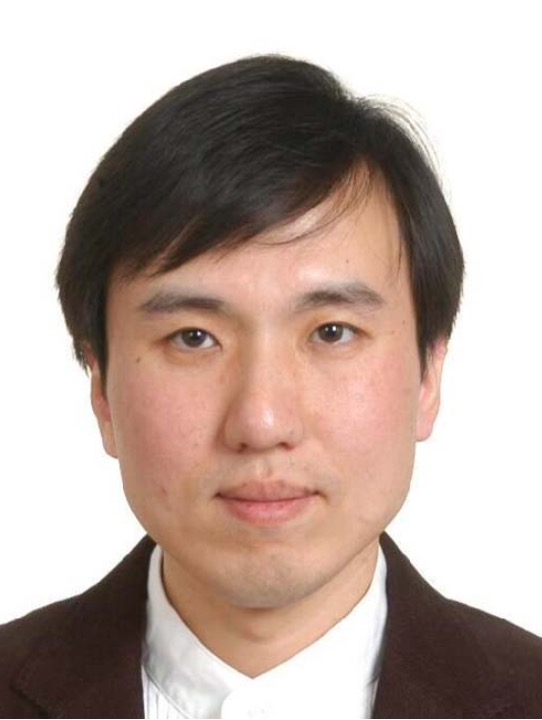}}]{Yi Yang} (Member, IEEE) received the B.Eng. degree in automation and the M.S. degree in control science and engineering from the Hebei University of Technology, Tianjin, China, in 2001 and 2004, respectively, and the Ph.D. degree in control science and engineering from the Beijing Institute of Technology, Beijing, China, in 2010. He is currently a Professor with the School of Automation, Beijing Institute of Technology.
His research interests include the areas of autonomous vehicles and robotics, with focus on intelligent navigation, cross-domain collaborative perception, and high mobility motion planning and control in dynamic open environment.
\end{IEEEbiography}

\begin{IEEEbiography}[{\includegraphics[width=1in,height=1.25in,clip,keepaspectratio]{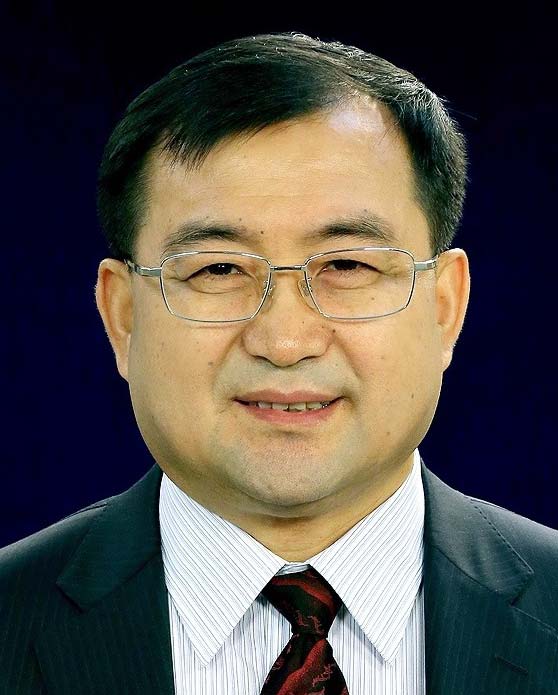}}]{Mengyin Fu} (Member, IEEE) received the B.S. degree from Liaoning University, China, the M.S. degree from the Beijing Institute of Technology, China, and the Ph.D. degree from the Chinese Academy of Sciences.
He was elected as an Academician of the Chinese Academy of Engineering in 2021 and Yangtze River Scholar Distinguished Professor in 2009. He is currently the President of the Nanjing University of Science and Technology. His research interests include integrated navigation, intelligent navigation, image processing, learning, and recognition, as well as their applications.
Prof. Fu won the Guanghua Engineering Science and Technology Award for Youth Award in 2010. In recent years, he has received the National Science and Technology Progress Award, for several times.
\end{IEEEbiography}

\end{document}